\documentclass[10pt, twocolumn, letterpaper]{article}

\usepackage[]{iccv}

\newcommand{\rb}[1]{\textbf{\color{red}#1}}
\newcommand{\gb}[1]{\textbf{\color{green}#1}}
\newcommand{\bb}[1]{\textbf{\color{blue}#1}}

\definecolor{iccvblue}{rgb}{0.21,0.49,0.74}

\usepackage{multirow}
\usepackage{pgfplots, pgfplotstable}
\usepgfplotslibrary{external, statistics, fillbetween}

\usepackage[pagebackref, breaklinks, colorlinks, allcolors=iccvblue]{hyperref}

\def\paperID{10360} % *** Enter the Paper ID here
\def\confName{ICCV}
\def\confYear{2025}

\title{SimBEV: A Synthetic Multi-Task Multi-Sensor Driving Data Generation Tool and Dataset}

\author{Goodarz Mehr\thanks{Work partially completed while at Virginia Tech.} \quad Azim Eskandarian\\
Virginia Commonwealth University\\
{\tt\small \{mehrg, eskandariana\}@vcu.edu}\\
{\tt\small\href{https://github.com/GoodarzMehr/SimBEV}{https://github.com/GoodarzMehr/SimBEV}}
}

\begin{document}
    \maketitle
    
    \begin{abstract}
Bird's-eye view (BEV) perception has garnered significant attention in autonomous driving in recent years, in part because BEV representation facilitates multi-modal sensor fusion. BEV representation enables a variety of perception tasks including BEV segmentation, a concise view of the environment useful for planning a vehicle's trajectory. However, this representation is not fully supported by existing datasets, and creation of new datasets for this purpose can be a time-consuming endeavor. To address this challenge, we introduce SimBEV. SimBEV is a randomized synthetic data generation tool that is extensively configurable and scalable, supports a wide array of sensors, incorporates information from multiple sources to capture accurate BEV ground truth, and enables a variety of perception tasks including BEV segmentation and 3D object detection. SimBEV is used to create the SimBEV dataset, a large collection of annotated perception data from diverse driving scenarios. SimBEV and the SimBEV dataset are open and available to the public.
\end{abstract}    
    \section{Introduction} \label{sec:intro}

Autonomous driving promises a future with safer, cleaner, more efficient and reliable transportation systems \cite{eskandarian2019research, wang2023multimodal}. As development of autonomous vehicle (AV) technology has accelerated in recent years, so has the need for perception algorithms capable of understanding complex driving scenarios in diverse environments \cite{zhao2023autonomous, cui2024survey}. High-quality driving datasets have been at the center of recent progress, serving as a foundation for training and benchmarking novel perception algorithms. It is vital for such datasets to encompass a wide variety of driving scenarios and encapsulate a diverse set of road types, weather conditions, and traffic patterns, so perception models can effectively generalize to real-world situations \cite{liu2024survey, song2023synthetic, guo2019safe}.

As essential in this context is multimodal sensor fusion, which enhances the performance of perception algorithms by compensating for the weaknesses of one modality with the strengths of others \cite{zhuang2021perception, chitta2022transfuser, xu2018pointfusion}. Sensor fusion improves an AV's understanding of its environment \cite{wang2023multisensor} (especially in adverse weather conditions \cite{bijelic2020seeing}), enables robust decision making in dynamic scenarios \cite{huang2020multi, shao2023safety}, and opens the door to perception models capable of performing multiple tasks simultaneously \cite{liu2022bevfusion, natan2022towards, phillips2021deep, huang2023fuller}. Consequently, it is imperative for driving datasets to support a wide array of sensors and perception tasks to facilitate the development of multifaceted perception systems that take advantage of the strengths of different sensing modalities.

Bird's-eye view (BEV) perception has attracted significant attention in recent years for two main reasons \cite{ma2024vision}. First, BEV representation is conducive to the fusion of information from different modalities, perspectives, and agents, and extracted BEV features can be used for various perception tasks. Second, BEV segmentation offers a concise, geometrically accurate, and semantically rich view of the environment, and can be used by motion planning, behavior prediction, and control algorithms. These two factors have led to the proliferation of perception algorithms that use BEV representation for 3D object detection, BEV segmentation, or both \cite{liu2022bevfusion, huang2023fuller, wang2023bi, xiong2023lxl, lin2024rcbevdet, jiao2023msmdfusion, li2022time3d, li2024unimode, zhang2023uni3d, xiong2023cape, man2023bev, wang2023unitr, yang2023bevformer, cai2023bevfusion4d, liu2023petrv2, mohapatra2021bevdetnet, gunn2024lift, liu2024seed, huang2021bevdet, wang2022detr3d, liang2022bevfusion, luo2022detr4d, li2024bevnext, zhao2024maskbev, dutta2022vit, gong2022gitnet, peng2023bevsegformer, xie2022m, xu2023cobevt, li2024fast, zhao2024improving}.

\begin{figure*}[t]
    \centering
    \includegraphics[width=\linewidth]{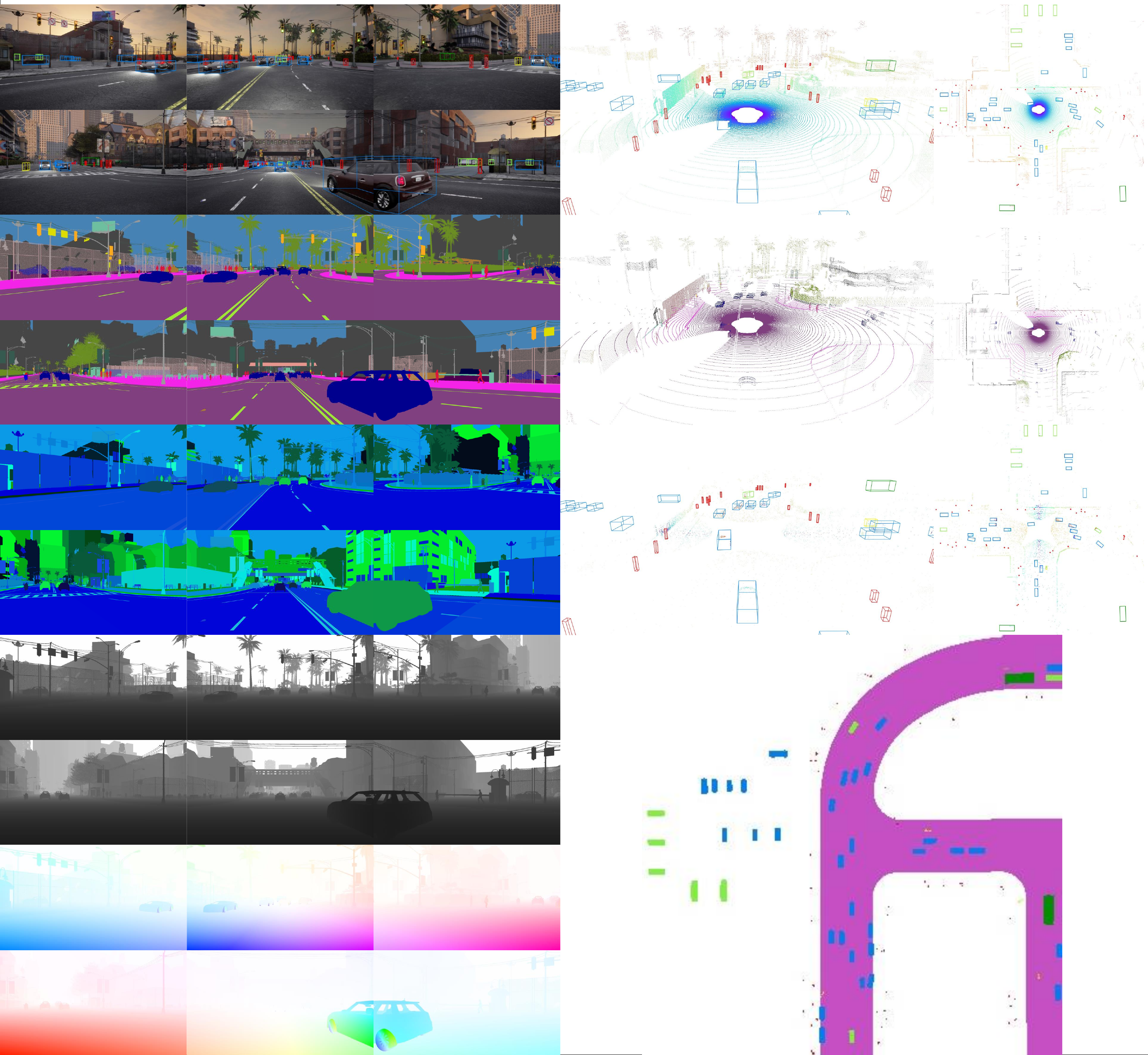}
    \setlength{\abovecaptionskip}{-8 pt}
    \setlength{\belowcaptionskip}{-14 pt}
    \caption{A data sample generated by SimBEV. The left half depicts a 360-degree view of the ego vehicle's surroundings using different camera types (from top to bottom RGB, sematic segmentation, instance segmentation, depth, and optical flow cameras, respectively). On the right half, views of lidar, semantic lidar, radar, and the BEV ground truth are shown from top to bottom, respectively. Some images also contain 3D object bounding boxes colored according to the object's class.}\label{fig:dataset-collage}
\end{figure*}

Despite growing interest in BEV perception, few existing datasets support BEV segmentation. For the ones that do, either BEV ground truth is limited to static map elements (drivable area, pedestrian crossing, etc.) \cite{caesar2020nuscenes}, or BEV ground truth is only provided for a small window around the ego vehicle (obtained by combining map elements with 3D object bounding boxes \cite{houston2021one}) as objects further away may be occluded from the ego vehicle's view.

Creating a new dataset to fill this gap is a challenging endeavor. Real-world driving data require (at least in part) labor-intensive hand annotation and need to be collected over a long period of time to ensure that weather conditions and traffic patterns present in the dataset are diverse \cite{uricar2019challenges, liu2024survey}. On the other hand, synthetic driving data often consist of user-designed scenarios that in most cases do not capture the full diversity of the environment. Moreover, simply capturing the overhead view of the ego vehicle in either case may not be enough to obtain the BEV ground truth due to the presence of vegetation and other structures (traffic lights, bridges, etc.) that obstruct that view \cite{liu2025h}.

\begin{table*}[t]
    \centering
    \footnotesize
    \begin{tabular}{c l c c c c c c c c}
        \toprule
         & \textbf{Dataset} & \textbf{Year} & \textbf{Scenes} & \textbf{Annotated frames} & \textbf{2D Det} & \textbf{3D Det} & \textbf{2D Seg} & \textbf{3D Seg} & \textbf{BEV Seg} \\
        \toprule
        \multirow{11}*{\rotatebox[origin=c]{90}{\textbf{Real-world}}} & KITTI \cite{geiger2013vision} & 2012 & 22 & 41K & \checkmark & \checkmark & - & - & - \\
         & Cityscapes \cite{cordts2016cityscapes} & 2016 & - & 25K & \checkmark & \checkmark & \checkmark & - & - \\
         & Mapillary \cite{neuhold2017mapillary} & 2017 & - & 25K & - & - & \checkmark & - & - \\
         & ApolloScape \cite{huang2018apolloscape} & 2018 & 103 & 144K & \checkmark & \checkmark & \checkmark & \checkmark & - \\
         & Argoverse \cite{chang2019argoverse} & 2019 & 113 & 22K & \checkmark & \checkmark & - & - & limited \\
         & Waymo Open \cite{sun2020scalability} & 2019 & 1150 & 230K & \checkmark & \checkmark & \checkmark & - & - \\
         & nuScenes \cite{caesar2020nuscenes} & 2019 & 1000 & 40K & \checkmark & \checkmark & \checkmark & \checkmark & limited \\
         & A*3D \cite{pham20203d} & 2020 & - & 39K & - & \checkmark & - & - & - \\
         & BDD100K \cite{yu2020bdd100k} & 2020 & 100K & 12M & \checkmark & - & \checkmark & - & - \\
         & Lyft Level 5 \cite{houston2021one} & 2021 & 366 & 46K & - & \checkmark & - & - & limited \\
         & Argoverse 2 \cite{wilson2023argoverse} & 2021 & 1000 & 6M & \checkmark & \checkmark & - & - & limited \\
        \midrule
        \multirow{6}*{\rotatebox[origin=c]{90}{\textbf{Synthetic}}} & SYNTHIA \cite{ros2016synthia} & 2016 & - & 13K & \checkmark & \checkmark & \checkmark & - & - \\
         & GTA-V \cite{richter2016playing} & 2016 & - & 25K & - & - & \checkmark & - & - \\
         & ViPER \cite{simon2005viper} & 2017 & - & 254K & \checkmark & \checkmark & \checkmark & - & - \\
         & All-in-One Drive \cite{weng2023all} & 2021 & 100 & 100K & \checkmark & \checkmark & \checkmark & \checkmark & - \\
         & SHIFT \cite{sun2022shift} & 2022 & 4850 & 2.5M & \checkmark & \checkmark & \checkmark & - & - \\
         & \textbf{SimBEV} & 2024 & 320 & 102K & \checkmark & \checkmark & \checkmark & \checkmark & \checkmark \\
        \bottomrule
    \end{tabular}
    \caption{Comparison of the size and supported tasks of the most notable existing single-vehicle driving datasets. SimBEV is the only dataset that provides full support for BEV perception.} \label{table:dataset-comparison}
\end{table*}

To overcome these challenges, our paper makes two main contributions. First, we introduce SimBEV, a synthetic data generation tool based on CARLA Simulator \cite{dosovitskiy2017carla} that uses domain randomization to create diverse driving scenarios. SimBEV supports a comprehensive array of sensors and incorporates information from multiple sources to capture accurate BEV ground truth and 3D object bounding boxes. It enables a variety of perception tasks, including BEV segmentation and 3D object detection, making it an invaluable tool for computer vision researchers and helping accelerate the development of more capable autonomous driving systems. Second, we use SimBEV to create the SimBEV dataset, a comprehensive large-scale dataset that can serve as a benchmark for a variety of perception tasks. A data sample generated by SimBEV is shown in \cref{fig:dataset-collage}.
    \section{Related Work} \label{sec:related-work}

Real-world driving datasets often target specific subsets of perception tasks, as the high costs associated with data collection and labeling limit their scope. One of the oldest and most prominent driving datasets is the KITTI dataset \cite{geiger2013vision}, which supports depth estimation and 2D/3D object detection and tracking. Other notable image-based datasets include Cityscapes \cite{cordts2016cityscapes}, and Mapillary \cite{neuhold2017mapillary}, which are geared towards segmentation, while A*3D \cite{pham20203d} focuses on 3D object detection. More recently, large-scale datasets such as BDD100K \cite{yu2020bdd100k}, Waymo Open \cite{sun2020scalability}, ApolloScape \cite{huang2018apolloscape}, Argoverse 2 \cite{wilson2023argoverse}, and nuScenes \cite{caesar2020nuscenes} have emerged, offering multi-modal data and multi-task annotations but primarily emphasizing object detection and tracking.

Synthetic driving datasets are compiled using graphics engines and physics simulators. For example, SYNTHIA \cite{ros2016synthia} incorporates RGB and semantically segmented images generated by its dedicated simulator. Video games have also served as a source of data. For instance, GTA-V \cite{richter2016playing} offers RGB and semantically segmented images extracted from GTA. ViPER \cite{simon2005viper} expands on GTA-V by including optical flow images and discrete environmental labels. The introduction of CARLA \cite{dosovitskiy2017carla} fostered systemic generation of driving datasets. All-in-One Drive \cite{weng2023all} is one such dataset, providing support for multiple sensors with a focus on simulating SPAD (Single-Photon Avalanche Detector)-lidars. Another is SHIFT \cite{sun2022shift}, a large-scale multi-task multi-modal dataset for autonomous driving, designed to simulate discrete and continuous changes in weather and traffic conditions to evaluate domain adaptation strategies.

Existing datasets offer limited support for BEV segmentation. In nuScenes \cite{caesar2020nuscenes}, BEV segmentation is only supported for static map elements (drivable area, lane line, pedestrian crossing, etc.). In Lyft Level 5 \cite{houston2021one} and Argoverse \cite{chang2019argoverse, wilson2023argoverse}, BEV ground truth is obtained by combining map elements and vehicle bounding boxes observable by the ego vehicle's perception sensors, limiting BEV ground truth area and/or missing occluded objects.

Some vehicle-to-everything (V2X) datasets provide limited support for BEV segmentation as well. H-V2X \cite{liu2025h} captures the BEV ground truth using overhead cameras installed along a 100 km highway, with data limited to highway driving and mostly suitable for highway motion forecasting. CARLA-based OPV2V \cite{xu2022opv2v} dataset provides the BEV ground truth, but only for the drivable area, lane line, and vehicle classes, and the one for drivable area can be inaccurate due to its sole reliance on CARLA-generated waypoints. Finally, CARLA-based V2X-Sim \cite{li2022v2x} dataset captures the BEV ground truth for several classes using an overhead camera, which can be inaccurate due to the presence of vegetation, traffic light poles, and other structures that obstruct the overhead view.
    \section{SimBEV} \label{sec:simbev}

\begin{figure*}[t]
    \centering
    \includegraphics[width=\linewidth]{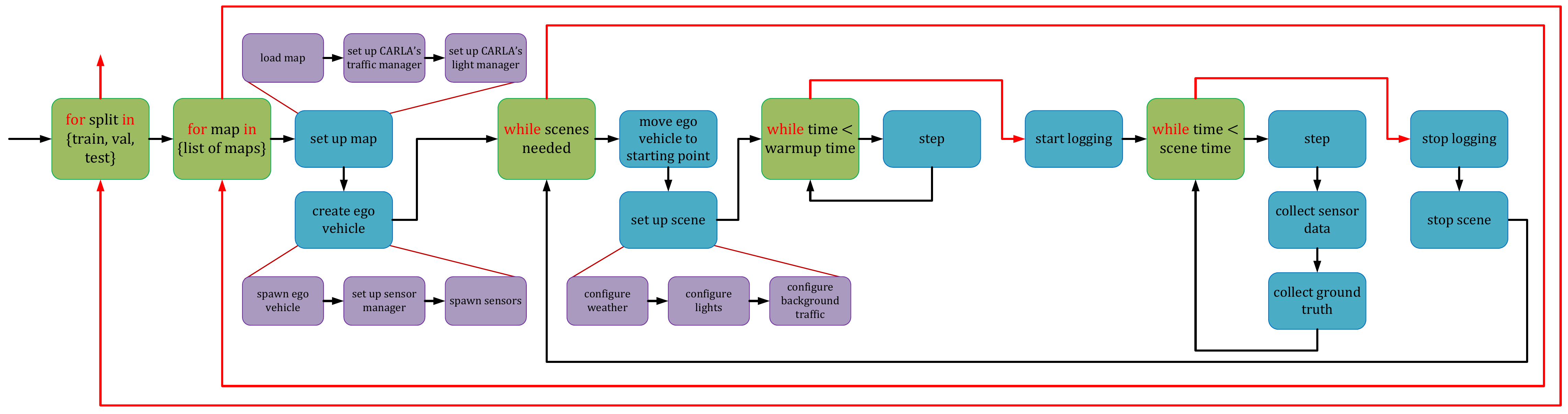}
    \setlength{\abovecaptionskip}{-8 pt}
    \setlength{\belowcaptionskip}{-8 pt}
    \caption{SimBEV's logic flow when creating a new dataset. The arrow exiting green nodes at the top indicates the action taken when the condition in that node is no longer satisfied.}\label{fig:simbev}
\end{figure*}

SimBEV relies on CARLA 0.9.15 \cite{dosovitskiy2017carla} equipped with a custom content library (see the Supplementary Material) to simulate the environment, perception sensors, and traffic behavior, although it is compatible with the standard release of CARLA as well. SimBEV streamlines, automates, and manages the entire data collection process for the user (who controls it through a single configuration file) by manipulating simulation elements through CARLA's Python API. It equips the user with CARLA's customizability when necessary and takes charge when not. This flexibility and ease of use enables researchers to quickly create custom datasets that suit their needs.

\subsection{Design} \label{subsec:simbev-overview}

SimBEV works by randomizing (within some bounds) as many simulation parameters as possible to create a statistically diverse yet realistic set of scenarios. To create a dataset, SimBEV generates and collects data from consecutive episodes, or scenes, each with a unique configuration. 

SimBEV's logic flow is shown in \cref{fig:simbev}. To start, the user configures the desired number of scenes for each map (i.e., the driving environment, can be an existing CARLA map or a custom one) for the training, validation, and test sets. SimBEV checks to see if a SimBEV dataset already exists. If so, it subtracts the number of existing scenes in that dataset from the number of desired scenes for each map. This lets the user expand an already existing SimBEV dataset, or continue with dataset creation in the event of a crash. SimBEV also lets the user replace individual scenes.

At the start of each scene, SimBEV creates uniformly distributed waypoints a certain distance (specified by the user) apart from each other across the map's roads. It selects one at random and spawns the ego vehicle and attached sensors there, though the user has the option to specify a set of spawn coordinates that SimBEV must choose from instead. SimBEV then configures the weather randomly and, if at night, changes the intensity of street lights at random.

For background traffic, SimBEV randomly selects the desired number of vehicles and pedestrians, although the user can specify each. SimBEV then uses the waypoints mentioned earlier to spawn random vehicles at random locations, and spawns pedestrians randomly on walkable areas of the map. CARLA's traffic manager controls the behavior of vehicles and pedestrians throughout the scene.

Because all vehicles and pedestrians start from rest, SimBEV runs the simulation for a user-specified period of time (called warm-up duration) to reach a more realistic state, before collecting data for a user-specified period of time. During that period, SimBEV saves data from the desired sensors at each time step, and calculates and saves both 3D object bounding boxes and the BEV ground truth. Following that, it saves meta-information about the collected data and a log of the scene, destroys the vehicles, pedestrians, and sensors, and moves on to the next scene.

\subsection{Sensors} \label{subsec:simbev-sensors}

SimBEV supports a variety of sensors available in CARLA, including five different camera types (RGB, semantic segmentation, instance segmentation, depth, and optical flow), two different lidar types (regular and semantic), radar, GNSS, and IMU, as shown in \cref{fig:dataset-collage}. The user has full control over each sensor type's parameters (e.g. a camera's resolution or FoV), but the placement of sensors is fixed for now. Similar to \cite{caesar2020nuscenes}, cameras are placed at six locations above the vehicle to offer a 360-degree view of the vehicle's surroundings, while a radar is placed on each of the four sides of the vehicle. The GNSS and IMU are placed at the center of the vehicle (the origin of the vehicle's coordinate system), and a lidar is placed high above that center.

\subsection{Scene configuration} \label{subsec:simbev-config}

As will be discussed in what follows, numerous parameters are randomized for each scene to ensure that generated scenes are as unique and diverse as possible.

\paragraph{Weather.} \label{par:weather}

Weather in CARLA is controlled using several parameters such as fog density, sun altitude angle, wind intensity, etc. By default, SimBEV randomly selects these parameters for each scene to create diverse weather conditions (subject to some constraints to ensure the realism of the weather), but the user also has the option to set any of the parameters to a fixed value. For instance, setting the sun altitude angle to anything less than zero creates a dataset of night-time scenes.

\begin{figure}[t]
    \centering
    \includegraphics[width=\linewidth]{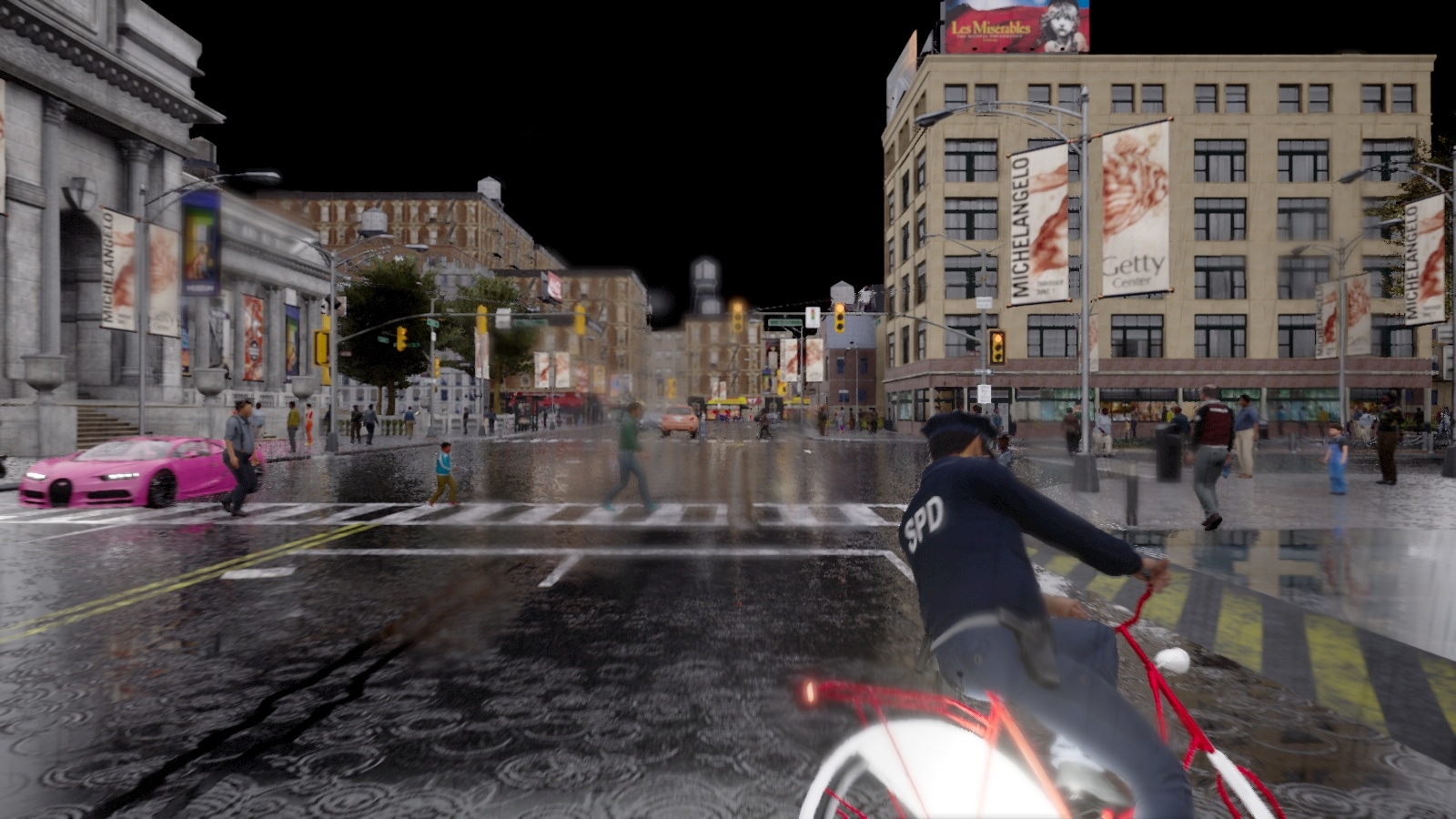}
    \setlength{\abovecaptionskip}{-8 pt}
    \setlength{\belowcaptionskip}{-8 pt}
    \caption{In a scene generated by SimBEV, a reckless ego vehicle runs over a cyclist.}\label{fig:reckless}
\end{figure}

\paragraph{Traffic.} \label{par:traffic}

SimBEV randomly selects background vehicles from CARLA's vehicle library, which includes sedans, vans, trucks, heavy goods vehicles (HGVs), buses, bicycles, motorcycles, and emergency vehicles, whose emergency lights are turned on randomly. When possible, vehicle colors are selected at random from a vast set of available colors (e.g. a sedan can change colors but a firetruck will only be red). Some vehicles support articulated doors, so when these vehicles come to a stop - e.g. at a traffic light - SimBEV may randomly open one or all of their doors.

SimBEV randomly chooses each vehicle's maximum speed (relative to the speed limit, e.g. 10\% over/under) and how close vehicles can get to each other when coming to a stop. It also randomly selects how long each traffic light stays green. However, the user always has the option to set any of these parameters to a fixed value.

SimBEV randomly chooses pedestrians from CARLA's walker library (which contains models of different age, gender, race, and body type), sets their walking speed at random, and gives each a random destination to go to.

\paragraph{Lights.} \label{par:lights}

SimBEV gives the user the option to turn off all street and/or building lights at night. It also lets the user randomize building light colors, and/or change the intensity of all street lights by a fixed, if desired random, value. In addition, SimBEV randomly turns off individual street lights based on a probability set by the user to simulate broken street lights in the real world.

\paragraph{Reckless driving and jaywalking.} \label{par:reckless}

If desired by the user, some vehicles (including the ego vehicle) can drive recklessly, ignoring traffic lights, traffic signs, and collisions with other vehicles and pedestrians, as shown in \cref{fig:reckless}. The user controls the likelihood of reckless driving, which can result in interesting edge cases. The user also has control over the share of pedestrians allowed to jaywalk and cross the road at any point, not just at crosswalks.

\begin{figure}[t]
    \centering
    \includegraphics[width=\linewidth]{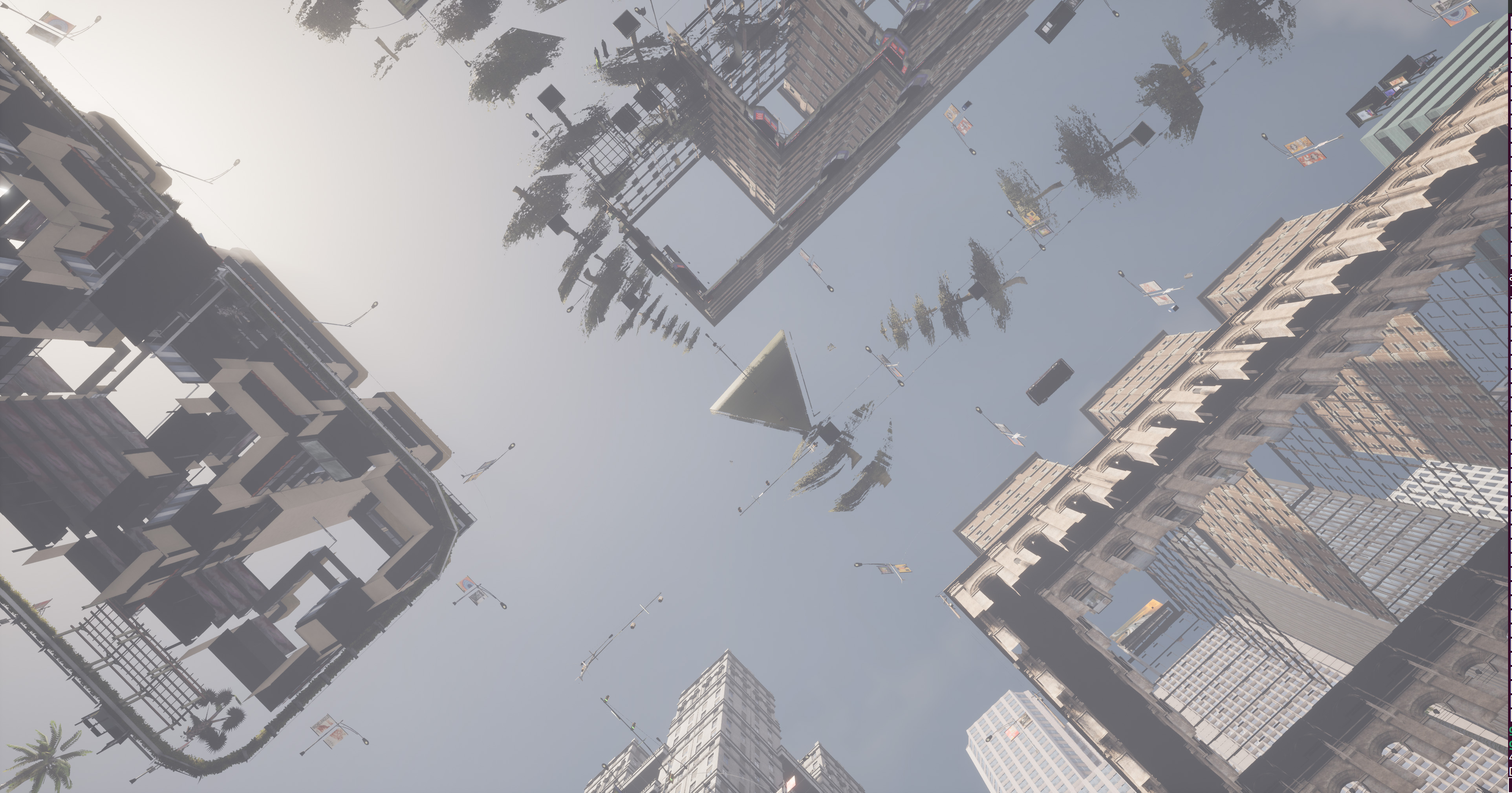}
    \setlength{\abovecaptionskip}{-8 pt}
    \setlength{\belowcaptionskip}{-8 pt}
    \caption{Ground elements (roads, sidewalks, etc.) in CARLA use one-way visible materials, appearing invisible to a camera placed below them. We use this property to capture accurate BEV ground truth by placing a camera below the ego vehicle looking up.}\label{fig:invis-road}
\end{figure}

\subsection{Data annotation} \label{subsec:simbev-gt}

SimBEV offers two main types of data annotation: 3D object bounding boxes and BEV ground truth. The output of some perception sensors such as segmentation, depth, and optical flow cameras and semantic lidar can serve as annotation as well, but we do not discussed them here.

\paragraph{3D object bounding boxes.} \label{par:bev-bbox}

At each time step, SimBEV collects 3D object bounding boxes that are within a user-configurable radius of the ego vehicle for the following six classes: \textit{car}, \textit{truck} (includes trucks, vans, HGVs, etc., but not buses), \textit{bus}, \textit{motorcycle}, \textit{bicycle}, and \textit{pedestrian}. Other object attributes are also collected alongside each bounding box, such as the object's ID, its linear and angular velocity, and its make, model, and color if the object is a vehicle. An optional post-processing step calculates the number of lidar and radar points that fall within each bounding box and adds a \textit{valid} label to boxes with at least one point inside, \textit{invalid} otherwise. This labeling is useful for training 3D object detection algorithms, as it can filter out objects that may not be visible to perception sensors \cite{caesar2020nuscenes}.

\begin{figure*}[t]
    \centering
    \includegraphics[width=0.32\linewidth]{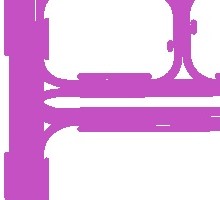}
    \includegraphics[width=0.32\linewidth]{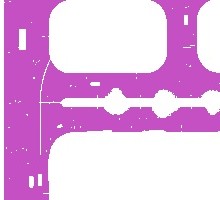}
    \includegraphics[width=0.32\linewidth]{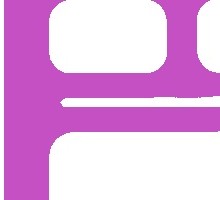}
    \setlength{\abovecaptionskip}{8 pt}
    \setlength{\belowcaptionskip}{-8 pt}
    \caption{Left: BEV road data calculated using CARLA-generated waypoints; there are clear gaps where lanes diverge. Middle: BEV road data obtained from the overhead camera; vehicles and vegetation obstruct a portion of the view. Right: BEV \textit{road} ground truth obtained by combining the two sources and performing \textit{binray closing}.}\label{fig:road-mask}
\end{figure*}

\paragraph{BEV ground truth.} \label{par:bev-gt}

SimBEV supports the following eight classes for BEV segmentation: \textit{road}, \textit{car}, \textit{truck}, \textit{bus}, \textit{motorcycle}, \textit{bicycle}, \textit{rider} (human on a \textit{motorcycle} or \textit{bicycle}), and \textit{pedestrian}. At each time step, the BEV ground truth is saved as a $C \times l \times l$ binary array, where $C$ is the number of classes and $l$ is the dimension of the BEV grid that is centered on the ego vehicle.

To calculate the BEV ground truth for non-\textit{road} classes, we take advantage of the fact that ground elements in CARLA (roads, sidewalks, etc.) use one-way visible materials, appearing solid from one direction and see-through from the opposite, as shown in \cref{fig:invis-road}. This means that we can place a semantic segmentation camera 1 km above the ego vehicle facing down (far enough to minimize perspective distortion) and another 1 km below the ego vehicle facing up to catch what the overhead camera cannot see due to obstructions. Both cameras have a $l \times l$ resolution and their field of view (FoV) is calculated so that each pixel represents a $d \times d$ area on the ground. The BEV ground truth for each non-\textit{road} class is obtained by merging data from the two cameras using a \textit{logical or} operation. By default, $l$ is set to 360 and $d$ is set to 0.4 m, creating a 144 m $\times$ 144 m box around the ego vehicle. This area is larger than what is typically used for BEV segmentation (100 m $\times$ 100 m), but it can help with data augmentation (rotation, translation, scaling) during training.

We follow an approach similar to \cite{xu2022opv2v} to obtain the ground truth for the \textit{road} class. Specifically, we use CARLA-generated waypoints a small distance apart from each other (specified by the user, we recommend setting it to $d$) across the map's roads and note each waypoint's lane width. We then calculate the mutual distance between these waypoints and the center of each cell of a $l \times l$ BEV grid that is centered on the ego vehicle, where each cell represents a $d \times d$ area. For each grid cell, if a waypoint exists whose distance to the center of that cell is less than that waypoint's lane width, that cell is labeled as \textit{road}. Where our approach differs from \cite{xu2022opv2v} is that we then combine this information with data from the overhead camera and perform \textit{binary closing} to patch any potential gaps, obtaining a much more accurate ground truth. This process is illustrated in \cref{fig:road-mask}.

In general, our method allows us to assign multiple labels to the same cell. For example, a cell occupied by a cyclist will have a \textit{rider} (obtained from the overhead camera), a \textit{bicycle} (obtained from the underground camera), and a \textit{road} (calculated using CARLA-generated waypoints) label.

Our approach works everywhere except when roads with large elevation differences are near the ego vehicle, e.g. when the ego vehicle is traveling under an overpass. In those situations, we do not use the overhead or underground cameras. Instead, we rely on CARLA-generated waypoints to calculate the BEV ground truth for the \textit{road} class and use 3D object bounding boxes to calculate the BEV ground truth for other classes. Although not as accurate as our overall approach, the resulting ground truth is still acceptable. SimBEV switches to this method when it detects two waypoints within 48.0 m of each other that have an elevation difference greater than 6.4 m, signaling that they are on two different roads.

    \section{The SimBEV Dataset} \label{sec:simbev-dataset}

To showcase SimBEV, we used it to create the SimBEV dataset, a collection of 320 scenes spread across all 11 CARLA maps according to \cref{table:simbev-split}. To the best of our knowledge, this is the first dataset that utilizes CARLA's largest maps, i.e., Town12, Town13, and Town15. Because Town13 shares many common features with Town12 but uses different building styles, textures, and vegetation, it is not included in the train set to evaluate the generalization performance of trained models and expose overfitting.

\begin{table}[t]
    \centering
    \footnotesize
    \begin{tabular}{c c c c}
        \toprule
        \textbf{Map} & \textbf{Train} & \textbf{Validation} & \textbf{Test} \\
        \toprule
        Town01 & 8 & 2 & 2 \\
        Town02 & 8 & 2 & 2 \\
        Town03 & 20 & 4 & 4 \\
        Town04 & 20 & 4 & 4 \\
        Town05 & 20 & 4 & 4 \\
        Town06 & 20 & 4 & 4 \\
        Town07 & 20 & 4 & 4 \\
        Town10HD & 20 & 4 & 4 \\
        Town12 & 48 & 8 & 8 \\
        Town13 & 0 & 8 & 8 \\
        Town15 & 36 & 6 & 6 \\
        \bottomrule
        \textbf{Total} & 220 & 50 & 50 \\
        \bottomrule
    \end{tabular}
    \setlength{\abovecaptionskip}{4 pt}
    \setlength{\belowcaptionskip}{-8 pt}
    \caption{Distribution of the scenes of the SimBEV dataset across all available CARLA maps.} \label{table:simbev-split}
\end{table}
\setlength{\lineskip}{1pt}
\begin{figure*}[t]
    \centering
    \setlength{\lineskip}{0pt}
    \includegraphics[width=0.142857\linewidth]{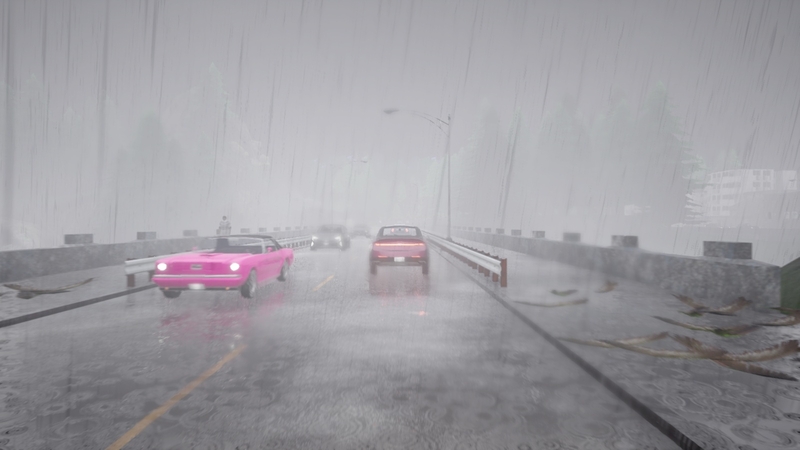}\hspace{-2.5pt}
    \includegraphics[width=0.142857\linewidth]{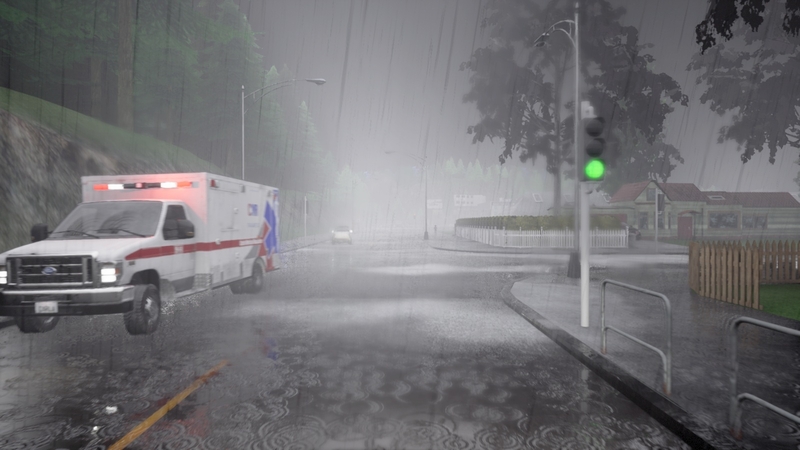}\hspace{-2.5pt}
    \includegraphics[width=0.142857\linewidth]{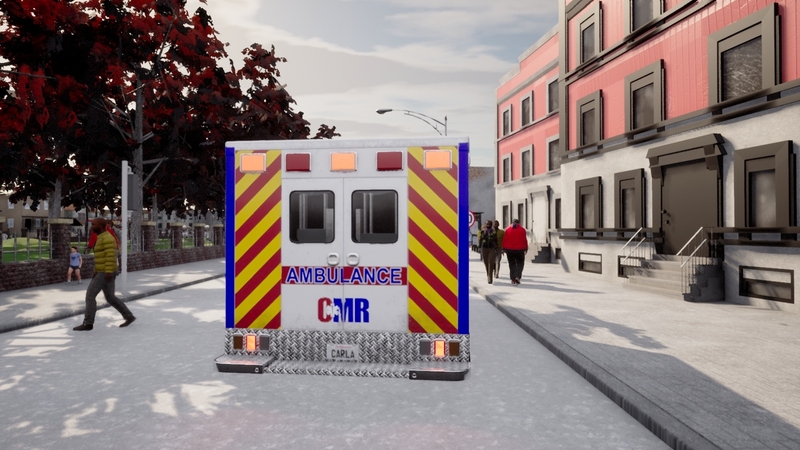}\hspace{-2.5pt}
    \includegraphics[width=0.142857\linewidth]{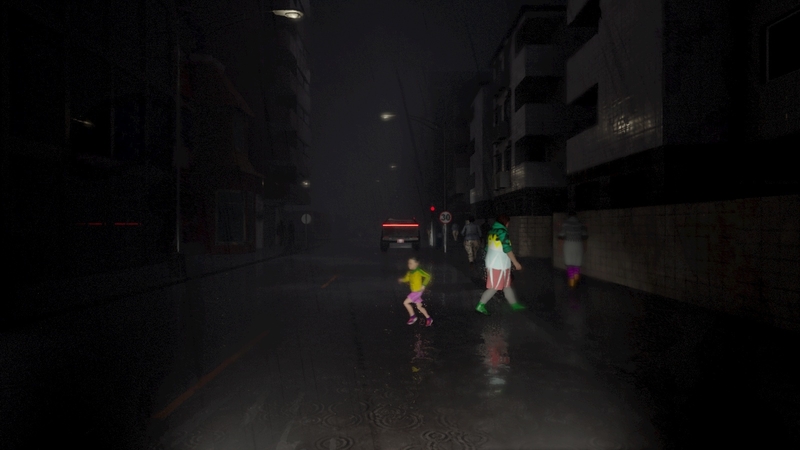}\hspace{-2.5pt}
    \includegraphics[width=0.142857\linewidth]{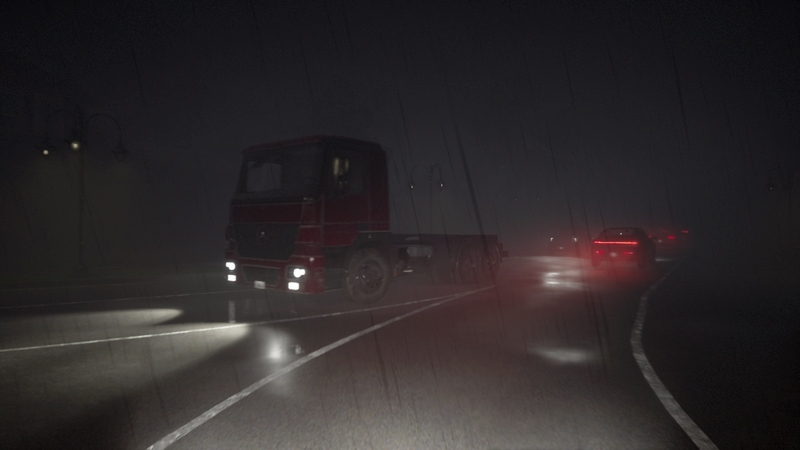}\hspace{-2.5pt}
    \includegraphics[width=0.142857\linewidth]{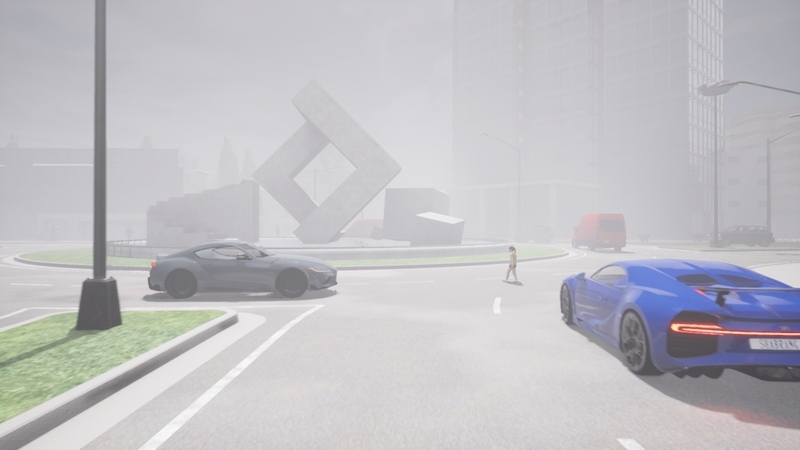}\hspace{-2.5pt}
    \includegraphics[width=0.142857\linewidth]{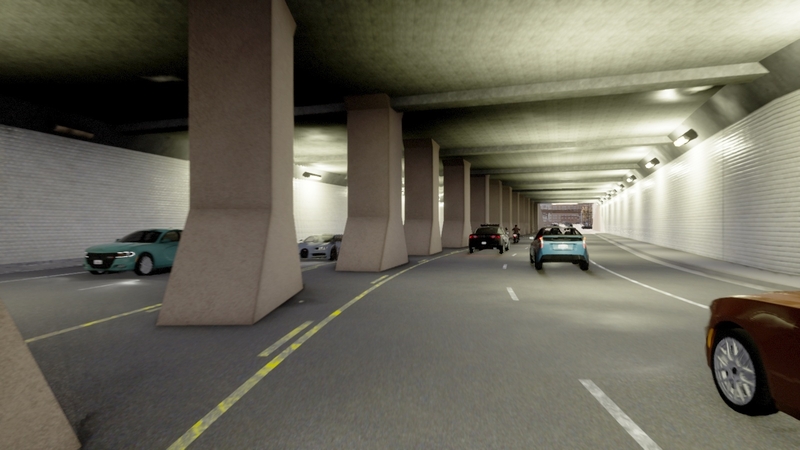}
    \includegraphics[width=0.142857\linewidth]{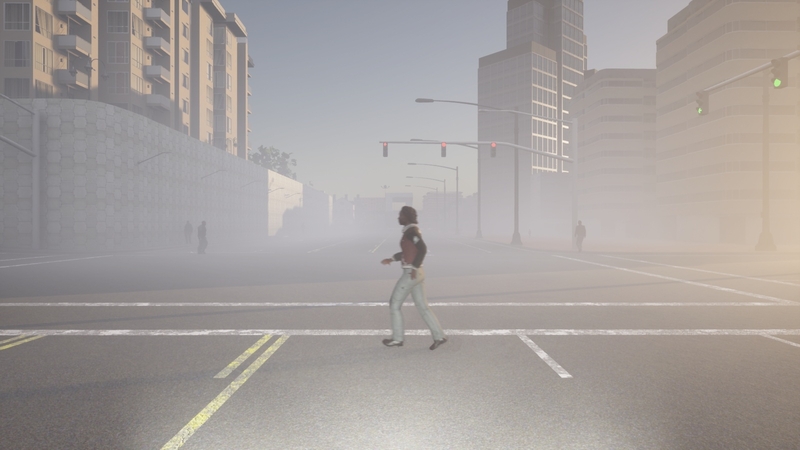}\hspace{-2.5pt}
    \includegraphics[width=0.142857\linewidth]{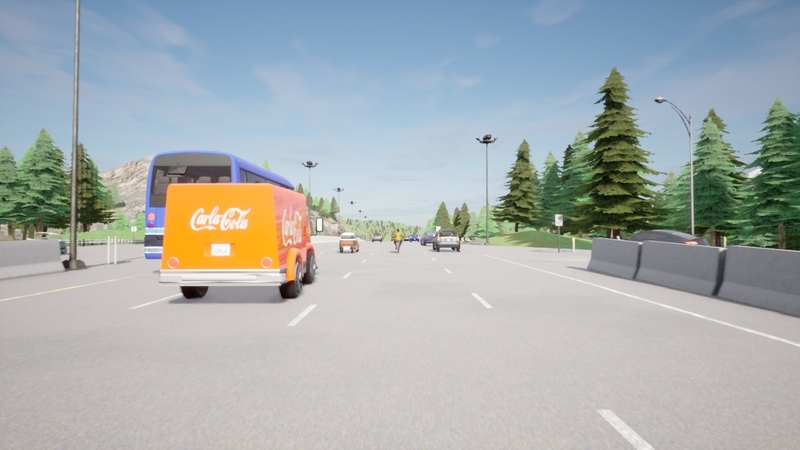}\hspace{-2.5pt}
    \includegraphics[width=0.142857\linewidth]{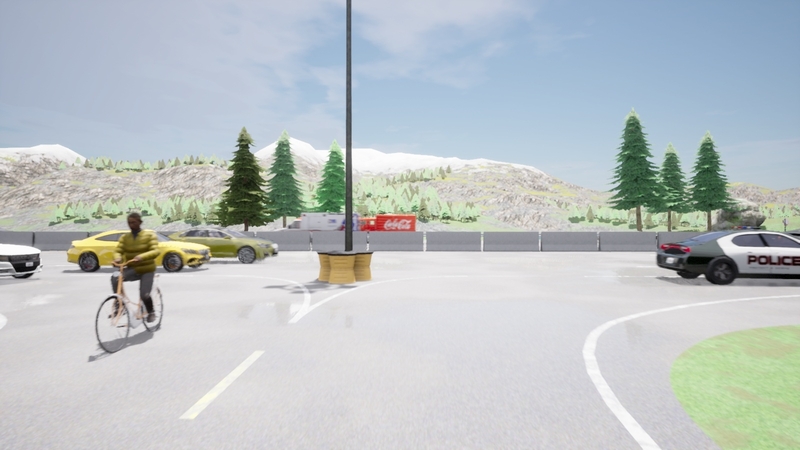}\hspace{-2.5pt}
    \includegraphics[width=0.142857\linewidth]{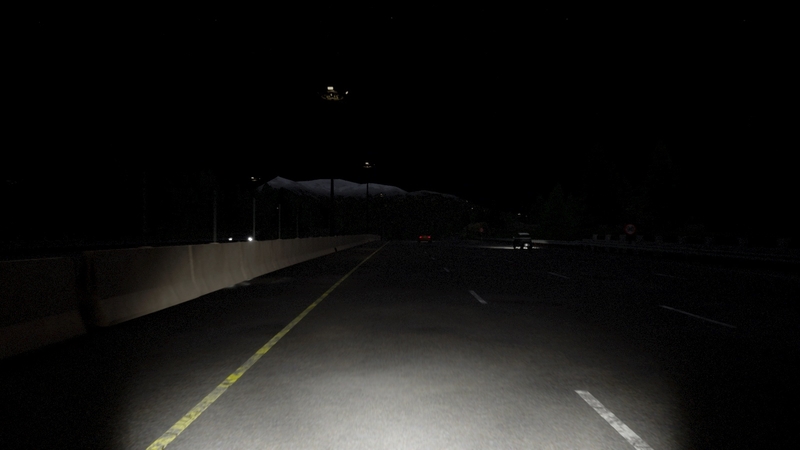}\hspace{-2.5pt}
    \includegraphics[width=0.142857\linewidth]{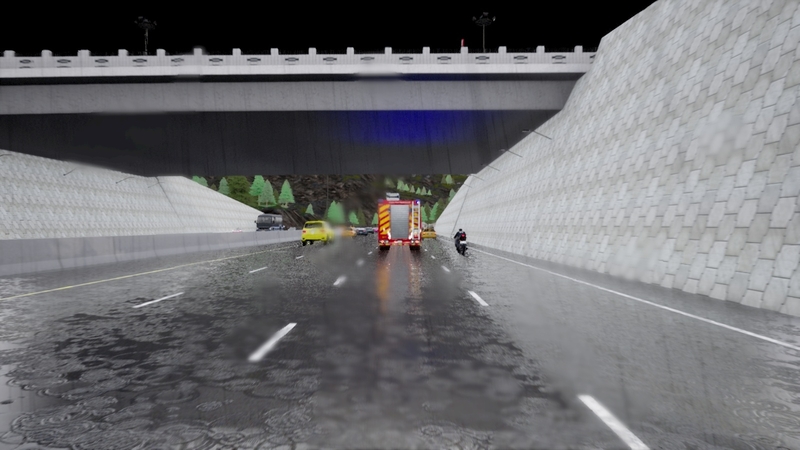}\hspace{-2.5pt}
    \includegraphics[width=0.142857\linewidth]{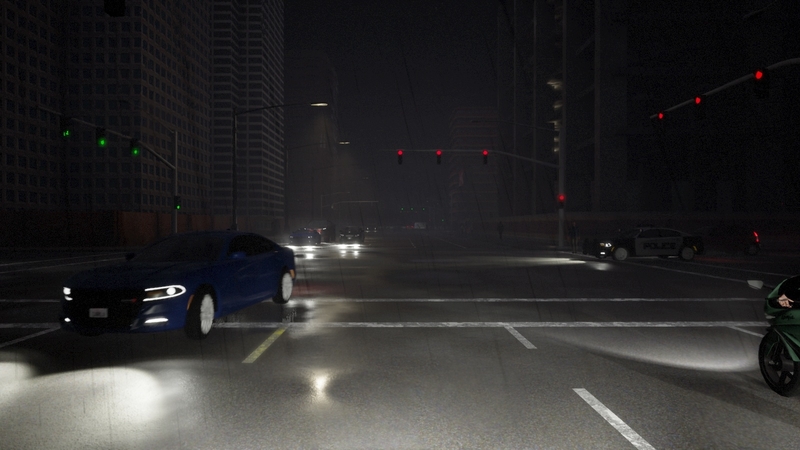}\hspace{-2.5pt}
    \includegraphics[width=0.142857\linewidth]{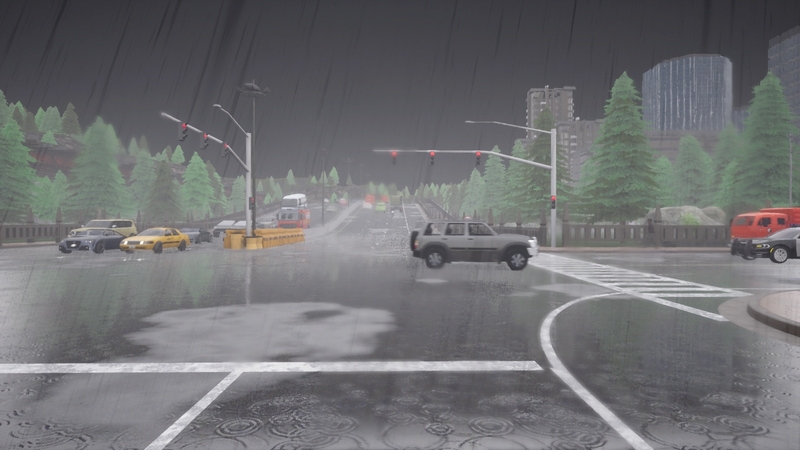}
    \includegraphics[width=0.142857\linewidth]{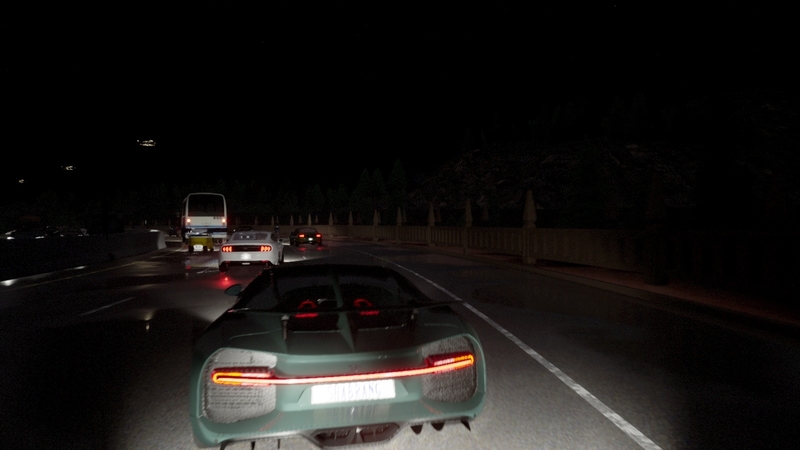}\hspace{-2.5pt}
    \includegraphics[width=0.142857\linewidth]{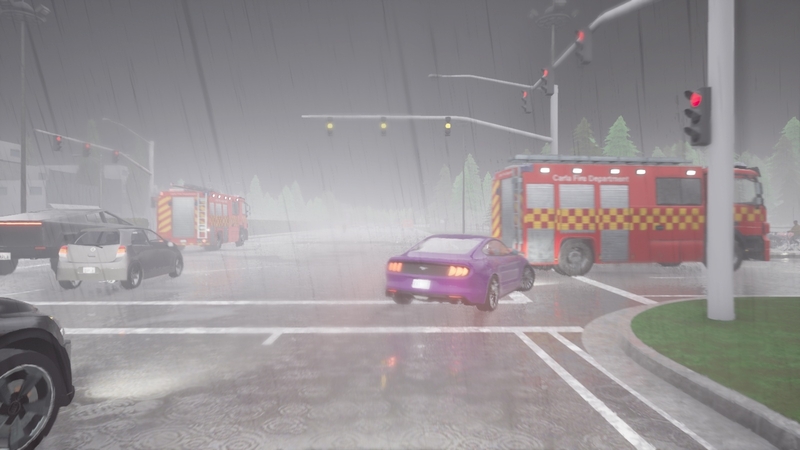}\hspace{-2.5pt}
    \includegraphics[width=0.142857\linewidth]{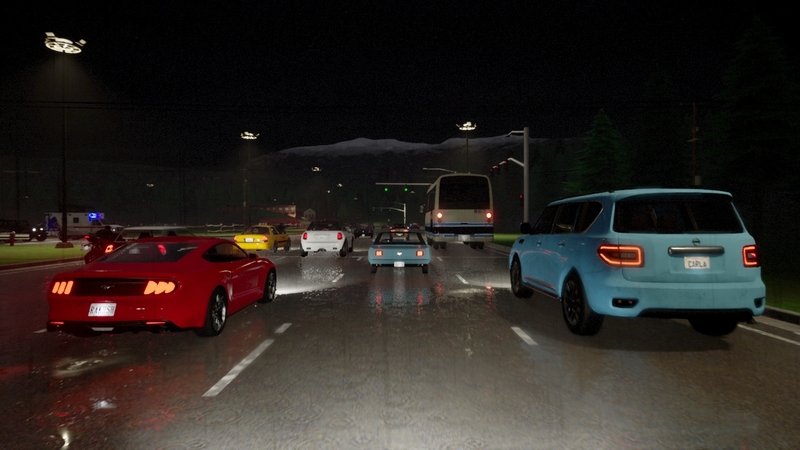}\hspace{-2.5pt}
    \includegraphics[width=0.142857\linewidth]{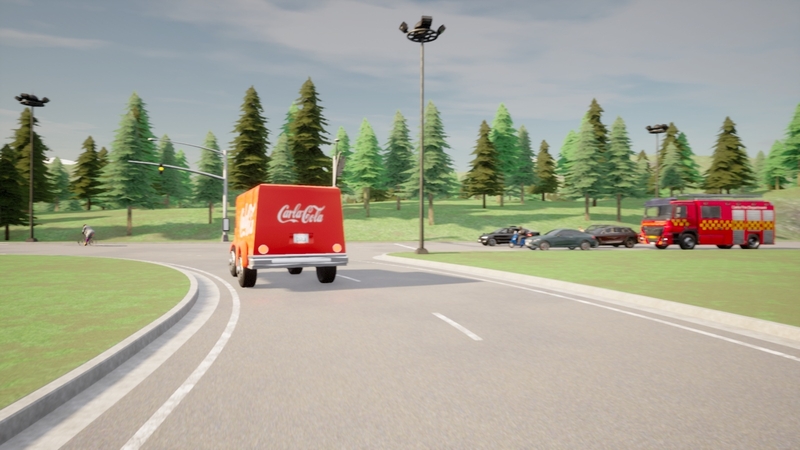}\hspace{-2.5pt}
    \includegraphics[width=0.142857\linewidth]{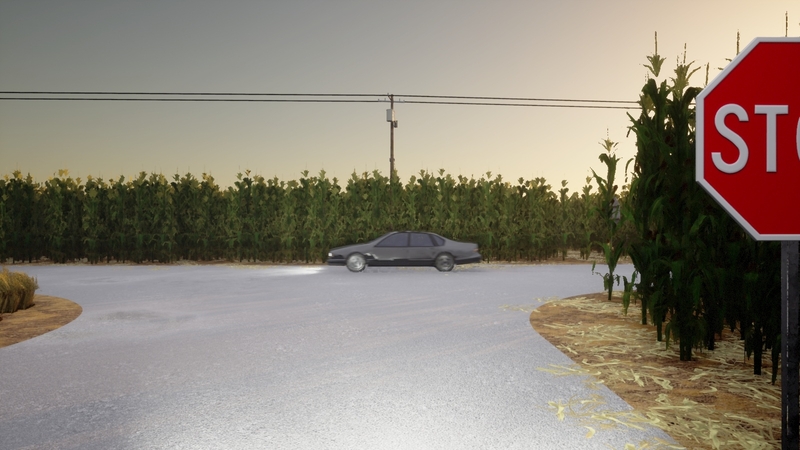}\hspace{-2.5pt}
    \includegraphics[width=0.142857\linewidth]{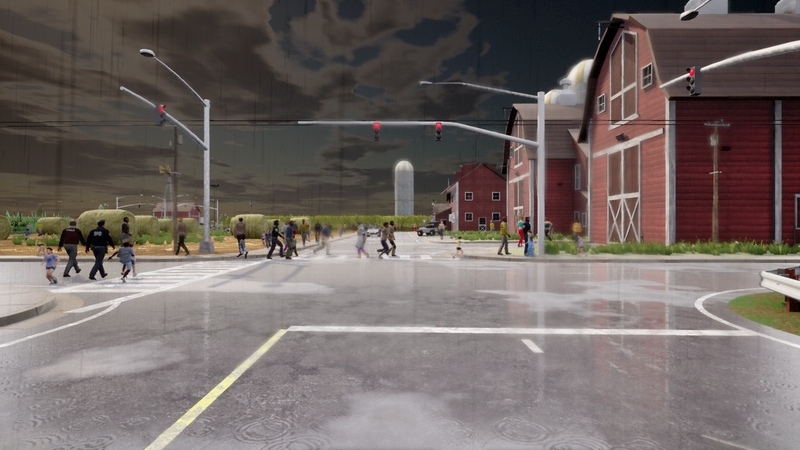}\hspace{-2.5pt}
    \includegraphics[width=0.142857\linewidth]{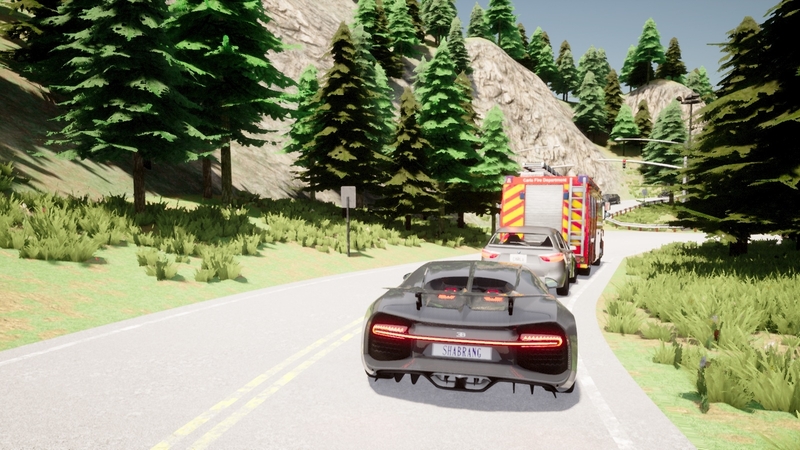}
    \includegraphics[width=0.142857\linewidth]{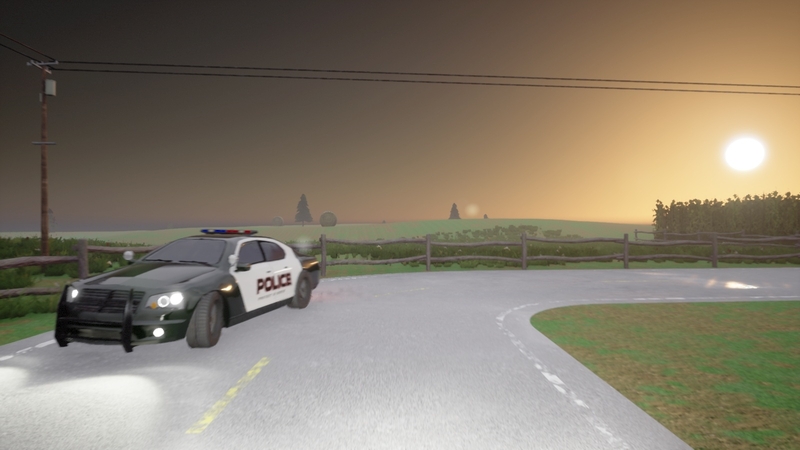}\hspace{-2.5pt}
    \includegraphics[width=0.142857\linewidth]{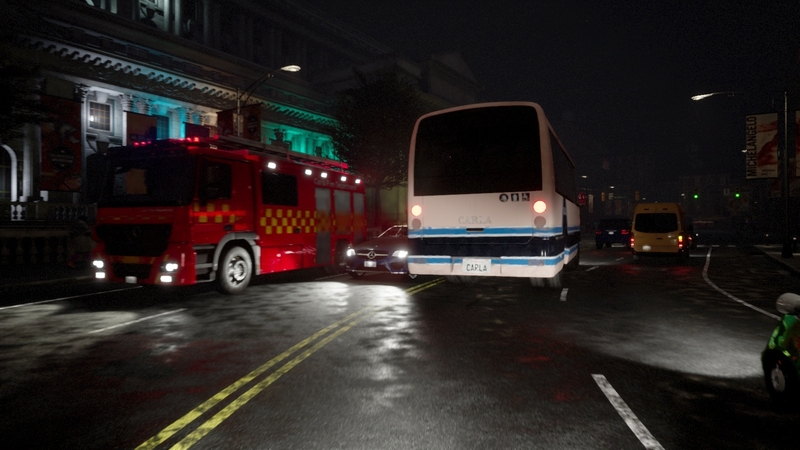}\hspace{-2.5pt}
    \includegraphics[width=0.142857\linewidth]{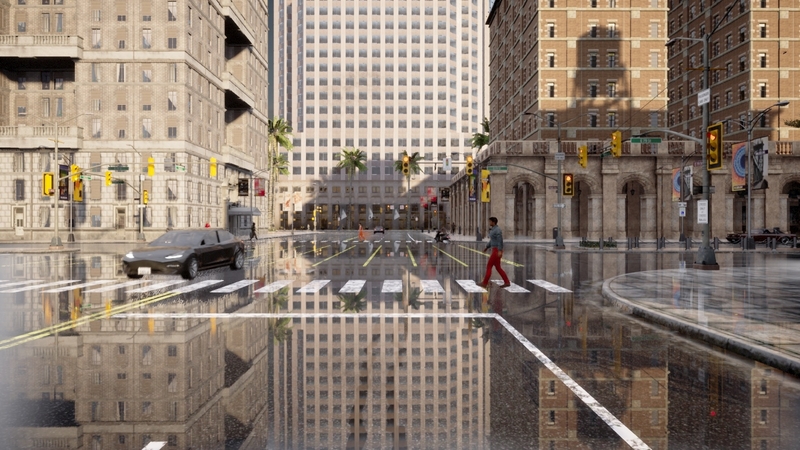}\hspace{-2.5pt}
    \includegraphics[width=0.142857\linewidth]{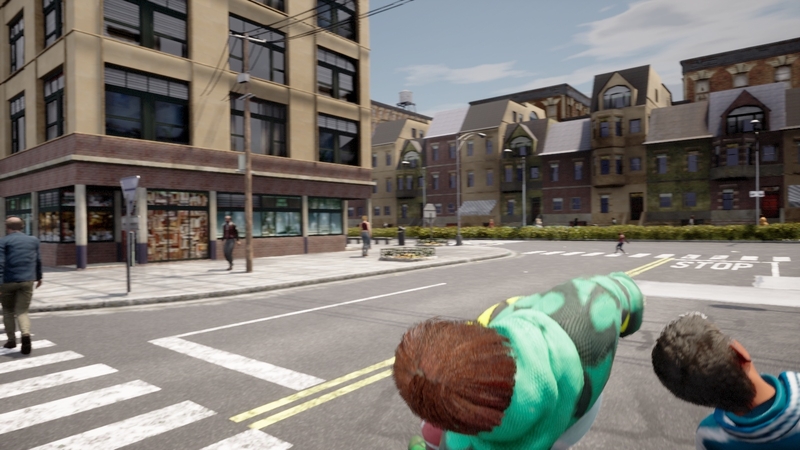}\hspace{-2.5pt}
    \includegraphics[width=0.142857\linewidth]{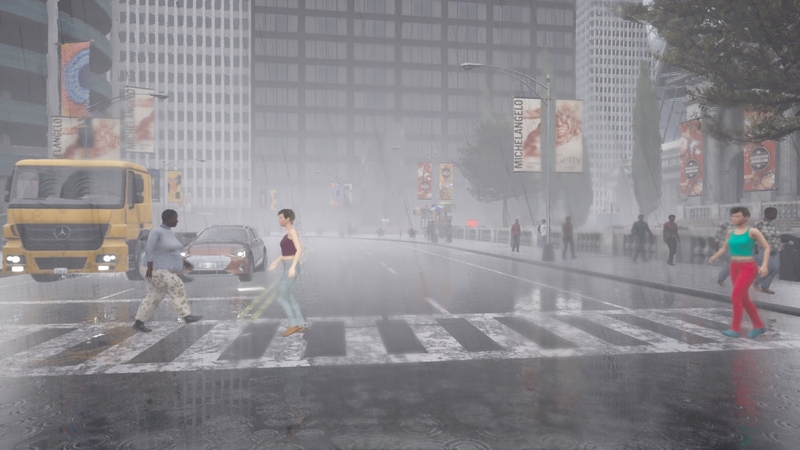}\hspace{-2.5pt}
    \includegraphics[width=0.142857\linewidth]{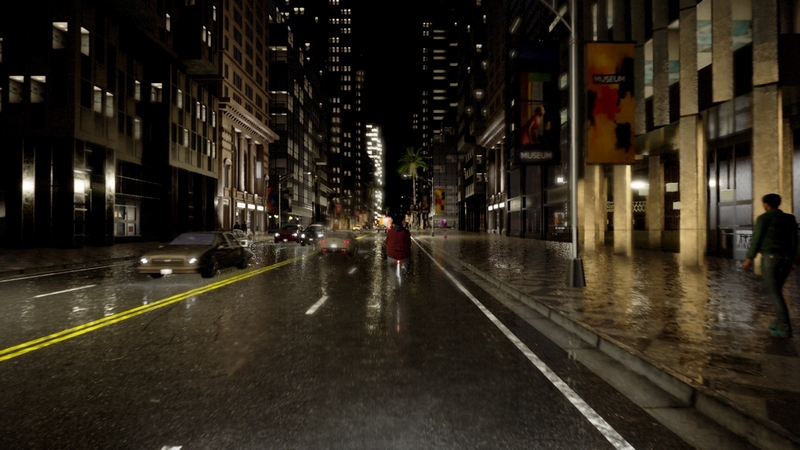}\hspace{-2.5pt}
    \includegraphics[width=0.142857\linewidth]{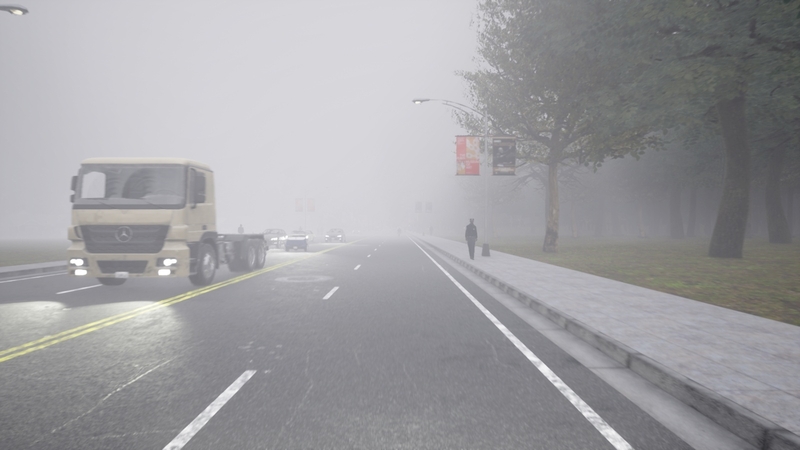}
    \includegraphics[width=0.142857\linewidth]{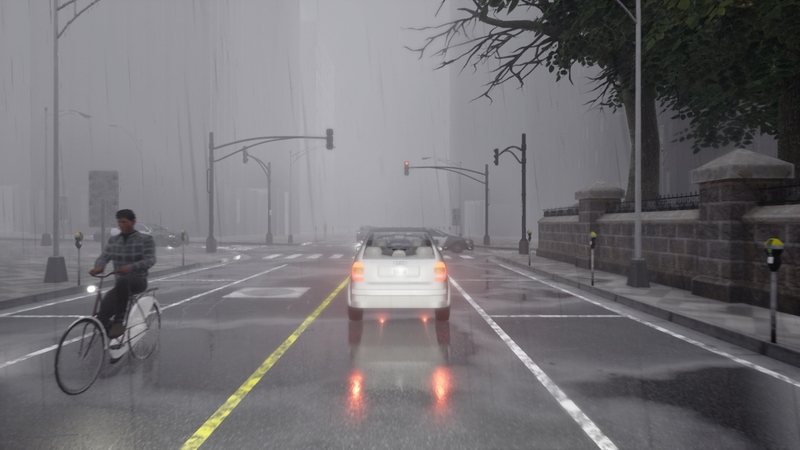}\hspace{-2.5pt}
    \includegraphics[width=0.142857\linewidth]{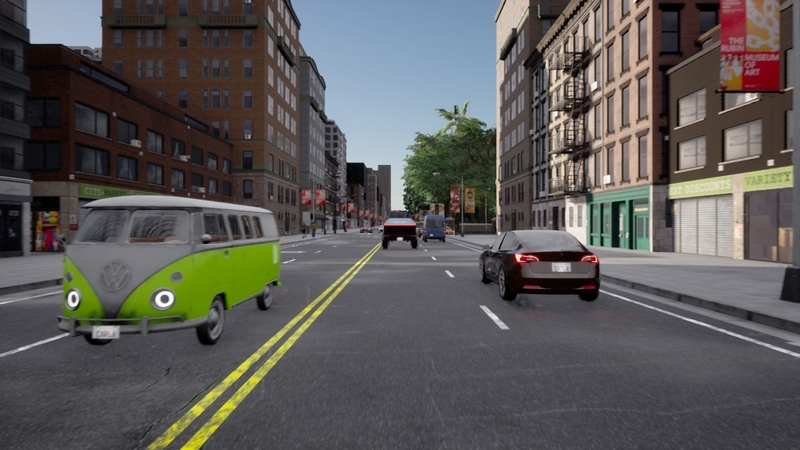}\hspace{-2.5pt}
    \includegraphics[width=0.142857\linewidth]{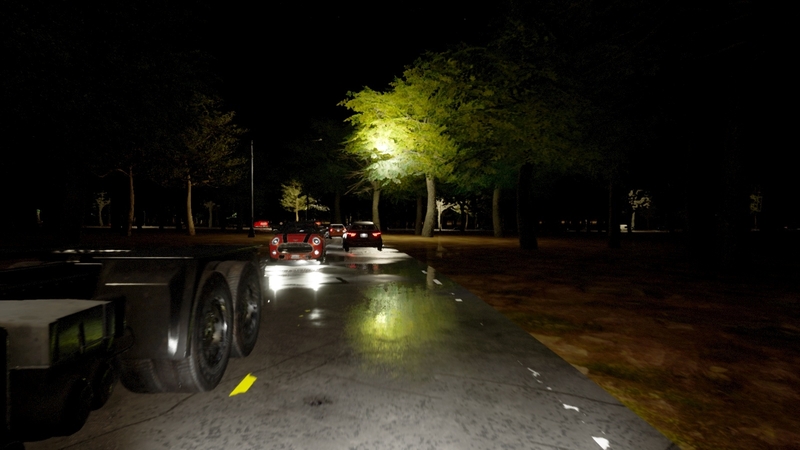}\hspace{-2.5pt}
    \includegraphics[width=0.142857\linewidth]{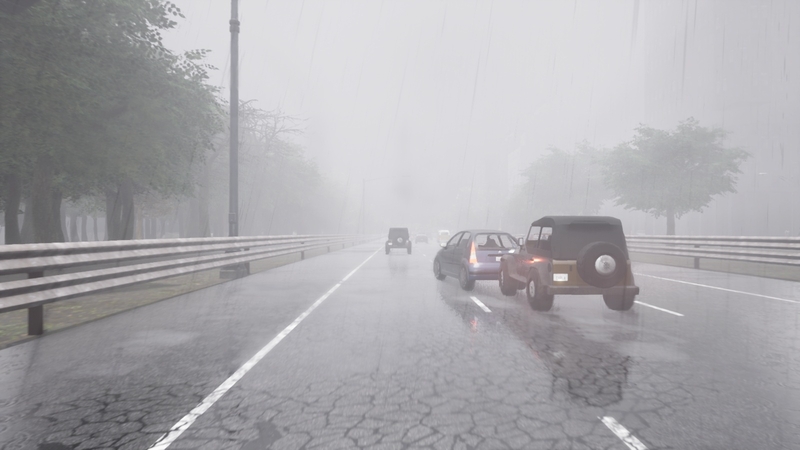}\hspace{-2.5pt}
    \includegraphics[width=0.142857\linewidth]{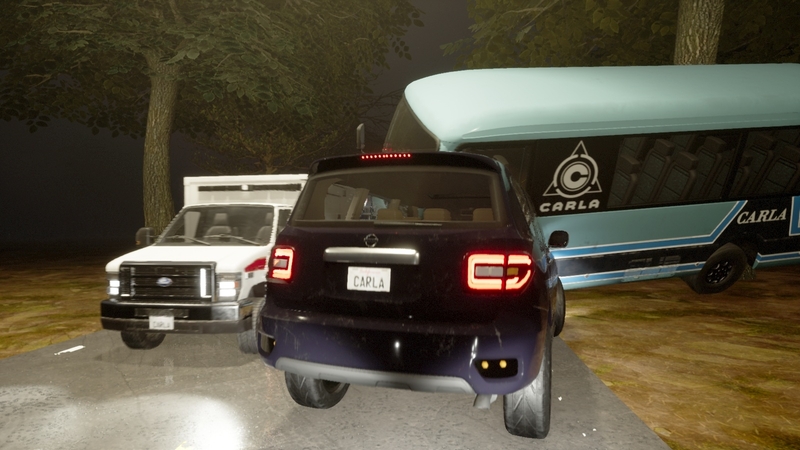}\hspace{-2.5pt}
    \includegraphics[width=0.142857\linewidth]{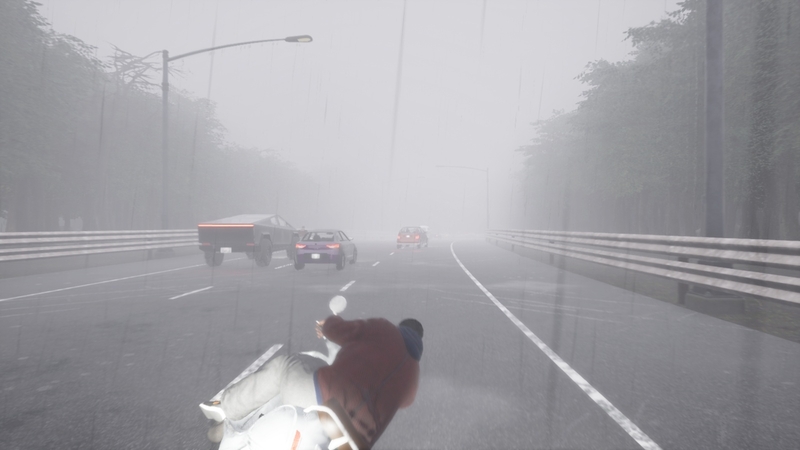}\hspace{-2.5pt}
    \includegraphics[width=0.142857\linewidth]{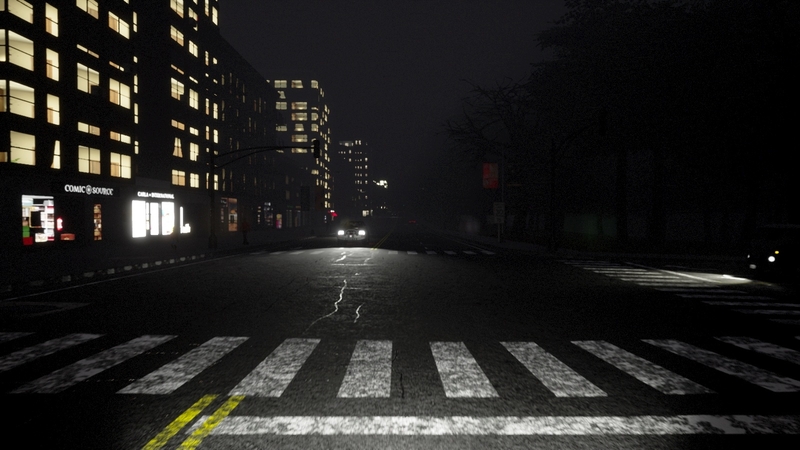}
    \includegraphics[width=0.142857\linewidth]{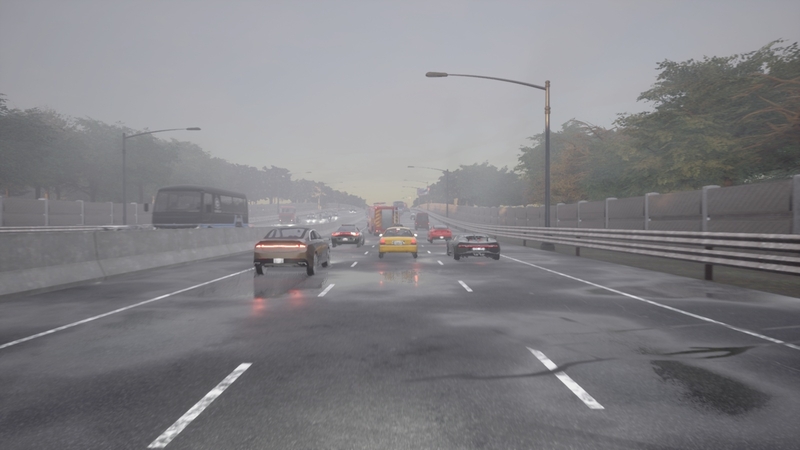}\hspace{-2.5pt}
    \includegraphics[width=0.142857\linewidth]{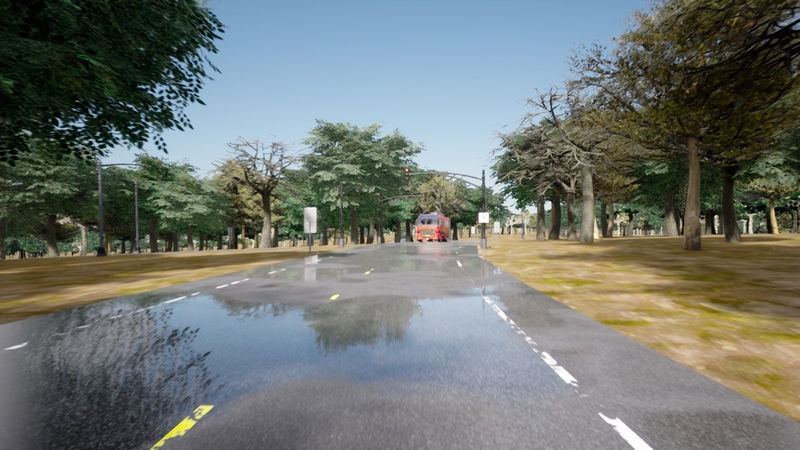}\hspace{-2.5pt}
    \includegraphics[width=0.142857\linewidth]{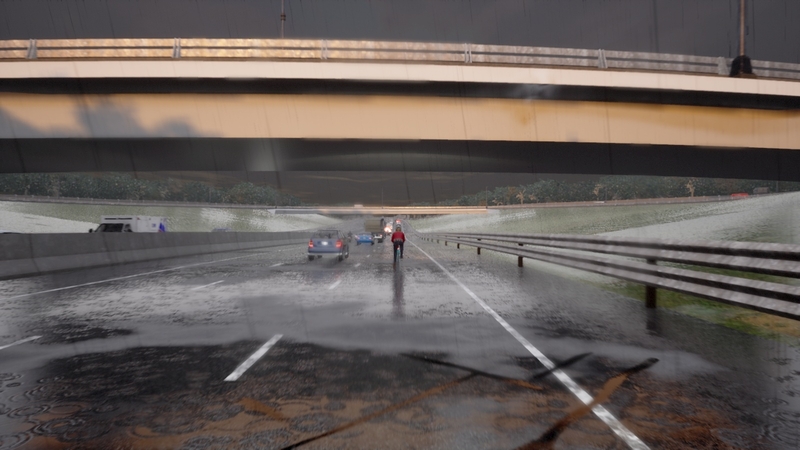}\hspace{-2.5pt}
    \includegraphics[width=0.142857\linewidth]{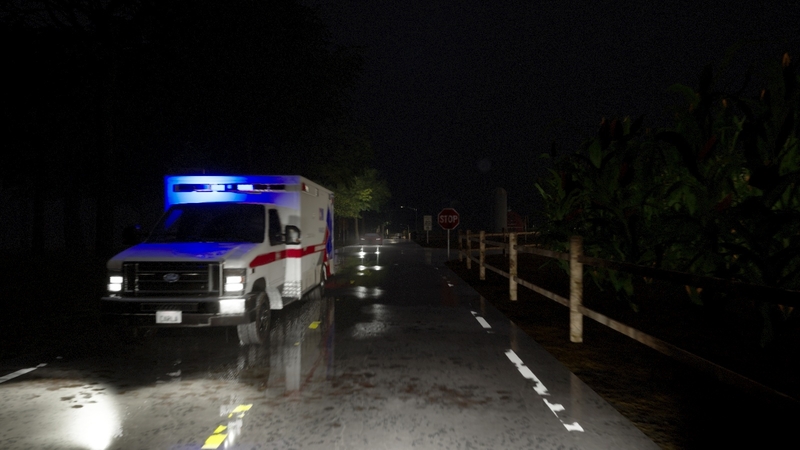}\hspace{-2.5pt}
    \includegraphics[width=0.142857\linewidth]{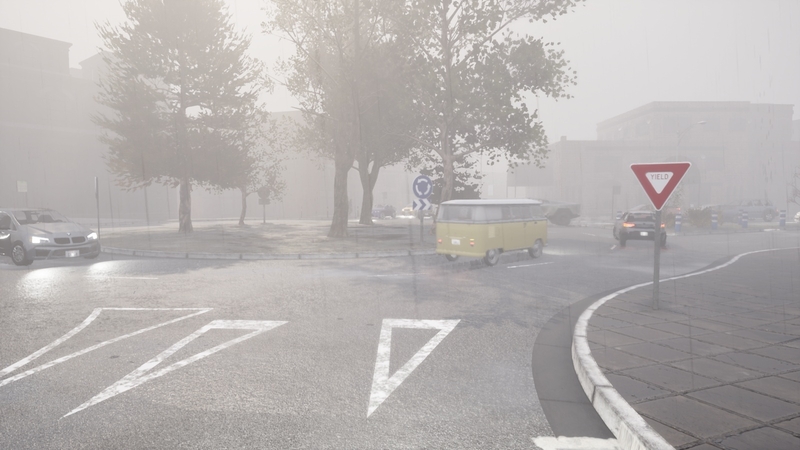}\hspace{-2.5pt}
    \includegraphics[width=0.142857\linewidth]{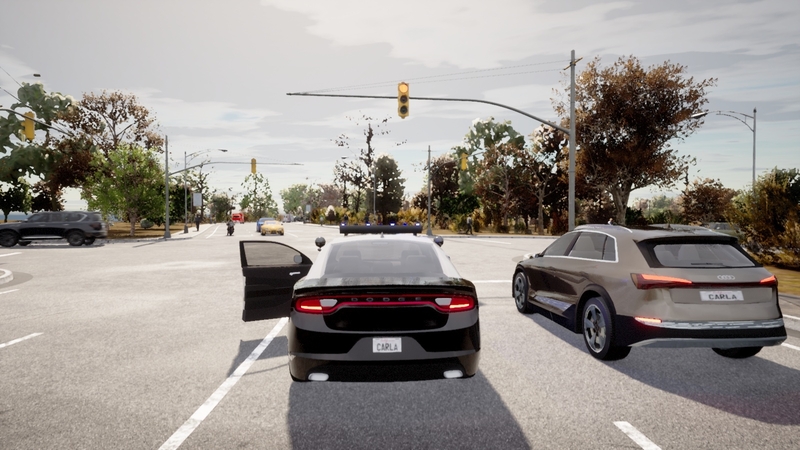}\hspace{-2.5pt}
    \includegraphics[width=0.142857\linewidth]{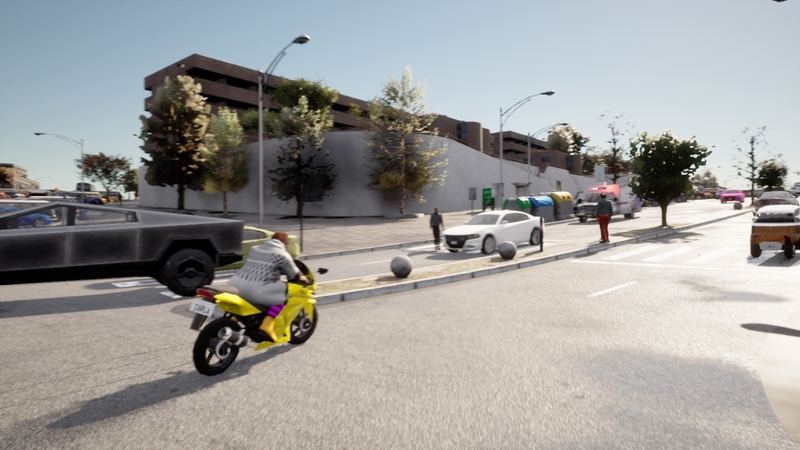}
    \includegraphics[width=0.142857\linewidth]{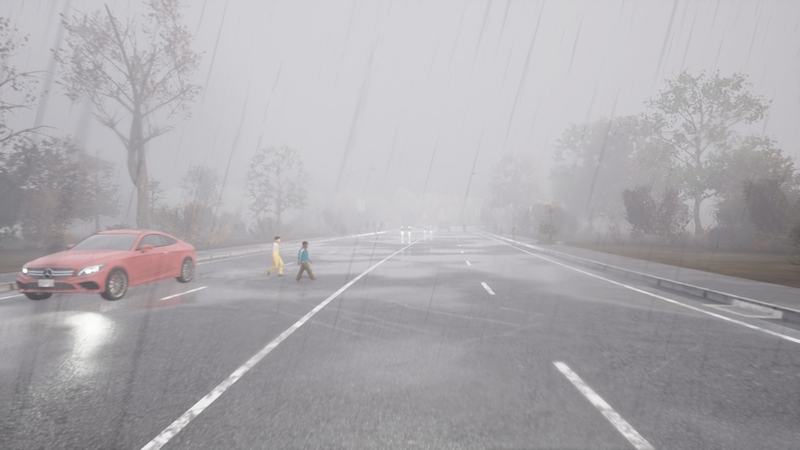}\hspace{-2.5pt}
    \includegraphics[width=0.142857\linewidth]{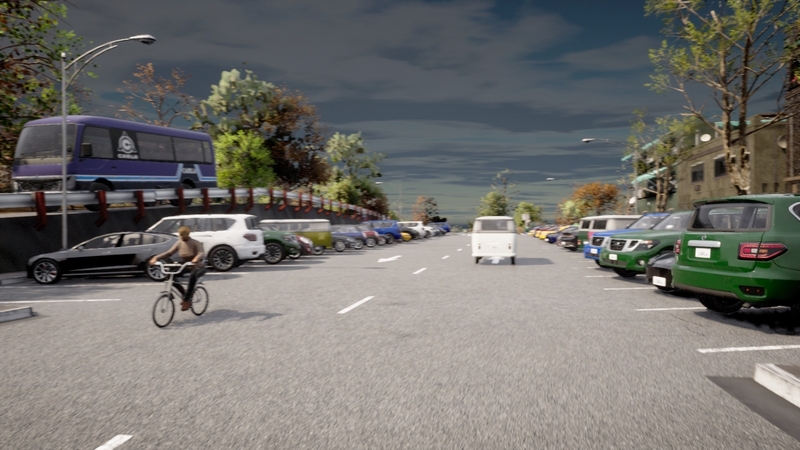}\hspace{-2.5pt}
    \includegraphics[width=0.142857\linewidth]{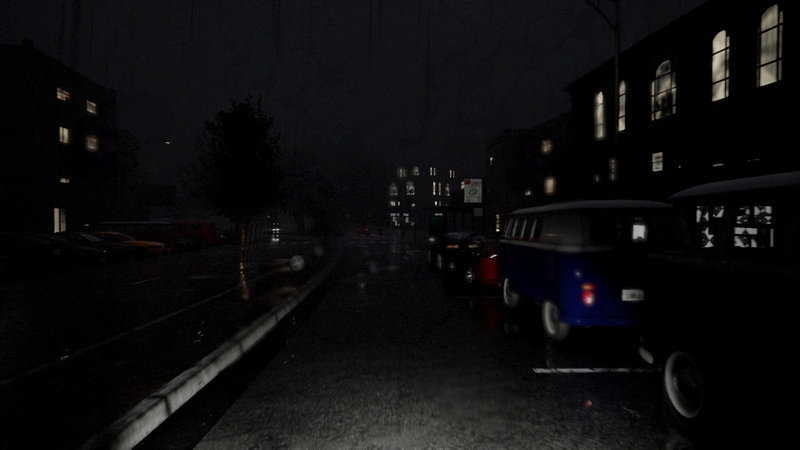}\hspace{-2.5pt}
    \includegraphics[width=0.142857\linewidth]{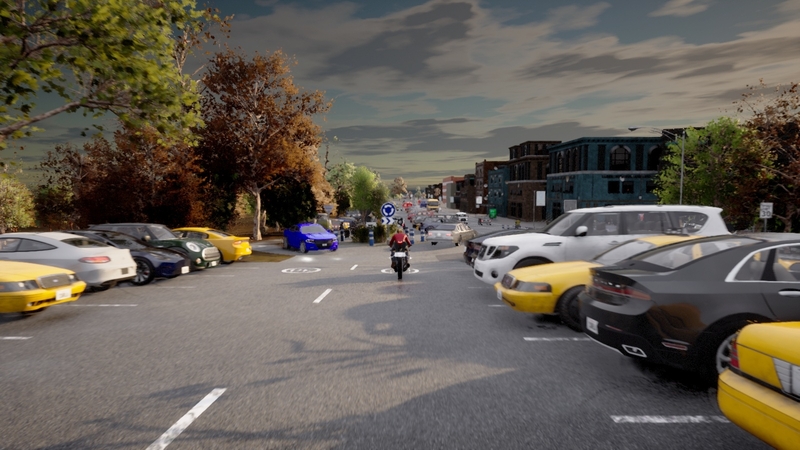}\hspace{-2.5pt}
    \includegraphics[width=0.142857\linewidth]{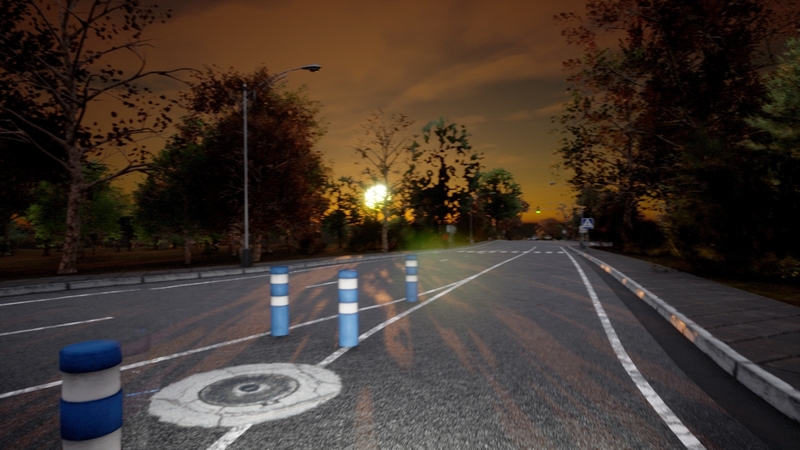}\hspace{-2.5pt}
    \includegraphics[width=0.142857\linewidth]{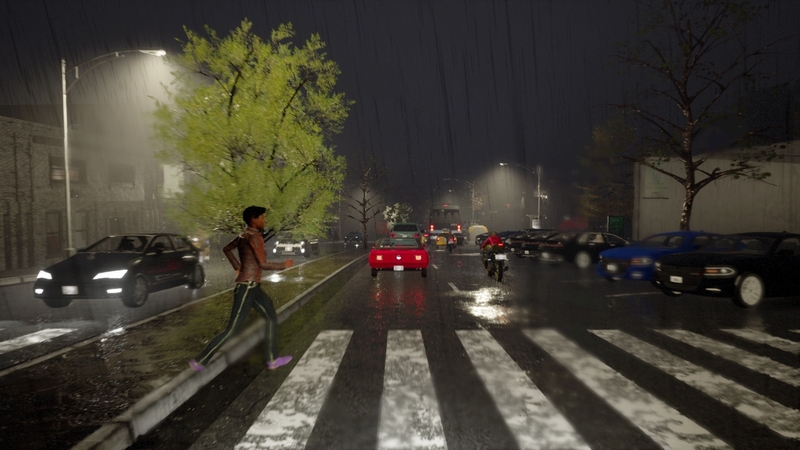}\hspace{-2.5pt}
    \includegraphics[width=0.142857\linewidth]{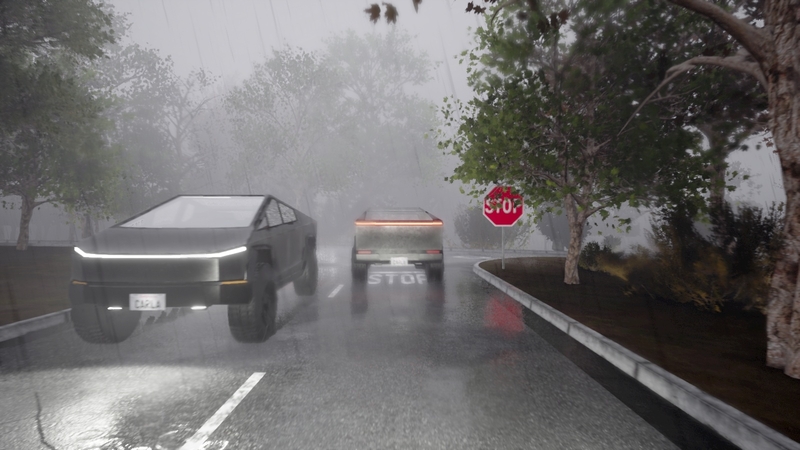}
    \setlength{\abovecaptionskip}{-10 pt}
    \setlength{\belowcaptionskip}{-14 pt}
    \caption{Front camera image samples from the SimBEV dataset.}\label{fig:image-samples}
\end{figure*}

The SimBEV dataset was created on a computer with a single Nvidia GeForce RTX 3090 graphics card over the course of 83 hours. It takes up 1.5 TB when compressed, and contains data from every sensor supported by SimBEV. Data is collected at a 20 Hz sample rate, with each scene lasting 16 s (320 frames). In total, the SimBEV dataset contains 102,400 annotated frames, 8,315,935 3D bounding boxes (3,792,499 of which are \textit{valid}), and 2,793,491,357 BEV ground truth labels. With 81.2 3D bounding boxes per frame (37.0 \textit{valid} bounding boxes per frame) it is on par with, if not surpassing, existing driving datasets \cite{caesar2020nuscenes, chang2019argoverse, wilson2023argoverse, weng2023all, sun2020scalability}. A collection of front camera images displayed in \cref{fig:image-samples} highlights the diversity of the SimBEV dataset.

More information about the SimBEV dataset, including sensor properties, SimBEV parameters, and dataset statistics, can be found in the Supplementary Material.
    \section{Evaluation and Analysis} \label{sec:eval}

The SimBEV dataset can be used for a variety of perception tasks, including 2D/3D segmentation, depth and optical flow estimation, and motion tracking and prediction. Here, we focus on BEV segmentation and 3D object detection.

\subsection{Tasks and metrics} \label{subsec:eval-metrics}

BEV segmentation results are evaluated using intersection over union (IoU), where for each class, a prediction is considered positive if its probability (score) is above a certain threshold (here 0.5). Our 3D object detection metrics are inspired by \cite{caesar2020nuscenes}, with average precision (AP) as the main metric. We consider two approaches to matching a predicted bounding box with a ground truth one. In the first, two boxes are matched if their 3D IoU is above a certain threshold \cite{geiger2013vision, cordts2016cityscapes}. In the second, two boxes are matched if the distance between their centers is below a certain threshold. As \cite{caesar2020nuscenes} notes, in the former, small translation errors for small objects (such as pedestrians) result in low or even zero IoU, making performance comparison of camera-only models that tend to have large localization errors difficult. More information about the metrics used for 3D object detection evaluation is available in the Supplementary Material.

\subsection{Evaluation results} \label{subsec:eval-results}

\begin{table*}[t]
    \centering
    \footnotesize
    \begin{tabular}{c c c c c c c c c c c}
        \toprule
        \textbf{Model} & \textbf{Modality} & \textbf{Road} & \textbf{Car} & \textbf{Truck} & \textbf{Bus} & \textbf{Motorcycle} & \textbf{Bicycle} & \textbf{Rider} & \textbf{Pedestrian} & \textbf{Mean} \\
        \toprule
        BEVFusion-C & C & 76.0 & 17.2 & 5.1 & 22.9 & 0.0 & 0.0 & 0.0 & 0.0 & 15.2 \\
        BEVFusion-L & L & 87.7 & 70.6 & \gb{73.5} & \rb{81.5} & 32.5 & \bb{3.6} & 18.4 & \bb{18.9} & \bb{48.3} \\
        BEVFusion & C + L & \bb{88.4} & \bb{72.7} & \rb{74.5} & \gb{80.0} & \gb{36.3} & \bb{3.6} & \bb{23.3} & \gb{20.0} & \rb{50.0} \\
        UniTR & C + L & \gb{92.8} & \rb{73.8} & 67.7 & 51.7 & \rb{36.5} & \rb{11.4} & \rb{36.2} & \rb{27.5} & \gb{49.7} \\
        UniTR+LSS & C + L & \rb{93.3} & \gb{72.8} & \bb{69.4} & \bb{58.5} & \bb{35.9} & \gb{6.3} & \gb{31.6} & 12.9 & 47.6 \\
        \bottomrule
    \end{tabular}
    \setlength{\abovecaptionskip}{4 pt}
    \setlength{\belowcaptionskip}{-4 pt}
    \caption{BEV segmentation IoUs (in \%) for different models evaluated on the SimBEV dataset \textit{test} set. The top three values are indicated in \rb{red}, \gb{green}, and \bb{blue}, respectively.} \label{table:seg-results}
\end{table*}
\begin{table*}[t]
    \centering
    \footnotesize
    \begin{tabular}{c c c c c c c c}
        \toprule
        \multirow{2}*{\textbf{Model}} & \multirow{2}*{\textbf{Modality}} & \textbf{mAP} & \textbf{mATE} & \textbf{mAOE} & \textbf{mASE} & \textbf{mAVE} & \textbf{SDS} \\
         & & \textbf{(\%) $\uparrow$} & \textbf{(m) $\downarrow$} & \textbf{(rad) $\downarrow$} & \textbf{$\downarrow$} & \textbf{(m/s) $\downarrow$} & \textbf{(\%) $\uparrow$} \\
        \toprule
        BEVFusion-C & C & 7.0 & 0.337 & 0.943 & 0.106 & 4.98 & 23.7 \\
        BEVFusion-L & L & 33.9 & \bb{0.105} & \gb{0.086} & 0.107 & 1.49 & 50.8 \\
        BEVFusion & C + L & \gb{34.1} & 0.107 & \rb{0.077} & \bb{0.101} & \bb{1.46} & \bb{51.0} \\
        UniTR & C + L & \bb{33.0} & \rb{0.081} & 0.140 & \gb{0.071} & \gb{0.51} & \gb{56.5} \\
        UniTR+LSS & C + L & \rb{34.2} & \gb{0.083} & \bb{0.131} & \rb{0.069} & \rb{0.49} & \rb{57.5} \\
        \bottomrule
    \end{tabular}
    \setlength{\abovecaptionskip}{4 pt}
    \setlength{\belowcaptionskip}{-4 pt}
    \caption{3D object detection results for different models evaluated on the SimBEV dataset \textit{test} set using the first (IoU-based) method. The top three values are indicated in \rb{red}, \gb{green}, and \bb{blue}, respectively.} \label{table:det-results-iou}
\end{table*}
\begin{table*}[ht!]
    \centering
    \footnotesize
    \begin{tabular}{c c c c c c c c}
        \toprule
        \multirow{2}*{\textbf{Model}} & \multirow{2}*{\textbf{Modality}} & \textbf{mAP} & \textbf{mATE} & \textbf{mAOE} & \textbf{mASE} & \textbf{mAVE} & \textbf{SDS} \\
         & & \textbf{(\%) $\uparrow$} & \textbf{(m) $\downarrow$} & \textbf{(rad) $\downarrow$} & \textbf{$\downarrow$} & \textbf{(m/s) $\downarrow$} & \textbf{(\%) $\uparrow$} \\
        \toprule
        BEVFusion-C & C & 22.1 & 0.744 & 1.044 & 0.137 & 4.65 & 25.1 \\
        BEVFusion-L & L & \rb{48.1} & \gb{0.144} & \gb{0.133} & 0.134 & 1.56 & 56.4 \\
        BEVFusion & C + L & \rb{48.1} & \bb{0.146} & \rb{0.122} & \bb{0.127} & \bb{1.54} & \bb{56.6} \\
        UniTR & C + L & \bb{47.7} & \rb{0.113} & 0.224 & \gb{0.090} & \gb{0.55} & \gb{61.7} \\
        UniTR+LSS & C + L & \gb{47.8} & \rb{0.113} & \bb{0.207} & \rb{0.085} & \rb{0.53} & \rb{62.2} \\
        \bottomrule
    \end{tabular}
    \setlength{\abovecaptionskip}{4 pt}
    \setlength{\belowcaptionskip}{-14 pt}
    \caption{3D object detection results for different models evaluated on the SimBEV dataset \textit{test} set using the second (distance-based) method. The top three values are indicated in \rb{red}, \gb{green}, and \bb{blue}, respectively.} \label{table:det-results-distance}
\end{table*}

We benchmark BEVFusion \cite{liu2022bevfusion} and UniTR \cite{wang2023unitr} - both multi-sensor models for multi-task perception - on the SimBEV dataset. BEVFusion has camera-only (BEVFusion-C), lidar-only (BEVFusion-L), and fused (camera + lidar) variants for each task (six variants in total), allowing us to compare the performance of different modalities. BEVFusion-C is a variant of BEVDet-Tiny \cite{huang2021bevdet} using a much heavier view transformer, while BEVFusion-L is the lidar-only variant of TransFusion (TransFusion-L) \cite{bai2022transfusion}. UniTR, along with the base model for each task, has a variant augmented by an additional LSS-based BEV fusion step (four variants in total) \cite{liu2022bevfusion, liang2022bevfusion, philion2020lift}.

\Cref{table:seg-results} shows BEV segmentation IoUs (in \%) for different models evaluated on the SimBEV dataset \textit{test} set. As expected, all models achieve higher IoUs for larger objects compared to smaller ones (\textit{motorcycle}, \textit{bicycle}, \textit{rider}, and \textit{pedestrian}). In addition, the IoUs for the \textit{road} class (which is the only BEV segmentation class shared between SimBEV and nuScenes) are consistent with \cite{liu2022bevfusion}.

\Cref{table:seg-results} shows that BEVFusion outperforms BEVFusion-L only by a small margin, probably because of SimBEV's dense lidar point cloud. Notably, BEVFusion gets ahead when it comes to detecting smaller objects, probably because of the extra semantic information obtained from camera images. However, both models perform poorly when it comes to the \textit{bicycle} class, though we found that BEVFusion achieves a 12.7\% IoU for that class when the threshold is lowered to 0.4. It seems that because bicycles are always accompanied by a rider (and are smaller than motorcycles), the model has difficulty distinguishing between the two and has lower confidence in its predictions.

\Cref{table:seg-results} also shows that BEVFusion outperforms UniTR, because the latter significantly underperforms in the \textit{bus} class, even though it is much better at detecting smaller objects than the former. We think that this is likely because UniTR's transformer backbone is unable to effectively utilize information in the $z$ direction. We can also see from \cref{table:seg-results} that BEVFusion-C performs poorly (except for the \textit{road} class) compared to the others. As noted above, because images lack explicit geometric information, camera-only models have difficulty localizing objects.

3D object detection results using the first and second methods are shown in \cref{table:det-results-iou} and \cref{table:det-results-distance}, respectively. In contrast to nuScenes benchmarks \cite{caesar2020nuscenes, wang2023unitr}, BEVFusion slightly outperforms UniTR and UniTR+LSS in mAP here. However, the UniTR variants score a much higher SDS because they do a much better job at predicting object velocities. We can also see that the second matching method (distance-based) produces higher mAP values. This is due to its more permissive nature, where, unlike the first matching method, two boxes can be matched even if they do not intersect at all. This permissiveness, which makes BEVFusion-C more comparable to the others, can be seen when juxtaposing the mATE, mAOE, and mASE values of the two methods, with those for the second method considerably higher. A breakdown of the results by class is available in the Supplementary Material.

Finally, we should note that while CARLA and the real world are statistically different domains, our results and those of \cite{sun2022shift} indicate that trends in CARLA are compatible with real-world observations and SimBEV, with its accurate BEV ground truth, can be a useful tool for evaluating both novel perception methods and domain adaptation strategies.

    \section{Conclusion} \label{sec:conclusion}

In this paper, we introduced SimBEV, a randomized synthetic data generation tool that is extensively configurable and scalable, supports a wide array of sensors, incorporates information from multiple sources to capture accurate BEV ground truth, and enables a variety of perception tasks including BEV segmentation and 3D object detection. To showcase SimBEV, we used it to create the SimBEV dataset, a comprehensive large-scale driving dataset, which we used to benchmark BEV perception models and compare different sensing modalities. We hope that SimBEV empowers researchers in exploring a variety of computer vision tasks. Future work will focus on improving SimBEV and enabling vehicle-to-everything (V2X) data collection.

\newpage
    
    % {
    %     \small
    %     \bibliographystyle{ieeenat_fullname}
    %     \bibliography{main}
    % }
    
    % WARNING: do not forget to delete the supplementary pages from your submission 
    \clearpage

\renewcommand\thesection{\Alph{section}}

\setcounter{page}{1}
\setcounter{section}{0}
\maketitlesupplementary

A preview of SimBEV can be accessed at \href{https://gitfront.io/r/SportCarGallery/yY1YEo7uEcLB/SimBEV-Preview/}{https://gitfront .io/r/SportCarGallery/yY1YEo7uEcLB/SimBEV-Preview/}.

\section{CARLA Simulator}
\label{appsec:carla}

\begin{figure}[t]
    \centering
    \setlength{\lineskip}{0pt}
    \includegraphics[width=\linewidth]{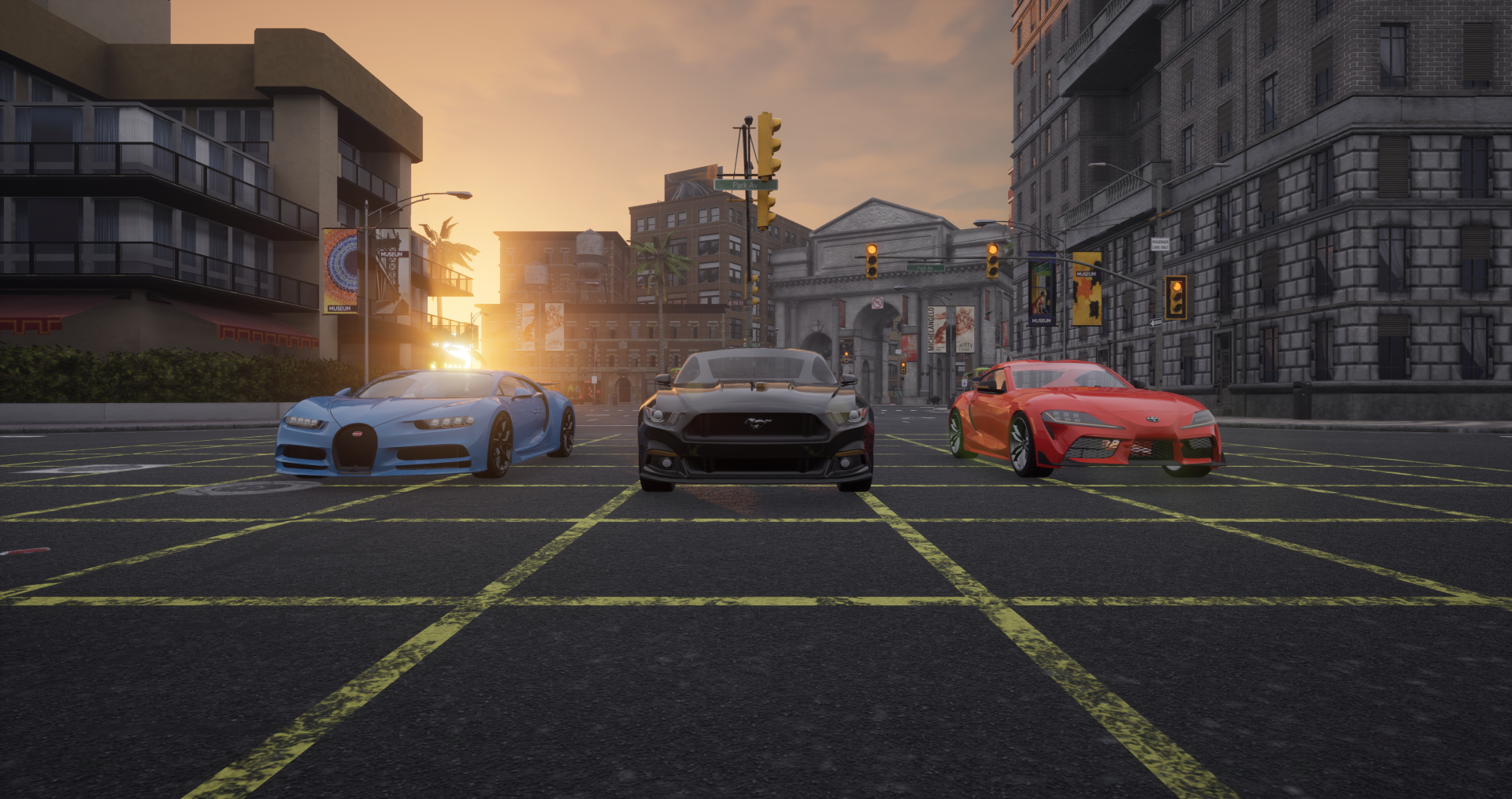}
    \includegraphics[width=\linewidth]{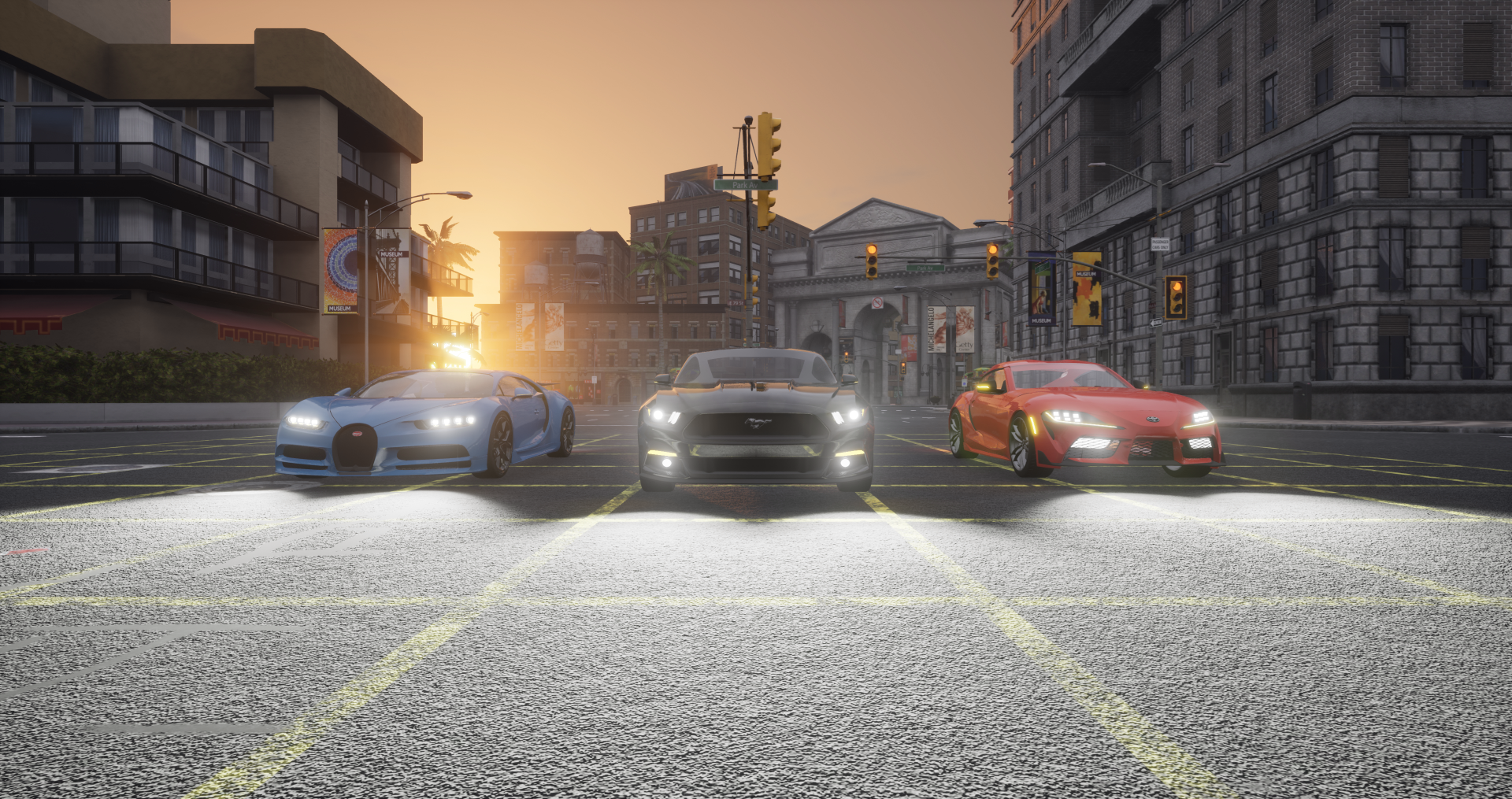}
    \caption{From left to right, the Bugatti Chiron, Ford Mustang, and Toyota GR Supra added to CARLA's vehicle library with their lights turned off (top) and on (bottom).}\label{appfig:car-trio}
\end{figure}

\begin{figure}[t]
    \centering
    \setlength{\belowcaptionskip}{-6 pt}
    \includegraphics[width=\linewidth]{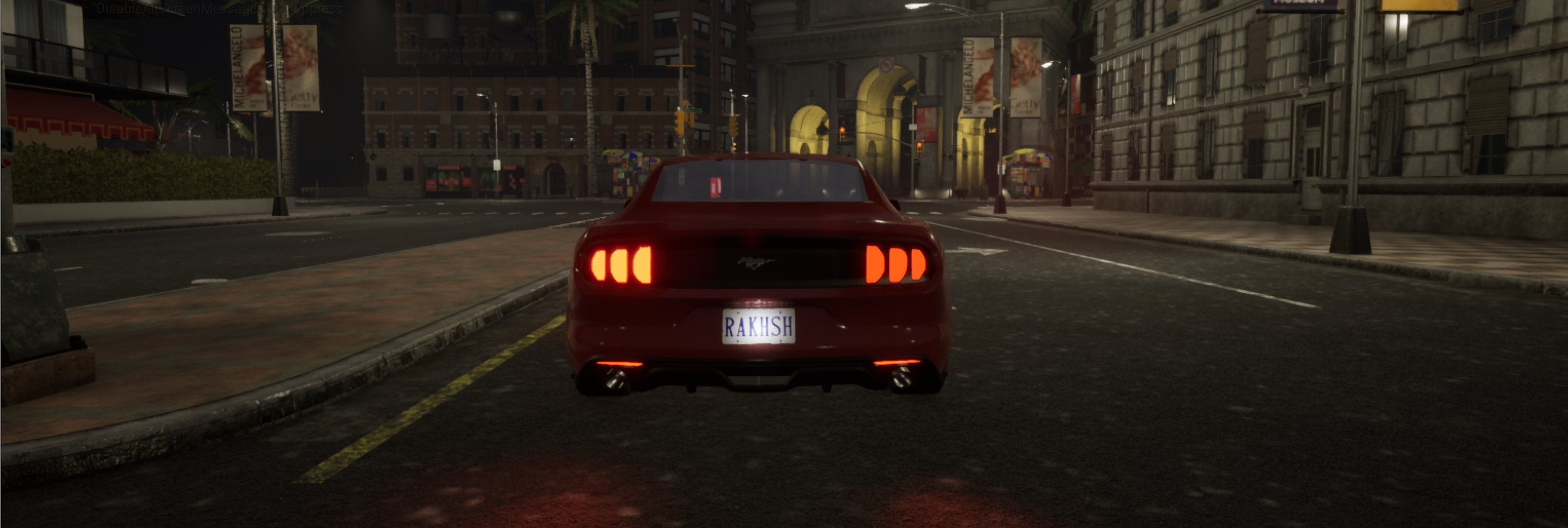}
    \caption{In contrast to CARLA's segmented light design approach, our multi-layer approach can realistically visualize vehicle lights that use the same element for multiple purposes. For instance, in the Ford Mustang pictured here both position and left blinker lights are turned on.}\label{appfig:mustang-lights}
\end{figure}

SimBEV relies on CARLA Simulator 0.9.15 \cite{dosovitskiy2017carla} equipped with an enhanced content library. Some of the improvements we made are listed below.
\begin{itemize}
    \item We added three new sports cars to CARLA's vehicle library using existing 3D models \cite{kentik2024mustang}\footnote{We used royalty-free 3D models of the three cars available on BlenderKit as the basis for the vehicles. However, the Supra and Chiron models had been removed from BlenderKit at the time of writing, so unfortunately we have no way of crediting their authors for their work.}: sixth generation Ford Mustang, Toyota GR Supra, and Bugatti Chiron, shown in \cref{appfig:car-trio}. They enhance the diversity of CARLA's vehicle library, especially when it comes to fast, high-performance cars. The Ford Mustang is the default data collection vehicle in SimBEV.
    \item We added lights (headlights, taillights, blinkers, etc.) to some of the older models in CARLA's vehicle library that lacked them, and redesigned existing vehicle lights in Blender using a new multi-layer approach that better visualizes modern multi-purpose lights, as shown in \cref{appfig:mustang-lights}.
    \item We added a set of 160 standard colors available to most vehicle models (apart from a few like the firetruck), and fixed color randomization issues for a few vehicles.
    \item We updated vehicle dynamics parameters of vehicle models to better match their vehicle's behavior and performance in the real world.
    \item We added or updated pedestrian navigation information for CARLA's Town12, Town13, and Town15 maps.
    \item We updated motorcycle and bicycle models so that they select their driver models randomly each time, instead of always being assigned the same model.
    \item We added lights to buildings in Town12 and fixed issues that prevented full control over building/street lights in Town12 and Town15.
\end{itemize}

SimBEV is compatible with the standard version of CARLA 0.9.15, but some features may not work properly.


\section{The SimBEV Dataset} \label{appsec:simbev-dataset-app}

\subsection{SimBEV configuration} \label{appsubsec:simbev-params}

\begin{figure}[t]
    \centering
    \includegraphics[width=\linewidth]{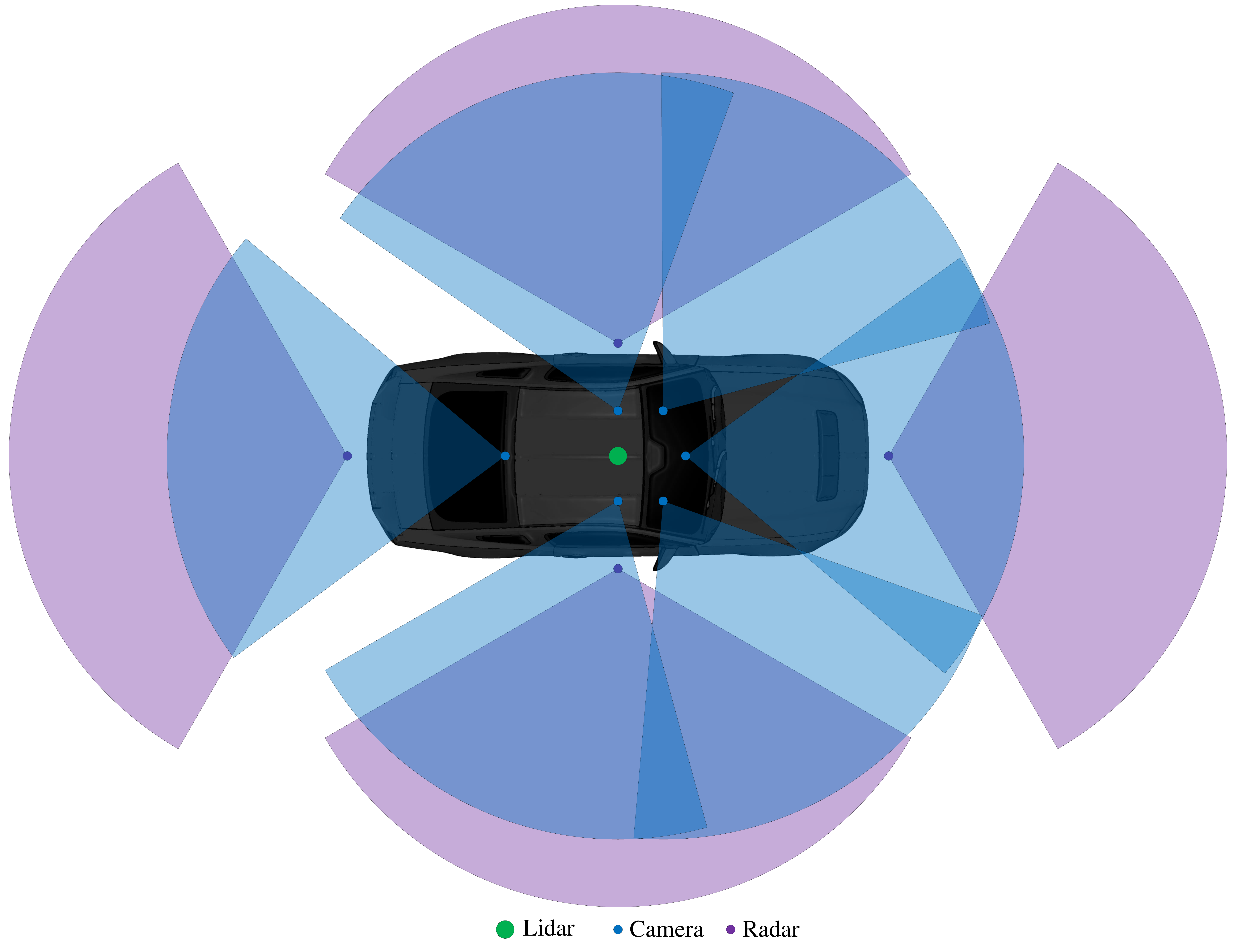}
    \caption{Position and FoV of the perception sensors used in SimBEV to create the SimBEV dataset.}\label{appfig:sensor-fov}
\end{figure}

\begin{figure}[t]
    \centering
    \includegraphics[width=\linewidth]{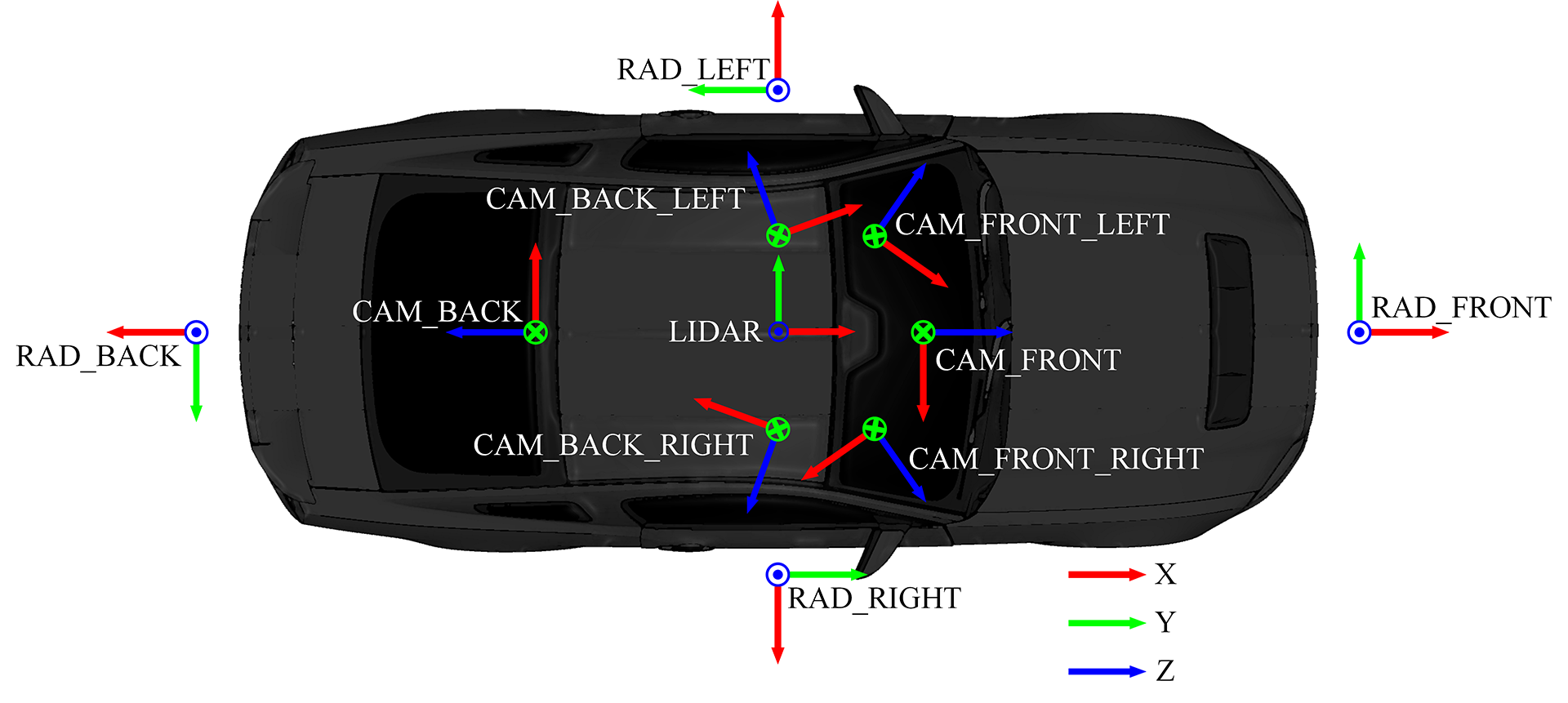}
    \caption{Coordinate frames of the perception sensors used in SimBEV to create the SimBEV dataset.}\label{appfig:sensor-coord}
\end{figure}

We configured SimBEV to generate a diverse set of unique scenarios for the SimBEV dataset, and collected data from all sensor types supported by SimBEV (RGB, semantic segmentation, instance segmentation, depth, and optical flow cameras; regular and semantic lidar; radar; GNSS; and IMU). Sensor configurations are listed in \cref{apptable:sensor-properties} and the arrangement of the sensors on the ego vehicle is shown in \cref{appfig:sensor-fov} and \cref{appfig:sensor-coord}, and detailed in \cref{apptable:sensor-positions}.

\begin{table*}[t]
    \centering
    \footnotesize
    \begin{tabular}{l l}
        \toprule
        \textbf{Sensor type} & \textbf{Properties} \\
        \toprule
        RGB camera & 1600$\times$900 resolution, 80 deg FoV, $f/1.8$ \\
        All other cameras & 1600$\times$900 resolution, 80 deg FoV \\
        \multirow{2}*{Lidar} & 128 channels, 120.0 m range, 20.0 Hz rotation frequency, 5,242,880 points per second, -30.67 to 10.67 vertical FoV, \\
         & 14\% general drop-off rate, 1 cm radial noise std \\
        Semantic lidar & 128 channels, 120.0 m range, 20.0 Hz rotation frequency, 5,242,880 points per second, -30.67 to 10.67 vertical FoV \\
        Radar & 120.0 m range, 100 deg horizontal FoV, 12 deg vertical FoV, 40,000 points per second \\
        GNSS & \{4e-2 m, 4e-7 deg, 4e-7 deg\} noise std for \{altitude, latitude, longitude\} \\
        IMU & 1.7e-4 rad/s gyroscope bias, \{1.7e-4 m/s$^ {2}$, 5.6e-6 rad/s\} noise std for \{accelerometer, gyroscope\} \\
        \bottomrule
    \end{tabular}
    \caption{Sensor configurations used for the collection of the SimBEV dataset. std: standard deviation.} \label{apptable:sensor-properties}
\end{table*}
\begin{table}[t]
    \centering
    \footnotesize
    \setlength{\belowcaptionskip}{8 pt}
    \begin{tabular}{l c c c c}
        \toprule
        \textbf{Sensor} & \textbf{$x$ (m)} & \textbf{$y$ (m)}& \textbf{$z$ (m)} & \textbf{$\gamma$ (deg)} \\
        \toprule
        Front left camera & 0.4 & 0.4 & 1.6 & 55 \\
        Front camera & 0.6 & 0.0 & 1.6 & 0\\
        Front right camera & 0.4 & -0.4 & 1.6 & -55 \\
        Back left camera & 0.0 & 0.4 & 1.6 & 110 \\
        Back camera & -1.0 & 0.0 & 1.6 & 180 \\
        Back right camera & 0.0 & -0.4 & 1.6 & -110 \\
        Left radar & 0.0 & 1.0 & 0.6 & 90 \\
        Front radar & 2.4 & 0.0 & 0.6 & 0 \\
        Right radar & 0.0 & -1.0 & 0.6 & -90 \\
        Back radar & -2.4 & 0.0 & 0.6 & 180 \\
        Lidar & 0.0 & 0.0 & 1.8 & N/A \\
        \bottomrule
    \end{tabular}
    \caption{Arrangement of data collection sensors used in SimBEV and the SimBEV dataset. Coordinates are relative to the center of the ground plane of the ego vehicle's 3D bounding box.} \label{apptable:sensor-positions}
\end{table}

Our sensor setup was mostly inspired by \cite{caesar2020nuscenes} (e.g. the 1600$\times$900 image resolution, the arrangement of the cameras, and the lidar's vertical FoV), though there are a few differences. Our lidars (both regular and semantic) have 128 channels instead of 32 to collect a much denser point cloud, which can be downsampled by the user later on if desired. For GNSS and IMU, we used the bias and noise standard deviation values of a GNSS/INS module found in a typical experimental autonomous driving platform.

\begin{table}[t]
    \centering
    \footnotesize
    \setlength{\belowcaptionskip}{14 pt}
    \setlength{\tabcolsep}{4 pt}
    \begin{tabular}{l c l}
        \toprule
        \textbf{Parameter} & \textbf{Symbol} & \textbf{Distribution} \\
        \toprule
        Cloudiness & $k_{c}$ & $100\times\mathcal{B}(0.8, 1.0)$ \\
        Precipitation & $k_{p}$ & $\mathcal{B}(0.8, 0.2)\times k_{c}$ if $k_{c} > 40.0$ else 0.0 \\
        Precipitation & \multirow{2}*{$k_{pd}$} & \multirow{2}*{$k_{p} + \mathcal{B}(1.2, 1.6)\times(100 - k_{p})$} \\
        deposits \\
        Wetness & $k_{w}$ & $\min(100.0, \max(\mathcal{N}(k_{p}, 10.0)))$ \\
        Wind intensity & $k_{wi}$ & $\mathcal{U}(0.0, 100.0)$ \\
        Sun azimuth & \multirow{2}*{$k_{az}$} & \multirow{2}*{$\mathcal{U}(0.0, 360.0)$} \\
        angle \\
        Sun altitude & \multirow{2}*{$k_{al}$} & \multirow{2}*{$180\times\mathcal{B}(3.6, 2.0) - 90.0$} \\
        angle \\
        \multirow{2}*{Fog density} & \multirow{2}*{$k_{f}$} & $100\times\mathcal{B}(1.6, 2.0)$ if $k_{c} > 40.0$ \\
         &  & or $k_{al} < 10.0$ else 0.0 \\
        Fog distance & $k_{fd}$ & $\mathcal{LN}(3.2, 0.8)$ if $k_{f} > 10.0$ else 0.0 \\
        Fog falloff & $k_{ff}$ & $5.0\times\mathcal{B}(1.2, 2.4)$ if $k_{f} > 10.0$ else 1.0 \\
        \bottomrule
    \end{tabular}
    \caption{Probability distribution used in SimBEV for weather parameters. $\mathcal{B}$: beta distribution. $\mathcal{N}$: normal distribution. $\mathcal{U}$: uniform distribution. $\mathcal{LN}$: log-normal distribution.} \label{apptable:weather-distributions}
\end{table}

SimBEV uses the probability distributions listed in \cref{apptable:weather-distributions} to randomize the parameters that control the weather in CARLA. These distributions are interdependent to ensure that the resulting weather is realistic (e.g. a combination of heavy rain and clear sky is unrealistic). Each of the configured parameters is briefly discussed below.

\begin{table*}[t]
    \centering
    \footnotesize
    \begin{tabular}{l c}
        \toprule
        \textbf{Parameter} & \textbf{Value or distribution} \\
        \toprule
        Warmup duration & 4 s \\
        Scene duration & 16 s \\
        Simulation time step & 50 ms \\
        3D bounding box collection radius & 120.0 m \\
        BEV grid resolution & $360\times360$ \\
        BEV grid cell dimensions & 0.4 m $\times$ 0.4 m \\
        Distance between CARLA-generated waypoints used for BEV ground truth calculation & 0.4 m \\
        Distance between CARLA-generated waypoints used as vehicle spawn location & 24.0 m \\
        Number of background vehicles ($s$: number of available spawn locations) & $\mathcal{U}_{i}(0, s - 3)$ \\
        Number of pedestrians & $\mathcal{U}_{i}(0, 640)$ \\
        Radius around the ego vehicle where background vehicles and pedestrians are spawned & 400.0 m \\
        Probability of vehicle door(s) getting open when stopped & 10.0\% \\
        Probability of emergency lights turned on & 50.0\% \\
        Probability of ego vehicle being reckless & 1.0\% \\
        Probability of other vehicles being reckless & 1.0\% \\
        Minimum speed of pedestrians & 0.8 m/s \\
        Maximum speed of pedestrians ($r$: minimum pedestrian speed) & $\max(r, \mathcal{LN}(0.16, 0.64))$ m/s \\
        Minimum intensity of street lights & 10,000 lm \\
        Change in the intensity of street lights ($m$: average intensity of all street lights in the scene) & $\mathcal{U}(-m, m)$ lm \\
        Probability of street light failure & 10.0\% \\
        Maximum vehicle speed relative to the speed limit & $\mathcal{U}(-20.0, 40.0)\%$ \\
        Distance to front vehicle when stopped & $\mathcal{N}(3.2, 1.0)$ m \\
        Traffic light green time & $\mathcal{U}(4.0, 28.0)$ s \\
        Walker cross factor & $\mathcal{B}(2.4, 1.6)$ \\
        \bottomrule
    \end{tabular}
    \caption{SimBEV configuration used for the collection of the SimBEV dataset. $\mathcal{B}$: beta distribution. $\mathcal{N}$: normal distribution. $\mathcal{U}$: uniform distribution. $\mathcal{U}_{i}$: uniform integer distribution. $\mathcal{LN}$: log-normal distribution.} \label{apptable:parameters}
\end{table*}
\begin{table}[t]
    \centering
    \footnotesize
    \setlength{\tabcolsep}{4 pt}
    \begin{tabular}{l c c c}
        \toprule
        \multirow{2}*{\textbf{Class}} & \textbf{Total 3D} & \textbf{Valid 3D}& \multirow{2}*{\textbf{BEV labes}} \\
         & \textbf{bounding boxes} & \textbf{bounding boxes} &  \\
        \toprule
        Road & N/A & N/A & 2,674,391,899 \\
        Car & 2,935,809 & 1,495,066 & 84,073,215 \\
        Truck & 497,729 & 298,280 & 22,759,787 \\
        Bus & 67,880 & 46,754 & 7,546,007 \\
        Motorcycle & 297,132 & 146,083 & 858,136 \\
        Bicycle & 214,619 & 100,640 & 187,869 \\
        Rider & N/A & N/A & 510,521 \\
        Pedestrian & 4,302,766 & 1,705,676 & 3,163,923 \\
        \toprule
        \textbf{Total} & 8,315,935 & 3,792,499 & 2,793,491,357 \\
        \bottomrule
    \end{tabular}
    \caption{Breakdown of the number of total and \textit{valid} 3D bounding boxes and BEV ground truth labels by class for the SimBEV dataset.} \label{apptable:overall-stats}
\end{table}

\begin{figure}[t]
    \centering
    \begin{tikzpicture}
        \begin{axis}[title = {}, xlabel = {Precipitation}, xlabel near ticks, ylabel = {Scenes}, ylabel near ticks, xmin = 0.4, xmax = 4.6, ymin = 0, ymax = 180, ybar, xtick = {1, 2, 3, 4}, xticklabels = {none, low, moderate, heavy}, xticklabel style = {rotate = 90, anchor = east}, ytick = {0, 30, 60, 90, 120, 150, 180}, label style = {font = \footnotesize}, tick pos = left, tick label style = {font = \footnotesize}, ymajorgrids = true, grid style = dashed, width = 0.4\linewidth, height = 0.6\linewidth]
            \addplot[color = blue, fill = blue] coordinates {(1, 163) (2, 29) (3, 79) (4, 49)};
        \end{axis}
    \end{tikzpicture}
    \begin{tikzpicture}
        \begin{axis}[title = {}, xlabel = {Fog density}, xlabel near ticks, xmin = 0.4, xmax = 4.6, ymin = 0, ymax = 140, ybar, xtick = {1, 2, 3, 4}, xticklabels = {none, low, moderate, heavy}, xticklabel style = {rotate = 90, anchor = east}, ytick = {0, 20, 40, 60, 80, 100, 120, 140}, label style = {font = \footnotesize}, tick pos = left, tick label style = {font = \footnotesize}, ymajorgrids = true, grid style = dashed, width = 0.4\linewidth, height = 0.6\linewidth]
            \addplot[color = green, fill = green] coordinates {(1, 123) (2, 69) (3, 93) (4, 35)};
        \end{axis}
    \end{tikzpicture}
    \begin{tikzpicture}
        \begin{axis}[title = {}, xlabel = {Lighting}, xlabel near ticks, xlabel shift = {-4.8 pt}, xmin = 0.4, xmax = 3.6, ymin = 0, ymax = 240, ybar, xtick = {1, 2, 3}, xticklabels = {night, dawn/dusk, day}, xticklabel style = {rotate = 90, anchor = east}, ytick = {0, 40, 80, 120, 160, 200, 240}, label style = {font = \footnotesize}, tick pos = left, tick label style = {font = \footnotesize}, ymajorgrids = true, grid style = dashed, width = 0.4\linewidth, height = 0.6\linewidth]
            \addplot[color = red, fill = red] coordinates {(1, 72) (2, 19) (3, 229)};
        \end{axis}
    \end{tikzpicture}
    \caption{Distribution of weather across SimBEV dataset scenes.} \label{appfig:weather-dist}
\end{figure}
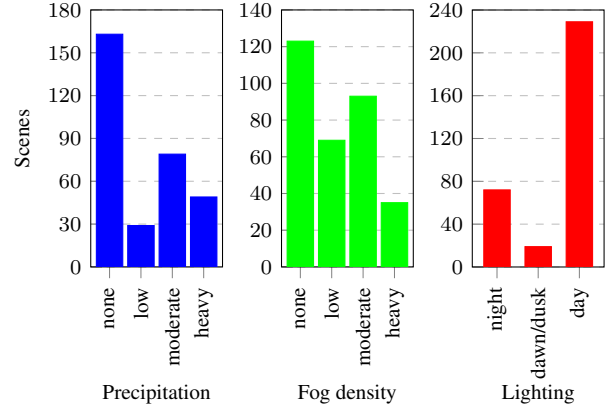

\begin{itemize}
    \item Cloudiness ($k_{c}$) controls the volume of clouds. Values range from 0 to 100.
    \item Precipitation ($k_{p}$) controls the intensity of rain. Values range from 0 to 100.
    \item Precipitation deposits ($k_{pd}$) controls the amount of puddles. Values range from 0 to 100, with 0 being no puddles and 100 a road filled with water.
    \item Wetness ($k_{w}$) controls the intensity of camera image blurriness caused by rain. Values range from 0 to 100.
    \item Wind intensity ($k_{wi}$) controls the strength of wind. Values range from 0 to 100.
    \item Sun azimuth angle ($k_{az}$) controls the azimuth angle of the sun. Values range from 0 to 360.
    \item Sun altitude angle ($k_{al}$) controls the altitude angle of the sun. Values range from -90 to 90, with -90 representing midnight and 90 midday.
    \item Fog density ($k_{f}$) controls fog concentration or thickness. Values range from 0 to 100, with 0 being no fog.
    \item Fog distance ($k_{fd}$) controls how far away the fog starts, and can be any nonnegative number.
    \item Fog falloff ($k_{ff}$) controls the density of the fog (as in specific mass), and can be any nonnegative number. If set to 0, the fog will be lighter than air and will cover the whole scene. If set to 1, the fog is approximately as dense as air. For values greater than 5 the fog will be so dense that it will be compressed to the ground level. Fog falloff is set to 0.01 in Town12, Town13, and Town15 due to their non-zero elevation.
\end{itemize}
SimBEV leaves other weather parameters (such as scattering intensity and dust storm) at their default value, though the user can change them if desired.

\Cref{apptable:parameters} lists several other SimBEV configurations used to create the SimBEV dataset. They control various aspects of SimBEV such as scene duration, number of spawned background vehicles and pedestrians, driving behavior, chance of reckless driving, etc.

\subsection{SimBEV dataset statistics} \label{appsubsec:stats}

The SimBEV dataset comprises 102,400 annotated frames, 8,315,935 3D object bounding boxes (3,792,499 of which are \textit{valid}), and 2,793,491,357 BEV ground truth labels, broken down by class in \cref{apptable:overall-stats}. Cars and pedestrians make up the largest share of 3D object bounding boxes, though those boxes include a large number of motorcycles and bicycles as well. This makes sense since the majority of models in CARLA's vehicle library are cars (compared to, e.g., only one bus model). BEV labels are dominated by the \textit{road} class, followed by \textit{car}, \textit{truck}, and \textit{bus} due to their larger footprint compared to the rest.

As discussed previously, SimBEV randomizes CARLA's weather parameters according to the distributions specified in \cref{apptable:weather-distributions}. \Cref{appfig:weather-dist} shows the distribution of weather across the SimBEV dataset, where precipitation (rain intensity, $k_{p}$) and fog density ($k_{f}$) values for each scene are categorized into none ($<$10\%), low (10 - 40\%), moderate (40 - 70\%), and heavy (70 - 100\%); while sun altitude angle ($k_{al}$) is categorized into night (-90 - 0 deg), dawn/dusk (0 - 6 deg), and day (6 - 90 deg). \Cref{appfig:weather-dist} shows that SimBEV contains a good mix of different weather conditions, with rain or fog present in about half of the scenes and nearly a quarter of the scenes occurring at night.

\begin{figure}[ht]
    \centering
    \setlength{\belowcaptionskip}{-4 pt}
    \begin{tikzpicture}
        \begin{axis}[title = {}, xlabel = {Count}, xlabel near ticks, ylabel = {Scenes}, ylabel near ticks, xmin = 0, xmax = 17, ymin = 0, ymax = 80, ybar = {0.4 pt}, bar width = {4.8 pt}, xtick = {1, 2, 3, 4, 5, 6, 7, 8, 9, 10, 11, 12, 13, 14, 15, 16}, xticklabels = {[0{,} 40), [40{,} 80), [80{,} 120), [120{,} 160), [160{,} 200), [200{,} 240), [240{,} 280), [280{,} 320), [320{,} 360), [360{,} 400), [400{,} 440), [440{,} 480), [480{,} 520), [520{,} 560), [560{,} 600), 600+}, xticklabel style = {rotate = 90, anchor = east}, ytick = {0, 10, 20, 30, 40, 50, 60, 70, 80}, label style = {font = \footnotesize}, tick pos = left, tick label style = {font = \footnotesize}, legend pos = north east, legend style = {font = \footnotesize}, ymajorgrids = true, grid style = dashed, width = 1.06\linewidth, height = 0.76\linewidth]
            \addplot[color = blue, fill = blue] coordinates {(1, 66) (2, 54) (3, 48) (4, 32) (5, 26) (6, 14) (7, 16) (8, 21) (9, 15) (10, 8) (11, 5) (12, 7) (13, 4) (14, 2) (15, 1) (16, 1)};
            \addplot[color = red, fill = red] coordinates {(1, 72) (2, 29) (3, 27) (4, 26) (5, 25) (6, 30) (7, 35) (8, 32) (9, 21) (10, 8) (11, 5) (12, 5) (13, 2) (14, 0) (15, 2) (16, 1)};
            \legend{Vehicle, Pedestrian}
        \end{axis}
    \end{tikzpicture}
    \caption{Distribution of the number of spawned vehicles (cars, trucks, buses, motorcycles, and bicycles) and pedestrians across the scenes of the SimBEV dataset.} \label{appfig:vehicle-ped-dist}
\end{figure}
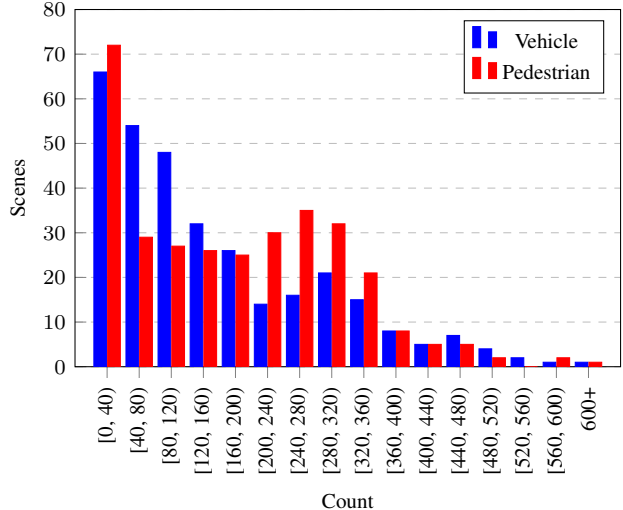

Looking at the distribution of the number of spawned vehicles (cars, trucks, buses, motorcycles, bicycles) and pedestrians across scenes of the SimBEV dataset, shown in \cref{appfig:vehicle-ped-dist}, it is clear that the scenes range from relatively empty to congested and crowded. The distribution of pedestrians is supposed to be uniform, but CARLA often spawns fewer pedestrians than requested, and the number of unspawned pedestrians grows rapidly when hundreds of pedestrians are requested. Moreover, in some cases CARLA cannot spawn pedestrians because there are no walkable areas around the ego vehicle (e.g., when the ego vehicle is traveling on a rural road). Hence, there are many scenes with 0 and 240 - 320 pedestrians and very few with more than 480.

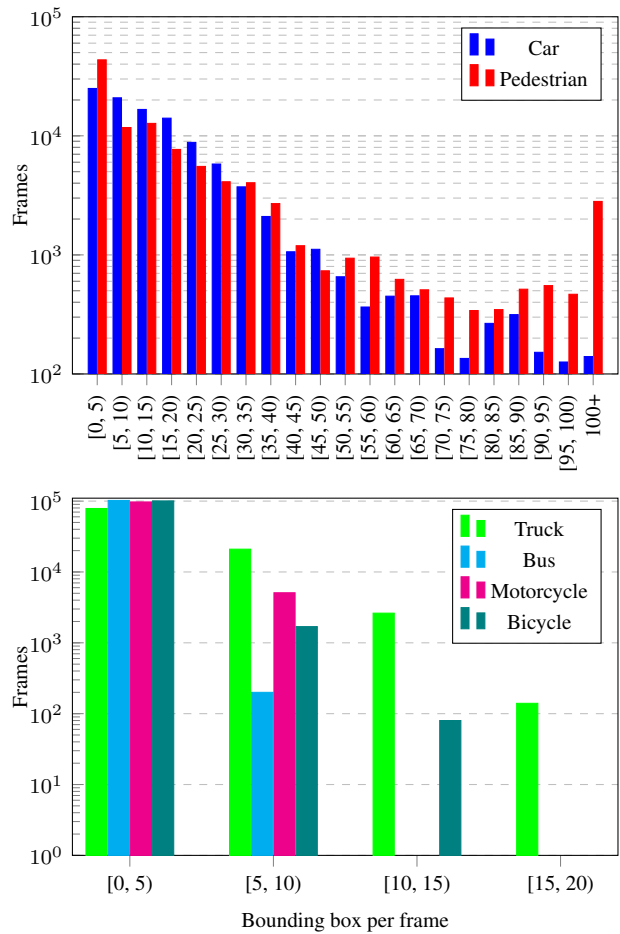
\begin{figure}[ht]
    \centering
    \setlength{\belowcaptionskip}{-24 pt}
    \begin{tikzpicture}
        \begin{semilogyaxis}[title = {}, xlabel near ticks, ylabel = {Frames}, ylabel near ticks, ylabel shift = {-6 pt}, xmin = 0, xmax = 22, ymin = 100, ymax = 100000, ybar = {0.4 pt}, bar width = {3.2 pt}, xtick = {1, 2, 3, 4, 5, 6, 7, 8, 9, 10, 11, 12, 13, 14, 15, 16, 17, 18, 19, 20, 21}, xticklabels = {[0{,} 5), [5{,} 10), [10{,} 15), [15{,} 20), [20{,} 25), [25{,} 30), [30{,} 35), [35{,} 40), [40{,} 45), [45{,} 50), [50{,} 55), [55{,} 60), [60{,} 65), [65{,} 70), [70{,} 75), [75{,} 80), [80{,} 85), [85{,} 90), [90{,} 95), [95{,} 100), 100+}, xticklabel style = {rotate = 90, anchor = east}, ytick = {100, 1000, 10000, 100000}, label style = {font = \footnotesize}, tick pos = left, tick label style = {font = \footnotesize}, legend pos = north east, legend style = {font = \footnotesize}, ymajorgrids = true, yminorgrids = true, grid style = dashed, width = 1.06\linewidth, height = 0.76\linewidth]
            \addplot[color = blue, fill = blue] coordinates {(1, 24985) (2, 20856) (3, 16656) (4, 14073) (5, 8809) (6, 5795) (7, 3728) (8, 2102) (9, 1062) (10, 1114) (11, 655) (12, 365) (13, 450) (14, 453) (15, 163) (16, 135) (17, 266) (18, 315) (19, 152) (20, 126) (21, 140)};
            \addplot[color = red, fill = red] coordinates {(1, 43457) (2, 11735) (3, 12716) (4, 7687) (5, 5534) (6, 4113) (7, 4035) (8, 2696) (9, 1194) (10, 735) (11, 938) (12, 959) (13, 623) (14, 509) (15, 435) (16, 341) (17, 347) (18, 515) (19, 552) (20, 467) (21, 2812)};
            \legend{Car, Pedestrian}
        \end{semilogyaxis}
    \end{tikzpicture}
    \begin{tikzpicture}
        \begin{semilogyaxis}[title = {}, xlabel = {Bounding box per frame}, xlabel near ticks, ylabel = {Frames}, ylabel near ticks, ylabel shift = {-6 pt}, xmin = 0.6, xmax = 4.4, ymin = 1, ymax = 110000, ybar = {0.4 pt}, bar width = {8 pt}, xtick = {1, 2, 3, 4}, xticklabels = {[0{,} 5), [5{,} 10), [10{,} 15), [15{,} 20)}, ytick = {1, 10, 100, 1000, 10000, 100000}, label style = {font = \footnotesize}, tick pos = left, tick label style = {font = \footnotesize}, legend pos = north east, legend style = {font = \footnotesize}, ymajorgrids = true, grid style = dashed, width = 1.06\linewidth, height = 0.76\linewidth]
            \addplot[color = green, fill = green] coordinates {(1, 78669) (2, 20969) (3, 2622) (4, 140)};
            \addplot[color = cyan, fill = cyan] coordinates {(1, 102200) (2, 200)};
            \addplot[color = magenta, fill = magenta] coordinates {(1, 97313) (2, 5087)};
            \addplot[color = teal, fill = teal] coordinates {(1, 100627) (2, 1693) (3, 80)};
            \legend{Truck, Bus, Motorcycle, Bicycle}
        \end{semilogyaxis}
    \end{tikzpicture}
    \caption{Breakdown of the number of \textit{valid} 3D object bounding boxes per frame by class across the SimBEV dataset.} \label{appfig:bbox-dist}
\end{figure}

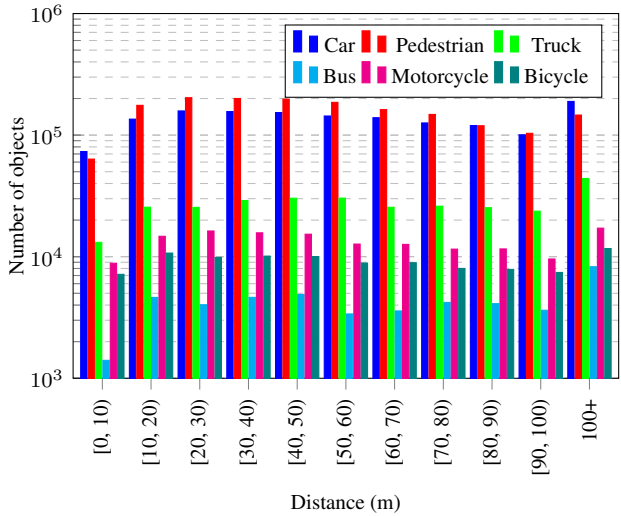
\begin{figure}[ht]
    \centering
    \setlength{\belowcaptionskip}{-7.2 pt}
    \begin{tikzpicture}
        \begin{semilogyaxis}[title = {}, xlabel = {Distance (m)}, xlabel near ticks, ylabel = {Number of objects}, ylabel near ticks, ylabel shift = {-6 pt}, xmin = 0.4, xmax = 11.6, ymin = 1000, ymax = 1000000, ybar = {0.4 pt}, bar width = {2.4 pt}, xtick = {1, 2, 3, 4, 5, 6, 7, 8, 9, 10, 11}, xticklabels = {[0{,} 10), [10{,} 20), [20{,} 30), [30{,} 40), [40{,} 50), [50{,} 60), [60{,} 70), [70{,} 80), [80{,} 90), [90{,} 100), 100+}, xticklabel style = {rotate = 90, anchor = east}, ytick = {1000, 10000, 100000, 1000000}, label style = {font = \footnotesize}, tick pos = left, tick label style = {font = \footnotesize}, legend pos = north east, legend style = {font = \footnotesize, legend columns = 3}, ymajorgrids = true, yminorgrids = true, grid style = dashed, width = 1.06\linewidth, height = 0.772\linewidth]
            \addplot[color = blue, fill = blue] coordinates {(1, 73296) (2, 135391) (3, 158103) (4, 156256) (5, 153405) (6, 143649) (7, 139089) (8, 125966) (9, 119928) (10, 100596) (11, 189387)};
            \addplot[color = red, fill = red] coordinates {(1, 63503) (2, 175807) (3, 203049) (4, 200182) (5, 197905) (6, 185989) (7, 162010) (8, 148013) (9, 119457) (10, 103498) (11, 146263)};
            \addplot[color = green, fill = green] coordinates {(1, 13125) (2, 25539) (3, 25460) (4, 28985) (5, 30271) (6, 30335) (7, 25506) (8, 26033) (9, 25300) (10, 23696) (11, 44030)};
            \addplot[color = cyan, fill = cyan] coordinates {(1, 1405) (2, 4626) (3, 4024) (4, 4632) (5, 4888) (6, 3382) (7, 3581) (8, 4208) (9, 4110) (10, 3627) (11, 8271)};
            \addplot[color = magenta, fill = magenta] coordinates {(1, 8835) (2, 14723) (3, 16277) (4, 15717) (5, 15315) (6, 12703) (7, 12630) (8, 11545) (9, 11589) (10, 9575) (11, 17174)};
            \addplot[color = teal, fill = teal] coordinates {(1, 7160) (2, 10714) (3, 9872) (4, 10118) (5, 10022) (6, 8869) (7, 8925) (8, 8011) (9, 7862) (10, 7420) (11, 11667)};
            \legend{Car, Pedestrian, Truck, Bus, Motorcycle, Bicycle}
        \end{semilogyaxis}
    \end{tikzpicture}
    \caption{Distribution of the distance of \textit{valid} objects from the ego vehicle across the SimBEV dataset.} \label{appfig:distance-dist}
\end{figure}

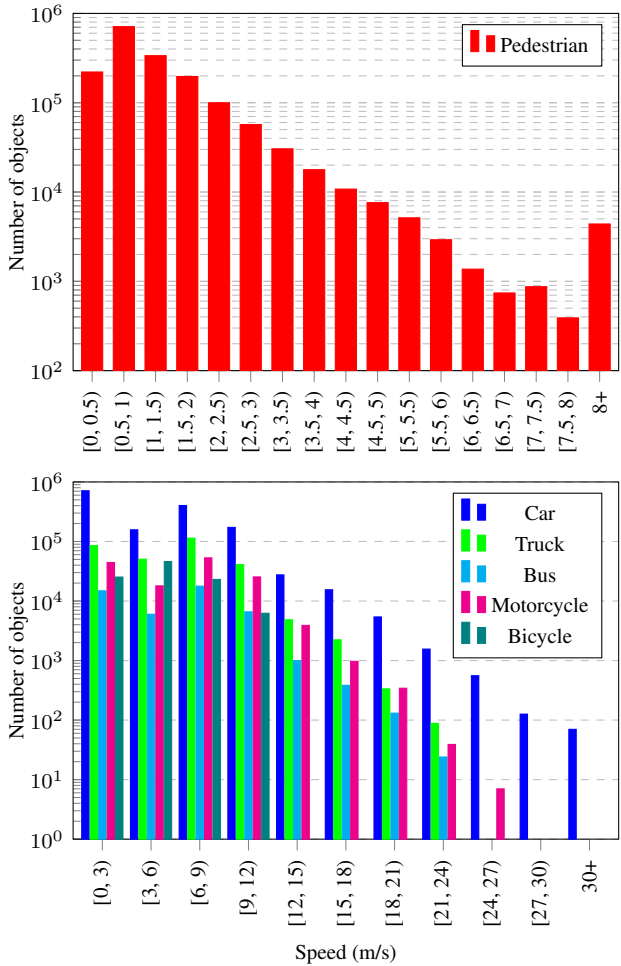
\begin{figure}[ht]
    \centering
    \setlength{\belowcaptionskip}{-24 pt}
    \begin{tikzpicture}
        \begin{semilogyaxis}[title = {}, xlabel near ticks, ylabel = {Number of objects}, ylabel near ticks, ylabel shift = {-6 pt}, xmin = 0.4, xmax = 17.6, ymin = 100, ymax = 1000000, ybar = {0.4 pt}, bar width = {8 pt}, xtick = {1, 2, 3, 4, 5, 6, 7, 8, 9, 10, 11, 12, 13, 14, 15, 16, 17}, xticklabels = {[0{,} 0.5), [0.5{,} 1), [1{,} 1.5), [1.5{,} 2), [2{,} 2.5), [2.5{,} 3), [3{,} 3.5), [3.5{,} 4), [4{,} 4.5), [4.5{,} 5), [5{,} 5.5), [5.5{,} 6), [6{,} 6.5), [6.5{,} 7), [7{,} 7.5), [7.5{,} 8), 8+}, xticklabel style = {rotate = 90, anchor = east}, ytick = {100, 1000, 10000, 100000, 1000000}, label style = {font = \footnotesize}, tick pos = left, tick label style = {font = \footnotesize}, legend pos = north east, legend style = {font = \footnotesize}, ymajorgrids = true, yminorgrids = true, grid style = dashed, width = 1.06\linewidth, height = 0.76\linewidth]
            \addplot[color = red, fill = red] coordinates {(1, 220987) (2, 711802) (3, 336667) (4, 196820) (5, 99965) (6, 56941) (7, 30475) (8, 17789) (9, 10792) (10, 7606) (11, 5147) (12, 2924) (13, 1370) (14, 741) (15, 871) (16, 390) (17, 4389)};
            \legend{Pedestrian}
        \end{semilogyaxis}
    \end{tikzpicture}
    \begin{tikzpicture}
        \begin{semilogyaxis}[title = {}, xlabel = {Speed (m/s)}, xlabel near ticks, ylabel = {Number of objects}, ylabel near ticks, ylabel shift = {-6 pt}, xmin = 0.4, xmax = 11.6, ymin = 1, ymax = 1000000, ybar = {0.4 pt}, bar width = {2.8 pt}, xtick = {1, 2, 3, 4, 5, 6, 7, 8, 9, 10, 11}, xticklabels = {[0{,} 3), [3{,} 6), [6{,} 9), [9{,} 12), [12{,} 15), [15{,} 18), [18{,} 21), [21{,} 24), [24{,} 27), [27{,} 30), 30+}, xticklabel style = {rotate = 90, anchor = east}, ytick = {1, 10, 100, 1000, 10000, 100000, 1000000}, label style = {font = \footnotesize}, tick pos = left, tick label style = {font = \footnotesize}, legend pos = north east, legend style = {font = \footnotesize}, ymajorgrids = true, grid style = dashed, width = 1.06\linewidth, height = 0.76\linewidth]
            \addplot[color = blue, fill = blue] coordinates {(1, 712094) (2, 157149) (3, 402544) (4, 172628) (5, 27485) (6, 15449) (7, 5406) (8, 1558) (9, 557) (10, 126) (11, 70)};
            \addplot[color = green, fill = green] coordinates {(1, 85285) (2, 50492) (3, 113952) (4, 41052) (5, 4847) (6, 2230) (7, 334) (8, 88)};
            \addplot[color = cyan, fill = cyan] coordinates {(1, 14849) (2, 6005) (3, 17773) (4, 6588) (5, 1000) (6, 384) (7, 131) (8, 24)};
            \addplot[color = magenta, fill = magenta] coordinates {(1, 44343) (2, 17977) (3, 53109) (4, 25400) (5, 3895) (6, 970) (7, 343) (8, 39) (9, 7)};
            \addplot[color = teal, fill = teal] coordinates {(1, 25269) (2, 46072) (3, 23070) (4, 6229)};
            \legend{Car, Truck, Bus, Motorcycle, Bicycle}
        \end{semilogyaxis}
    \end{tikzpicture}
    \caption{Breakdown of the speed of \textit{valid} objects across the SimBEV dataset.} \label{appfig:velocity-dist}
\end{figure}

\begin{figure}[t]
    \centering
    \setlength{\belowcaptionskip}{-8 pt}
    \begin{tikzpicture}
        \begin{semilogyaxis}[title = {}, xlabel = {}, xlabel near ticks, ylabel = {Number of points}, ylabel near ticks, ylabel shift = {-6 pt}, xmin = 0, xmax = 120, ymin = 1, ymax = 100000, xtick = {}, xticklabel = {\empty}, ytick = {1, 10, 100, 1000, 10000, 100000}, label style = {font = \footnotesize}, tick pos = left, tick label style = {font = \footnotesize}, legend pos = north east, legend style = {font = \footnotesize, legend columns = 3}, width = 1.06\linewidth, height = \linewidth]
            \addlegendimage{only marks, color = blue}\addlegendentry{Car}
            \addlegendimage{only marks, color = red}\addlegendentry{Pedestrian}
            \addlegendimage{only marks, color = green}\addlegendentry{Truck}
            \addlegendimage{only marks, color = cyan}\addlegendentry{Bus}
            \addlegendimage{only marks, color = magenta}\addlegendentry{Motorcycle}
            \addlegendimage{only marks, color = teal}\addlegendentry{Bicycle}
            \addplot graphics[xmin = 0, xmax = 120, ymin = 1, ymax = 100000] {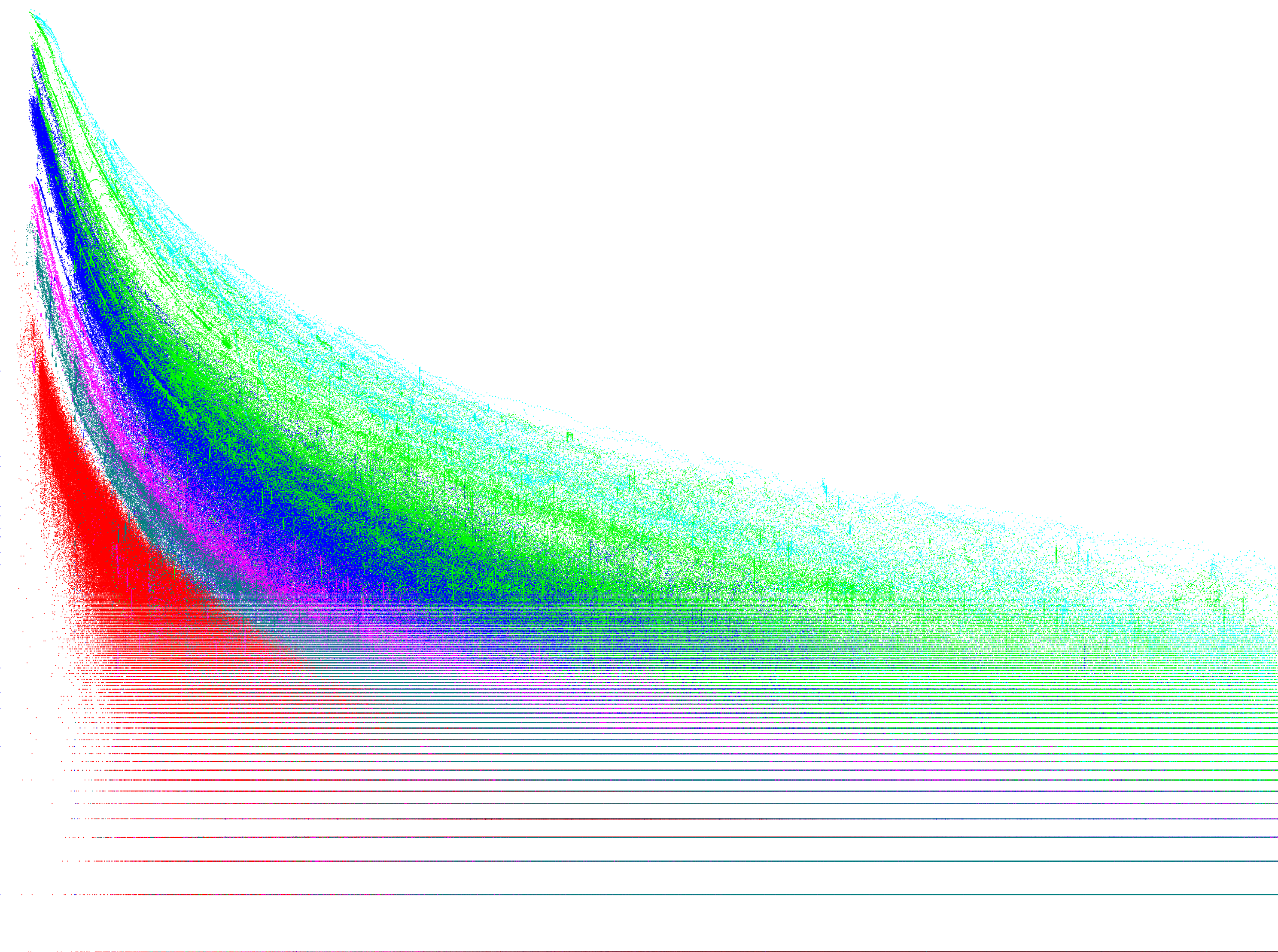};
        \end{semilogyaxis}
    \end{tikzpicture}
    \begin{tikzpicture}
        \begin{semilogyaxis}[title = {}, xlabel = {Distance (m)}, xlabel near ticks, ylabel = {Number of points}, ylabel near ticks, ylabel shift = {-6 pt}, xmin = 0, xmax = 120, ymin = 1, ymax = 10000, xtick = {0, 20, 40, 60, 80, 100, 120}, ytick = {1, 10, 100, 1000, 10000}, label style = {font = \footnotesize}, tick pos = left, tick label style = {font = \footnotesize}, legend pos = north east, legend style = {font = \footnotesize, legend columns = 3}, width = 1.06\linewidth, height = \linewidth]
            \addlegendimage{only marks, color = blue}\addlegendentry{Car}
            \addlegendimage{only marks, color = red}\addlegendentry{Pedestrian}
            \addlegendimage{only marks, color = green}\addlegendentry{Truck}
            \addlegendimage{only marks, color = cyan}\addlegendentry{Bus}
            \addlegendimage{only marks, color = magenta}\addlegendentry{Motorcycle}
            \addlegendimage{only marks, color = teal}\addlegendentry{Bicycle}
            \addplot graphics[xmin = 0, xmax = 120, ymin = 1, ymax = 10000] {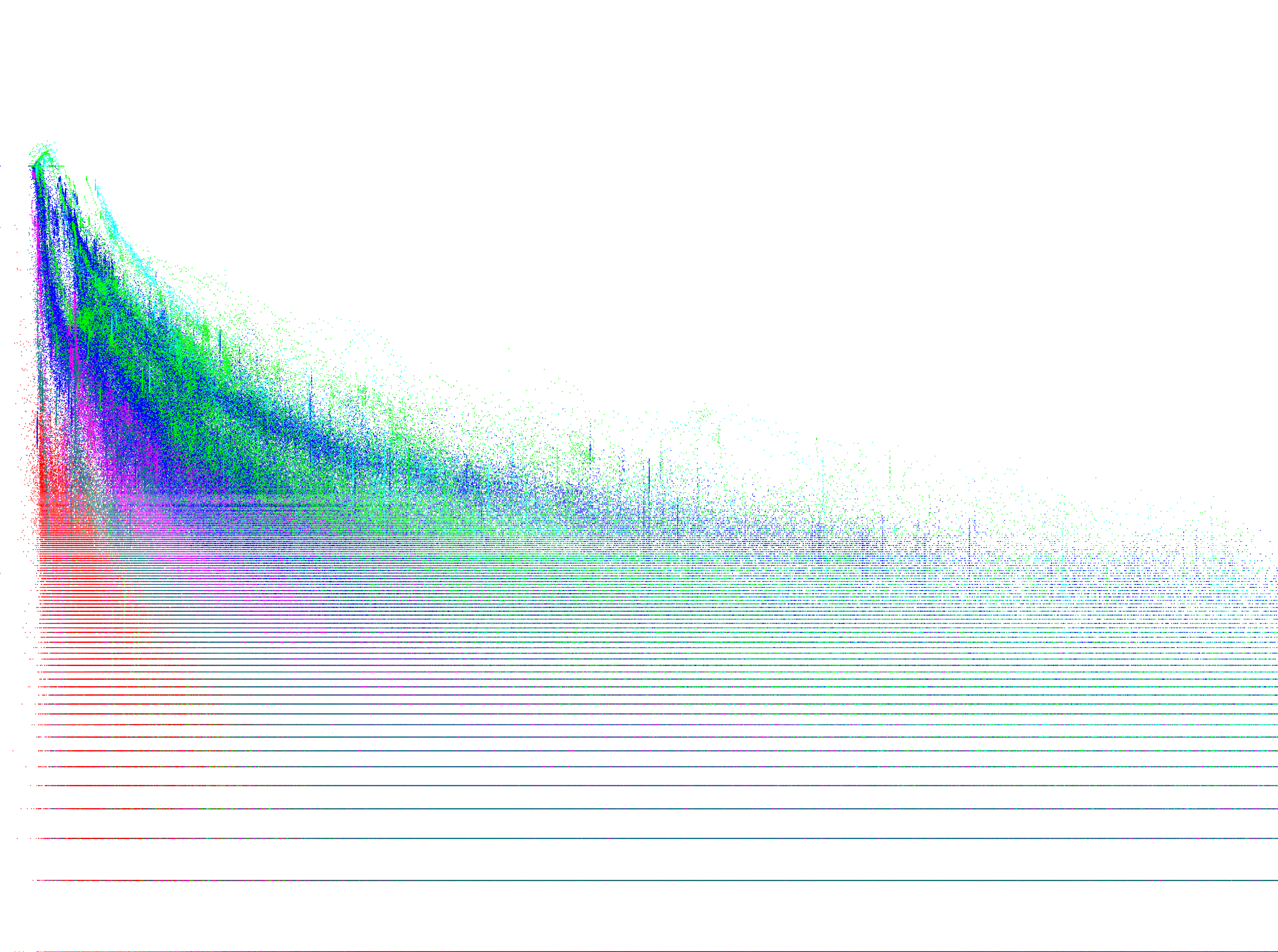};
        \end{semilogyaxis}
    \end{tikzpicture}
    \caption{Distribution of the number of lidar (top) and radar (bottom) points within \textit{valid} 3D object bounding boxes with respect to distance from the ego vehicle.} \label{appfig:num-points-dist}
\end{figure}

\begin{figure*}[t]
    \centering
    \begin{subfigure}{0.24\linewidth}
        \includegraphics[width=\linewidth]{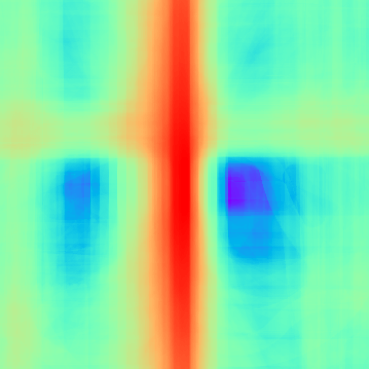}
        \caption{Road} \label{subfig:road}
    \end{subfigure}
    \begin{subfigure}{0.24\linewidth}
        \includegraphics[width=\linewidth]{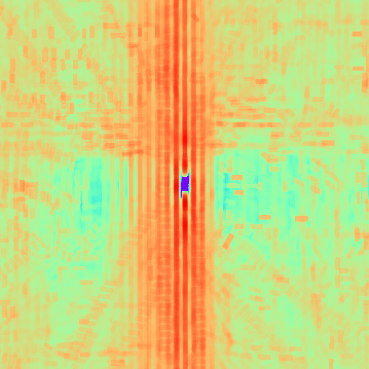}
        \caption{Car} \label{subfig:car}
    \end{subfigure}
    \begin{subfigure}{0.24\linewidth}
        \includegraphics[width=\linewidth]{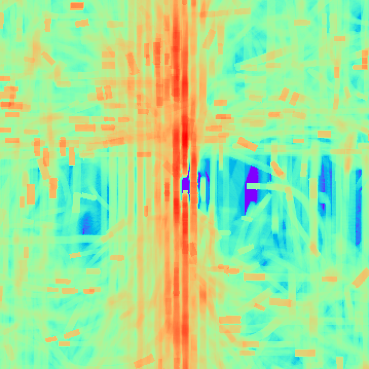}
        \caption{Truck} \label{subfig:truck}
    \end{subfigure}
    \begin{subfigure}{0.24\linewidth}
        \includegraphics[width=\linewidth]{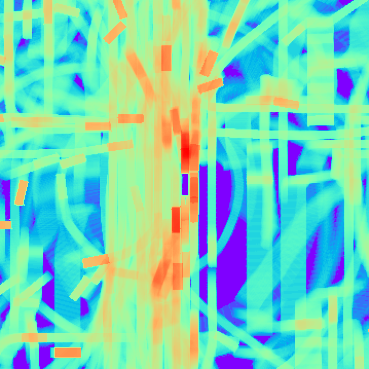}
        \caption{Bus} \label{subfig:bus}
    \end{subfigure}
    \begin{subfigure}{0.24\linewidth}
        \includegraphics[width=\linewidth]{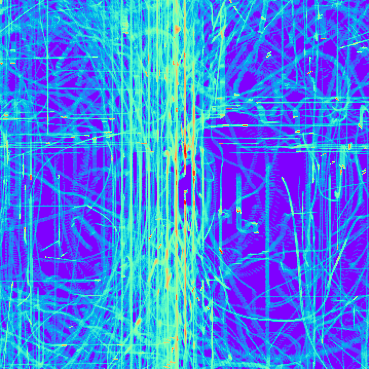}
        \caption{Motorcycle} \label{subfig:motorcycle}
    \end{subfigure}
    \begin{subfigure}{0.24\linewidth}
        \includegraphics[width=\linewidth]{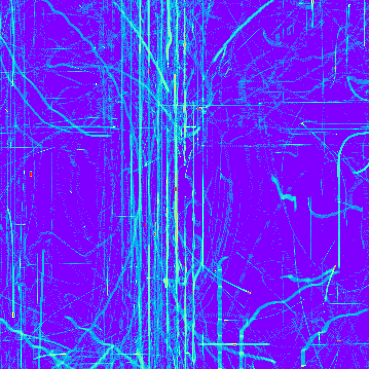}
        \caption{Bicycle} \label{subfig:bicycle}
    \end{subfigure}
    \begin{subfigure}{0.24\linewidth}
        \includegraphics[width=\linewidth]{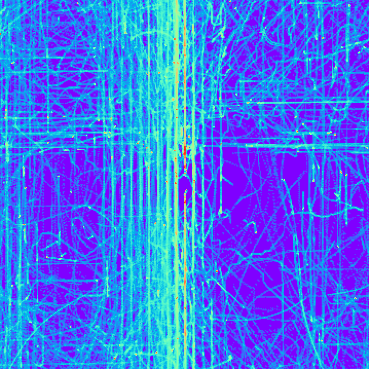}
        \caption{Rider} \label{subfig:rider}
    \end{subfigure}
    \begin{subfigure}{0.24\linewidth}
        \includegraphics[width=\linewidth]{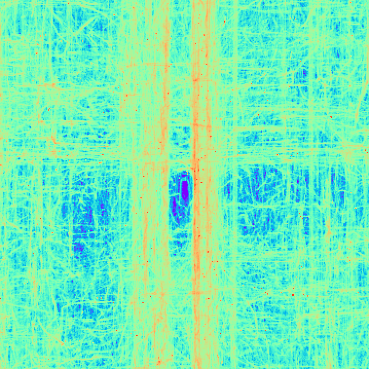}
        \caption{Pedestrian} \label{subfig:pedestrian}
    \end{subfigure}
    \setlength{\abovecaptionskip}{8 pt}
    \setlength{\belowcaptionskip}{-8 pt}
    \caption{Logarithmic BEV heat maps of the SimBEV dataset for different classes.} \label{appfig:bev-heat-map}
\end{figure*}

Because \cref{appfig:vehicle-ped-dist} shows the total number of spawned vehicles and pedestrians, many of which may be far from the ego vehicle, it may not fully represent what the ego vehicle observes. Hence in \cref{appfig:bbox-dist} we break down the number of \textit{valid} 3D object bounding boxes per frame by class. The distribution of the bounding boxes is similar to \cite{caesar2020nuscenes}, although our dataset offers a sizable number of frames with many (65+) \textit{valid} \textit{car}/\textit{pedestrian} bounding boxes as well. As expected, due to having fewer models in CARLA's vehicle library, the vast majority of frames only include a handful of trucks, buses, motorcycles, and bicycles.

\Cref{appfig:distance-dist} to \cref{appfig:num-points-dist} provide more insight into the SimBEV dataset. \Cref{appfig:distance-dist} shows that the distances of \textit{valid} 3D object bounding boxes from the ego vehicle are nearly uniformly distributed for all classes, in contrast to \cite{caesar2020nuscenes}, which is likely due to the higher density and range of our lidar point cloud. \Cref{appfig:velocity-dist} shows a reasonable speed range for all classes, which is comparable to \cite{caesar2020nuscenes} with a few exceptions. Our dataset has a large number of running pedestrians (3+ m/s), which can serve as edge cases for perception and behavior prediction algorithms. For other classes, our data was collected from both urban and highway environments (unlike \cite{caesar2020nuscenes}, which only collected data from urban environments), leading to many fast-moving objects. \cref{appfig:num-points-dist} shows the distribution of the number of lidar and radar points within \textit{valid} 3D object bounding boxes with respect to distance from the ego vehicle. Consistent with \cite{caesar2020nuscenes}, larger object bounding boxes have more points inside and the number of points for all classes decreases with increasing distance.

Finally, logarithmic BEV ground truth heat maps for all classes of the SimBEV dataset are shown in \cref{appfig:bev-heat-map}. As expected, \textit{road} is concentrated in the direction of travel of the ego vehicle, which also results in the concentration of labels of all vehicular classes in that region. In contrast, \textit{pedestrian} labels are relatively evenly distributed.

\section{3D Object Detection Evaluation}

For both approaches to evaluating the results of 3D object detection, AP is calculated from the area under the precision-recall curve. For the first method, we use IoU thresholds of $\mathbb{T} = \{0.3, 0.4, 0.5, 0.6, 0.7, 0.8, 0.9\}$ to match the bounding boxes. For the second method, similar to \cite{caesar2020nuscenes}, we use distance thresholds of $\mathbb{T} = \{0.5, 1, 2, 4\}$ m to match the bounding boxes. For both approaches, we define mAP as the average over all classes and all matching thresholds:
\begin{equation} \label{eq:map}
    \mathrm{mAP} = \frac{1}{|\mathbb{T}||\mathbb{C}|}\sum_{t \in \mathbb{T}, c \in \mathbb{C}}\mathrm{AP}_{t, c}.
\end{equation}

Similar to \cite{caesar2020nuscenes}, we measure a set of True Positive metrics (TP metrics) for each predicted bounding box that is matched to a ground truth bounding box: Average Translation Error (ATE), which is the Euclidean distance (in m) between box centers; Average Orientation Error (AOE), which is the smallest yaw angle difference (in rad) between the two boxes; Average Scale Error (ASE), which is equal to one minus the 3D IoU value of the two boxes after aligning for orientation and translation; and Average Velocity Error (AVE), which is the L2 norm of the difference in box velocities (in m/s). The mean TP metric (mTP) for each metric is computed by averaging over all classes and thresholds:
\begin{equation} \label{eq:mtp}
    \mathrm{mTP} = \frac{1}{|\mathbb{T}||\mathbb{C}|}\sum_{t \in \mathbb{T}, c \in \mathbb{C}}\mathrm{TP}_{t, c}.
\end{equation}

Finally, similar to \cite{caesar2020nuscenes}, we define the SimBEV Detection Score (SDS) as:
\begin{equation} \label{eq:sds}
    \mathrm{SDS} = \frac{1}{8}\Big(4\,\mathrm{mAP} + \sum_{\mathrm{mTP} \in \mathbb{TP}}(1-\min(1, \mathrm{mTP}))\Big).
\end{equation}

\section{Model Implementation} \label{appsec:model-implementation}

All variants of BEVFusion \cite{liu2022bevfusion} and UniTR \cite{wang2023unitr} were trained on an Nvidia DGX A100 640GB node using the settings and hyperparameters used by their authors for benchmarking on the nuScenes dataset \cite{caesar2020nuscenes}. SimBEV dataset data were augmented (translated, rotated, scaled) during training for all models.

\section{Comprehensive Evaluation Results} \label{appsec:comp-eval-results}

\begin{table*}[ht]
    \centering
    \footnotesize
    \begin{tabular}{l c c c c c c c c c c c}
        \toprule
        \textbf{Class} & \textbf{Model} & \textbf{Modality} & \textbf{0.1} & \textbf{0.2} & \textbf{0.3} & \textbf{0.4} & \textbf{0.5} & \textbf{0.6} & \textbf{0.7} & \textbf{0.8} & \textbf{0.9} \\
        \toprule
        \multirow{5}*{Road} & BEVFusion-C & C & 59.5 & 67.1 & 71.5 & 74.5 & 76.0 & 75.2 & 72.6 & 68.9 & 62.3 \\
         & BEVFusion-L & L & 48.6 & 55.9 & 66.1 & 85.1 & 87.7 & 87.2 & 84.9 & 81.1 & 74.6 \\
         & BEVFusion & C + L & \bb{59.7} & \bb{72.0} & \bb{80.0} & \bb{85.5} & \bb{88.4} & \bb{88.1} & \bb{85.9} & \bb{82.4} & \bb{76.3} \\
         & UniTR & C + L & \gb{85.7} & \gb{89.1} & \gb{91.0} & \gb{92.2} & \gb{92.8} & \gb{92.5} & \gb{91.4} & \gb{89.3} & \gb{85.5} \\
         & UniTR+LSS & C + L & \rb{86.0} & \rb{89.4} & \rb{91.3} & \rb{92.6} & \rb{93.3} & \rb{93.0} & \rb{92.0} & \rb{90.1} & \rb{86.4} \\
        \midrule
        \multirow{5}*{Car} & BEVFusion-C & C & 3.5 & 8.0 & 18.8 & 22.4 & 17.2 & 11.3 & 9.7 & 8.7 & 6.3 \\
         & BEVFusion-L & L & 5.3 & 37.8 & 56.5 & 67.1 & 70.6 & 63.6 & 51.8 & 37.6 & 18.8 \\
         & BEVFusion & C + L & \bb{11.7} & \bb{39.4} & \bb{58.6} & \bb{69.4} & \bb{72.7} & \bb{65.5} & \bb{54.0} & \bb{40.1} & \bb{20.5} \\
         & UniTR & C + L & \gb{31.2} & \gb{49.6} & \rb{63.1} & \rb{71.3} & \rb{73.8} & \rb{67.4} & \rb{57.4} & \rb{45.8} & \rb{29.8} \\
         & UniTR+LSS & C + L & \rb{32.2} & \rb{50.6} & \gb{63.0} & \gb{70.9} & \gb{72.8} & \gb{66.1} & \gb{55.8} & \gb{44.3} & \gb{28.8} \\
        \midrule
        \multirow{5}*{Truck} & BEVFusion-C & C & 2.1 & 6.7 & 11.7 & 9.8 & 5.1 & 2.1 & 0.4 & 0.0 & 0.0 \\
         & BEVFusion-L & L & 11.4 & 44.7 & 61.2 & \gb{70.6} & \gb{73.5} & \gb{67.4} & \gb{55.2} & 39.5 & 16.3 \\
         & BEVFusion & C + L & \bb{12.3} & \bb{47.0} & \bb{61.4} & \rb{70.9} & \rb{74.5} & \rb{69.2} & \rb{57.6} & \rb{43.2} & \bb{20.6} \\
         & UniTR & C + L & \gb{33.3} & \gb{51.2} & \gb{61.7} & 67.2 & 67.7 & 61.3 & 51.9 & \bb{40.1} & \gb{22.7} \\
         & UniTR+LSS & C + L & \rb{34.4} & \rb{53.2} & \rb{63.6} & \bb{69.0} & \bb{69.4} & \bb{63.4} & \bb{53.6} & \gb{41.6} & \rb{23.6} \\
        \midrule
        \multirow{5}*{Bus} & BEVFusion-C & C & 2.1 & 9.0 & 19.9 & 24.6 & 22.9 & 16.8 & 10.3 & 6.0 & 1.1 \\
         & BEVFusion-L & L & 19.1 & \gb{56.9} & \rb{72.0} & \rb{79.7} & \rb{81.5} & \rb{78.1} & \rb{69.7} & \rb{59.4} & \rb{44.1} \\
         & BEVFusion & C + L & \bb{19.7} & \bb{56.8} & \gb{70.3} & \gb{78.2} & \gb{80.8} & \gb{77.2} & \gb{68.9} & \gb{59.0} & \rb{44.1} \\
         & UniTR & C + L & \gb{39.5} & 53.5 & 56.7 & 55.5 & 51.7 & 45.1 & 37.8 & 29.8 & \bb{17.9} \\
         & UniTR+LSS & C + L & \rb{44.4} & \rb{57.2} & \bb{61.7} & \bb{62.0} & \bb{58.5} & \bb{51.4} & \bb{42.9} & \bb{33.5} & \gb{21.3} \\
        \midrule
        \multirow{5}*{Motorcycle} & BEVFusion-C & C & 0.3 & 0.7 & 0.0 & 0.0 & 0.0 & 0.0 & 0.0 & 0.0 & 0.0 \\
         & BEVFusion-L & L & 4.8 & \bb{13.7} & \gb{23.6} & 32.7 & 32.5 & 15.8 & 0.8 & 0.0 & 0.0 \\
         & BEVFusion & C + L & \bb{5.0} & 13.5 & \gb{23.6} & \bb{34.6} & \gb{36.3} & \bb{18.3} & \bb{1.5} & 0.0 & 0.0 \\
         & UniTR & C + L & \gb{6.2} & \gb{17.6} & \rb{29.1} & \rb{37.4} & \rb{36.5} & \rb{22.7} & \rb{7.2} & \rb{0.3} & 0.0 \\
         & UniTR+LSS & C + L & \rb{8.3} & \rb{19.5} & \rb{29.1} & \gb{36.3} & \bb{35.9} & \gb{21.6} & \gb{5.8} & \rb{0.3} & 0.0 \\
        \midrule
        \multirow{5}*{Bicycle} & BEVFusion-C & C & 0.0 & 0.0 & 0.0 & 0.0 & 0.0 & 0.0 & 0.0 & 0.0 & 0.0 \\
         & BEVFusion-L & L & 1.8 & 5.1 & 10.0 & \gb{13.3} & \bb{3.6} & 0.0 & 0.0 & 0.0 & 0.0 \\
         & BEVFusion & C + L & \bb{1.9} & \bb{5.6} & \rb{11.1} & \bb{12.7} & \bb{3.6} & 0.0 & 0.0 & 0.0 & 0.0 \\
         & UniTR & C + L & \rb{3.3} & \rb{7.9} & \gb{11.0} & \rb{13.6} & \rb{11.4} & \rb{5.5} & \gb{0.7} & 0.0 & 0.0 \\
         & UniTR+LSS & C + L & \gb{3.2} & \gb{7.7} & \bb{10.5} & 10.9 & \gb{6.3} & \gb{2.7} & \rb{1.4} & \rb{0.2} & 0.0 \\
        \midrule
        \multirow{5}*{Rider} & BEVFusion-C & C & 0.3 & 0.1 & 0.0 & 0.0 & 0.0 & 0.0 & 0.0 & 0.0 & 0.0 \\
         & BEVFusion-L & L & 4.6 & 11.7 & 20.8 & 30.3 & 18.4 & 0.5 & 0.0 & 0.0 & 0.0 \\
         & BEVFusion & C + L & \bb{4.8} & \bb{11.9} & \bb{21.0} & \bb{31.0} & \bb{23.3} & \bb{1.2} & 0.0 & 0.0 & 0.0 \\
         & UniTR & C + L & \gb{5.3} & \gb{15.5} & \rb{25.7} & \rb{35.7} & \rb{36.2} & \rb{17.4} & \rb{1.7} & 0.0 & 0.0 \\
         & UniTR+LSS & C + L & \rb{6.5} & \rb{15.7} & \gb{24.4} & \gb{32.1} & \gb{31.6} & \gb{14.8} & \gb{1.3} & 0.0 & 0.0 \\
        \midrule
        \multirow{5}*{Pedestrian} & BEVFusion-C & C & \bb{0.2} & 0.0 & 0.0 & 0.0 & 0.0 & 0.0 & 0.0 & 0.0 & 0.0 \\
         & BEVFusion-L & L & \rb{3.1} & 9.6 & 17.6 & \bb{28.4} & \bb{18.9} & \bb{0.1} & 0.0 & 0.0 & 0.0 \\
         & BEVFusion & C + L & \rb{3.1} & \bb{9.9} & \gb{18.3} & \gb{28.7} & \gb{20.2} & \gb{0.2} & 0.0 & 0.0 & 0.0 \\
         & UniTR & C + L & \gb{3.0} & \rb{11.1} & \rb{19.9} & \rb{30.2} & \rb{27.5} & \rb{3.9} & 0.0 & 0.0 & 0.0 \\
         & UniTR+LSS & C + L & \rb{3.1} & \gb{10.2} & \bb{17.7} & 25.1 & 12.9 & \gb{0.2} & 0.0 & 0.0 & 0.0 \\
        \midrule
        \midrule
        \multirow{5}*{Mean} & BEVFusion-C & C & 8.5 & 11.4 & 15.2 & 16.4 & 15.2 & 13.2 & 11.6 & 10.5 & 8.7 \\
         & BEVFusion-L & L & 12.4 & 29.4 & 41.0 & \gb{50.9} & \bb{48.3} & \bb{39.1} & \gb{32.8} & \gb{27.2} & 19.2 \\
         & BEVFusion & C + L & \bb{14.8} & \bb{32.0} & \bb{43.0} & \rb{51.3} & \rb{50.0} & \rb{40.0} & \rb{33.5} & \rb{28.1} & \rb{20.2} \\
         & UniTR & C + L & \gb{25.9} & \gb{36.9} & \gb{44.8} & \bb{50.4} & \gb{49.7} & \gb{39.5} & 31.0 & 25.7 & \bb{19.5} \\
         & UniTR+LSS & C + L & \rb{27.3} & \rb{38.0} & \rb{45.2} & 49.9 & 47.6 & \bb{39.1} & \bb{31.6} & \bb{26.2} & \gb{20.0} \\
        \bottomrule
    \end{tabular}
    \caption{BEV segmentation IoUs (in \%) by class and IoU threshold for different models evaluated on the SimBEV dataset \textit{test} set. The top three values are indicated in \rb{red}, \gb{green}, and \bb{blue}, respectively.} \label{apptable:comp-seg}
\end{table*}
\begin{table*}[ht]
    \centering
    \footnotesize
    \begin{tabular}{l c c c c c c c}
        \toprule
        \multirow{2}*{\textbf{Class}} & \multirow{2}*{\textbf{Model}} & \multirow{2}*{\textbf{Modality}} & \textbf{mAP} & \textbf{mATE} & \textbf{mAOE} & \textbf{mASE} & \textbf{mAVE} \\
         & & & \textbf{(\%) $\uparrow$} & \textbf{(m) $\downarrow$} & \textbf{(rad) $\downarrow$} & \textbf{$\downarrow$} & \textbf{(m/s) $\downarrow$} \\
        \toprule
        \multirow{5}*{Car} & BEVFusion-C & C & 12.5 & 0.518 & 0.710 & 0.177 & 5.67 \\
         & BEVFusion-L & L & \gb{41.0} & 0.129 & \gb{0.080} & 0.113 & 1.40 \\
         & BEVFusion & C + L & \rb{41.1} & \bb{0.128} & \rb{0.078} & \bb{0.112} & \bb{1.37} \\
         & UniTR & C + L & 38.8 & \gb{0.100} & 0.123 & \rb{0.084} & \gb{0.49} \\
         & UniTR+LSS & C + L & \bb{39.8} & \rb{0.099} & \bb{0.106} & \gb{0.087} & \rb{0.47} \\
        \midrule
        \multirow{5}*{Truck} & BEVFusion-C & C & 14.1 & 0.568 & 0.902 & 0.123 & 6.64 \\
         & BEVFusion-L & L & \rb{38.7} & \bb{0.143} & \rb{0.040} & 0.100 & 1.73 \\
         & BEVFusion & C + L & \rb{38.7} & 0.149 & \gb{0.042} & \bb{0.096} & \bb{1.70} \\
         & UniTR & C + L & \bb{36.2} & \rb{0.108} & 0.098 & \rb{0.066} & \gb{0.56} \\
         & UniTR+LSS & C + L & \gb{36.6} & \gb{0.110} & \bb{0.096} & \gb{0.070} & \rb{0.55} \\
        \midrule
        \multirow{5}*{Bus} & BEVFusion-C & C & 17.4 & 0.967 & 1.225 & \rb{0.020} & 5.80 \\
         & BEVFusion-L & L & \rb{30.6} & \bb{0.159} & \gb{0.044} & 0.067 & \bb{2.32} \\
         & BEVFusion & C + L & \gb{30.5} & 0.164 & \rb{0.037} & 0.058 & 2.42 \\
         & UniTR & C + L & 25.6 & \rb{0.114} & 0.143 & \bb{0.038} & \gb{0.85} \\
         & UniTR+LSS & C + L & \bb{27.4} & \gb{0.124} & \bb{0.121} & \gb{0.036} & \rb{0.75} \\
        \midrule
        \multirow{5}*{Motorcycle} & BEVFusion-C & C & 11.6 & 0.261 & 0.688 & 0.135 & 6.63 \\
         & BEVFusion-L & L & 39.6 & \gb{0.091} & \gb{0.080} & 0.131 & 1.74 \\
         & BEVFusion & C + L & \gb{40.1} & \bb{0.092} & \rb{0.071} & \bb{0.125} & \bb{1.59} \\
         & UniTR & C + L & \bb{39.8} & \rb{0.074} & 0.102 & \gb{0.092} & \rb{0.54} \\
         & UniTR+LSS & C + L & \rb{40.4} & \rb{0.074} & \bb{0.097} & \rb{0.088} & \gb{0.56} \\
        \midrule
        \multirow{5}*{Bicycle} & BEVFusion-C & C & 8.4 & 0.227 & 0.818 & 0.200 & 3.20 \\
         & BEVFusion-L & L & \bb{38.4} & \bb{0.088} & \gb{0.071} & 0.186 & 1.46 \\
         & BEVFusion & C + L & 38.3 & 0.089 & \rb{0.054} & \bb{0.177} & \bb{1.41} \\
         & UniTR & C + L & \gb{39.5} & \gb{0.071} & 0.110 & \gb{0.120} & \rb{0.43} \\
         & UniTR+LSS & C + L & \rb{41.3} & \rb{0.069} & \bb{0.103} & \rb{0.103} & \gb{0.44} \\
        \midrule
        \multirow{5}*{Pedestrian} & BEVFusion-C & C & 0.2 & 0.111 & 1.42 & \rb{0.035} & 1.18 \\
         & BEVFusion-L & L & 37.0 & \bb{0.064} & \gb{0.262} & 0.076 & \bb{0.39} \\
         & BEVFusion & C + L & \bb{37.2} & 0.066 & \rb{0.235} & 0.069 & \gb{0.38} \\
         & UniTR & C + L & \gb{40.1} & \gb{0.056} & 0.376 & \gb{0.048} & \rb{0.21} \\
         & UniTR+LSS & C + L & \rb{40.5} & \rb{0.055} & \bb{0.360} & \bb{0.049} & \rb{0.21} \\
        \midrule
        \midrule
        \multirow{5}*{Mean} & BEVFusion-C & C & 7.0 & 0.337 & 0.943 & 0.106 & 4.98 \\
         & BEVFusion-L & L & \bb{33.9} & \bb{0.105} & \gb{0.086} & 0.107 & 1.49 \\
         & BEVFusion & C + L & \gb{34.1} & 0.107 & \rb{0.077} & \bb{0.101} & \bb{1.46} \\
         & UniTR & C + L & 33.0 & \rb{0.081} & 0.140 & \gb{0.071} & \gb{0.51} \\
         & UniTR+LSS & C + L & \rb{34.2} & \gb{0.083} & \bb{0.131} & \rb{0.069} & \rb{0.49} \\
        \bottomrule
    \end{tabular}
    \caption{3D object detection results for different models evaluated on the SimBEV \textit{test} set using the first method. The top three values are indicated in \rb{red}, \gb{green}, and \bb{blue}, respectively.} \label{apptable:comp-det-iou}
\end{table*}
\begin{table*}[ht]
    \centering
    \footnotesize
    \begin{tabular}{l c c c c c c c}
        \toprule
        \multirow{2}*{\textbf{Class}} & \multirow{2}*{\textbf{Model}} & \multirow{2}*{\textbf{Modality}} & \textbf{mAP} & \textbf{mATE} & \textbf{mAOE} & \textbf{mASE} & \textbf{mAVE} \\
         & & & \textbf{(\%) $\uparrow$} & \textbf{(m) $\downarrow$} & \textbf{(rad) $\downarrow$} & \textbf{$\downarrow$} & \textbf{(m/s) $\downarrow$} \\
        \toprule
        \multirow{5}*{Car} & BEVFusion-C & C & 23.3 & 0.824 & 0.896 & 0.217 & 4.94 \\
         & BEVFusion-L & L & \gb{46.1} & 0.165 & \gb{0.109} & 0.127 & 1.44 \\
         & BEVFusion & C + L & \rb{46.5} & \bb{0.162} & \rb{0.106} & \bb{0.125} & \bb{1.40} \\
         & UniTR & C + L & \rb{46.5} & \gb{0.132} & 0.179 & \rb{0.095} & \gb{0.52} \\
         & UniTR+LSS & C + L & \bb{46.0} & \rb{0.128} & \bb{0.153} & \gb{0.097} & \rb{0.50} \\
        \midrule
        \multirow{5}*{Truck} & BEVFusion-C & C & 20.4 & 0.751 & 0.695 & 0.148 & 5.55 \\
         & BEVFusion-L & L & \rb{46.3} & \bb{0.162} & \rb{0.045} & 0.110 & 1.75 \\
         & BEVFusion & C + L & \gb{46.2} & 0.168 & \gb{0.049} & \bb{0.106} & \bb{1.74} \\
         & UniTR & C + L & 45.2 & \rb{0.123} & \bb{0.134} & \rb{0.074} & \gb{0.58} \\
         & UniTR+LSS & C + L & \bb{45.8} & \gb{0.128} & 0.140 & \gb{0.078} & \rb{0.56} \\
        \midrule
        \multirow{5}*{Bus} & BEVFusion-C & C & 18.7 & 0.829 & 1.185 & \rb{0.022} & 5.55 \\
         & BEVFusion-L & L & 34.1 & \bb{0.169} & \gb{0.049} & 0.072 & \bb{2.40} \\
         & BEVFusion & C + L & \bb{34.3} & 0.176 & \rb{0.040} & 0.063 & 2.44 \\
         & UniTR & C + L & \gb{35.1} & \rb{0.122} & 0.216 & \bb{0.041} & \gb{0.80} \\
         & UniTR+LSS & C + L & \rb{35.2} & \gb{0.129} & \bb{0.176} & \gb{0.037} & \rb{0.72} \\
        \midrule
        \multirow{5}*{Motorcycle} & BEVFusion-C & C & 26.5 & 0.604 & 0.841 & \bb{0.140} & 6.63 \\
         & BEVFusion-L & L & \bb{51.9} & 0.114 & \gb{0.118} & 0.159 & 1.79 \\
         & BEVFusion & C + L & 51.6 & \bb{0.113} & \rb{0.104} & 0.153 & \bb{1.65} \\
         & UniTR & C + L & \rb{52.3} & \rb{0.075} & 0.216 & \gb{0.105} & \gb{0.62} \\
         & UniTR+LSS & C + L & \gb{52.1} & \gb{0.091} & \bb{0.148} & \rb{0.096} & \rb{0.61} \\
        \midrule
        \multirow{5}*{Bicycle} & BEVFusion-C & C & 25.1 & 0.574 & 1.117 & 0.219 & 4.12 \\
         & BEVFusion-L & L & \rb{55.5} & \bb{0.114} & \gb{0.087} & 0.213 & \bb{1.53} \\
         & BEVFusion & C + L & \gb{55.3} & 0.115 & \rb{0.073} & \bb{0.207} & \gb{1.52} \\
         & UniTR & C + L & 54.4 & \gb{0.087} & 0.168 & \gb{0.146} & \rb{0.49} \\
         & UniTR+LSS & C + L & \bb{55.0} & \rb{0.085} & \bb{0.155} & \rb{0.121} & \rb{0.49} \\
        \midrule
        \multirow{5}*{Pedestrian} & BEVFusion-C & C & 18.9 & 0.884 & 1.529 & \rb{0.073} & 1.10 \\
         & BEVFusion-L & L & \gb{54.5} & \bb{0.141} & \gb{0.392} & 0.120 & \bb{0.47} \\
         & BEVFusion & C + L & \rb{54.8} & 0.142 & \rb{0.362} & 0.109 & \bb{0.47} \\
         & UniTR & C + L & 52.8 & \gb{0.118} & 0.499 & \gb{0.079} & \gb{0.29} \\
         & UniTR+LSS & C + L & \bb{53.1} & \rb{0.116} & \bb{0.472} & \bb{0.081} & \rb{0.28} \\
        \midrule
        \midrule
        \multirow{5}*{Mean} & BEVFusion-C & C & 22.1 & 0.744 & 1.044 & 0.137 & 4.65 \\
         & BEVFusion-L & L & \rb{48.1} & \gb{0.144} & \gb{0.133} & 0.134 & 1.56 \\
         & BEVFusion & C + L & \rb{48.1} & \bb{0.146} & \rb{0.122} & \bb{0.127} & \bb{1.54} \\
         & UniTR & C + L & \bb{47.7} & \rb{0.113} & 0.224 & \gb{0.090} & \gb{0.55} \\
         & UniTR+LSS & C + L & \gb{47.8} & \rb{0.113} & \bb{0.207} & \rb{0.085} & \rb{0.53} \\
        \bottomrule
    \end{tabular}
    \caption{3D object detection results for different models evaluated on the SimBEV \textit{test} set using the second method. The top three values are indicated in \rb{red}, \gb{green}, and \bb{blue}, respectively.} \label{apptable:comp-det-distance}
\end{table*}

BEV segmentation IoUs (in \%) by class and IoU threshold for different models are shown in \cref{apptable:comp-seg}, and a breakdown of 3D object detection results by class is shown in \cref{apptable:comp-det-iou} and \cref{apptable:comp-det-distance} for the first and second methods, respectively. As discussed before, the biggest takeaway from the results is that camera-only models (for both BEV segmentation and 3D object detection) perform worse than lidar-only and fusion models.

\newpage

    {
        \small
        \bibliographystyle{ieeenat_fullname}
        \bibliography{main}
    }

\end{document}